\documentclass[journal]{IEEEtran}
\usepackage{cite}

\usepackage[colorlinks, citecolor=green]{hyperref}
\usepackage{booktabs}
\usepackage{subfigure} 
\usepackage{graphicx}
\usepackage{float} 
\usepackage{multirow}
\usepackage{amsmath}
\usepackage{tabularx}
\usepackage{booktabs}

\usepackage{algorithmic}
\usepackage{algorithm}

\usepackage{colortbl}
\usepackage{amssymb}
\usepackage{makecell}
\usepackage{textcomp}
\usepackage{array} 
\usepackage{longtable}
\usepackage{booktabs}
\usepackage{mathptmx}
\usepackage{fixltx2e}

\renewcommand\arraystretch{1.0}
\usepackage[export]{adjustbox}
\begin{document}
%
\title{GeoSAM-Lite: A Lightweight Foundation Model for Onboard Remote Sensing Segmentation}
%
%
%

\author{Yongcong~Wang,
        Jie~Zhang,
        Rui~Jiang,
        Xubing~Yang,
        Ting~Yun,
        and~Li~Zhang
\thanks{Manuscript received April 19, 2005; revised September 17, 2014.}

\thanks{This work is supported in part by National Natural Science Foundation of China (NSFC) under grant (No. 32371876).}

\thanks{Yongcong Wang, Jie Zhang, Xubing Yang, Ting Yun and Li Zhang are  with College of Information Science and Technology \& Artificial Intelligence, Nanjing Forestry University, Nanjing 210016, China (e-mail:\{ycwang1031, lizhang\}@njfu.edu.cn). \textit{(Corresponding author: Li Zhang)}}

\thanks{Rui Jiang is with College of Telecommunications and Information Engineering, Nanjing University of Posts and
Telecommunications, Nanjing 210003, China.}}

\markboth{IEEE GEOSCIENCE AND REMOTE SENSING LETTERS}%
{Wang \MakeLowercase{\textit{et al.}}: GeoSAM-Lite: A Lightweight Foundation Model for Onboard Remote Sensing Segmentation}

\maketitle

\begin{abstract}
The deployment of large-scale foundation models like Segment Anything Model (SAM) on resource-constrained Earth observation platforms is hindered by prohibitive computational costs and the domain shift between natural and remote sensing imagery. To address these challenges, we propose \textit{Geo}spatial \textit{S}egment \textit{A}nything \textit{M}odel-Lite (GeoSAM-Lite), a lightweight, prompt-free segmentation framework designed for efficient onboard remote sensing segmentation. GeoSAM-Lite incorporates two core innovations: (1) Geospatial-Domain Initialization (Geo-Init), a domain-aware pre-training strategy that distills geospatial priors from a specialized teacher to bridge the domain gap; and (2) Feature Fusion Layers (FFL), which recalibrate spatial features and restore high-frequency boundary cues to overcome the capacity bottlenecks of lightweight backbones. Experiments across representative datasets, with a primary focus on cloud scenarios to evaluate performance under extreme scale variations and complex boundaries, demonstrate that GeoSAM-Lite achieves competitive accuracy while reducing parameters by 92.8\% compared to the heavyweight RSAM-Seg. By establishing a superior Pareto frontier between efficiency and fidelity, GeoSAM-Lite offers a practical solution for real-time segmentation on edge devices.
\end{abstract}


%
\IEEEpeerreviewmaketitle

\section{Introduction}

\IEEEPARstart{W}{ith} the rapid evolution of Earth observation technology, high-resolution remote sensing (RS) imagery has become indispensable across a spectrum of applications. Precise semantic segmentation of key targets—with cloud detection \cite{38clouddataset} serving as a highly representative challenge due to its extreme scale variations and complex boundaries—serves as a pivotal step for interpreting these complex geospatial scenes. While deep learning paradigms, particularly Convolutional Neural Networks (CNNs) \cite{38clouddataset}, Vision Transformers (ViTs) \cite{3} and recent Mamba \cite{hdamnet}, have achieved remarkable benchmarks, deploying them on resource-constrained platforms like satellites and UAVs remains challenging. To address this, the community has explored lightweight architectures, including optimized CNNs \cite{14} and efficient ViTs \cite{25}, striving to balance computational efficiency with segmentation accuracy.


\begin{figure}[t]
    \centering
    \includegraphics[width=0.7\linewidth]{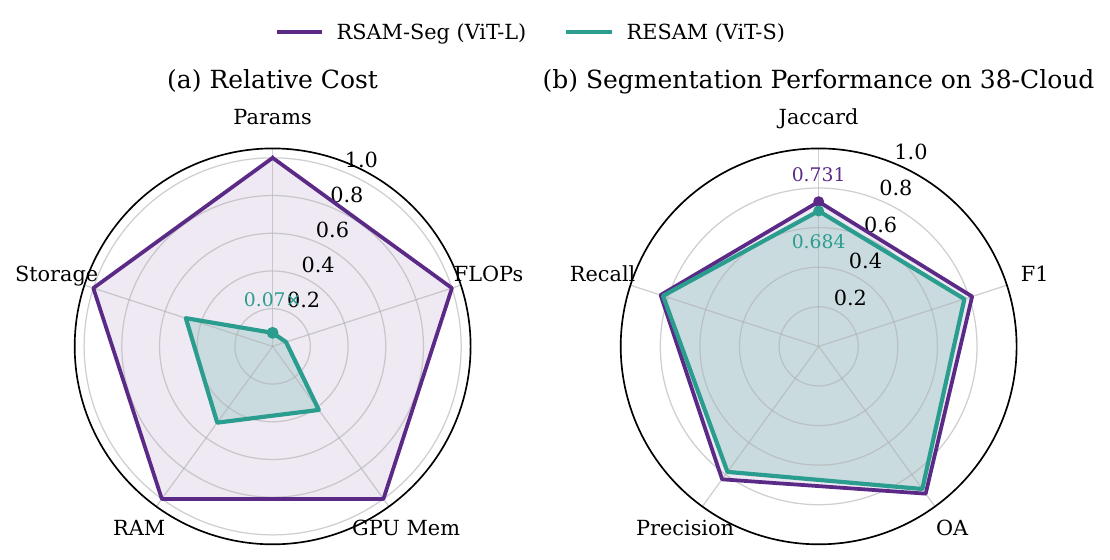}
    \caption{GeoSAM-Lite achieves a superior accuracy--efficiency trade-off.}
    \label{fig:teaser}
\end{figure}

Recently, the Segment Anything Model (SAM) \cite{sam} marked a paradigm shift in computer vision, offering unprecedented zero-shot generalization via promptable segmentation. In the RS domain, SAM has attracted significant attention for automating geospatial analysis, with adaptations like RSPrompter \cite{rsprompter} and specialized fine-tuning strategies \cite{sam_oe} demonstrating promise. Despite this potential, direct deployment of the original SAM faces a critical hurdle: its heavyweight image encoder (ViT-H, over 600M parameters) incurs prohibitive computational costs, rendering it unsuitable for onboard processing.

To alleviate this computational burden, the computer vision community introduced lightweight variants such as MobileSAM \cite{mobilesam} and EfficientSAM \cite{54}, which transfer SAM's capabilities to compact backbones like TinyViT using knowledge distillation. While effective in general domains, these generic lightweight models expose significant limitations when migrated to RS tasks. Primarily, they suffer from a substantial ``domain shift'', as they distill knowledge from a vanilla SAM teacher trained on natural images, thereby lacking essential geospatial priors. Furthermore, aggressive parameter reduction often induces a ``low-pass filtering'' effect \cite{2022vision}, leading to the attenuation of high-frequency details and blurred boundaries—a critical issue for fine-grained satellite imagery targets.

The dual challenges of domain shift and feature degradation in lightweight models motivate the proposal of \textbf{Geo}spatial \textbf{S}egment \textbf{A}nything \textbf{M}odel-Lite (GeoSAM-Lite). Unlike methods that rely on generic distillation, GeoSAM-Lite decouples representation learning into two specialized stages: domain-aware initialization and frequency-spatial adaptation. This design ensures that the model inherits geospatial semantics while explicitly recovering the high-frequency details lost during compression. Our main contributions are as follows.
\begin{enumerate}
    \item We propose GeoSAM-Lite, a prompt-free lightweight framework that reduces parameters by 92.8\% compared to teacher model while maintaining competitive segmentation accuracy.
    \item We introduce Geospatial-Domain Initialization (Geo-Init), a domain-aware pre-training strategy that distills RS-specific knowledge from an expert teacher via masked feature alignment.
    \item We design Feature Fusion Layers (FFL) that integrate spatial recalibration with frequency-domain reconstruction, explicitly compensating for the spectral bias and boundary blurring inherent in compact networks.
\end{enumerate}

\section{Methodology}

\subsection{Overall Framework of GeoSAM-Lite}

\begin{figure*}[!t]
    \centering
    \includegraphics[width=0.8\textwidth]{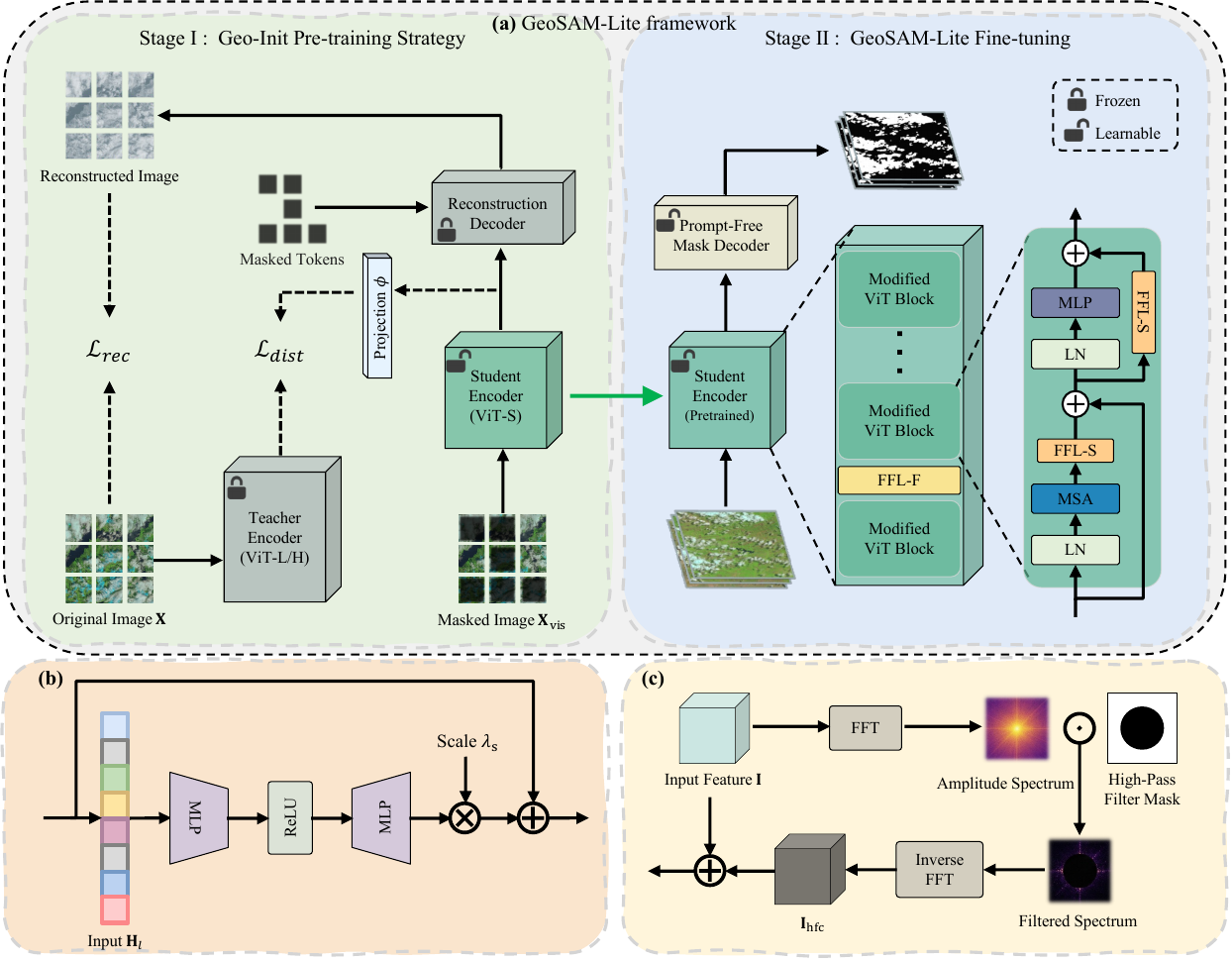} 
    \vspace{-1em}
    \caption{Overview of the proposed GeoSAM-Lite framework, with detailed illustrations of (c) FFL-S and (d) FFL-F modules.}
    \label{fig:overall_framework}
\end{figure*}

As illustrated in Fig.~\ref{fig:overall_framework}, GeoSAM-Lite introduces a decoupled two-stage paradigm to reconcile the trade-off between segmentation fidelity and onboard efficiency by explicitly separating domain alignment from feature refinement. In Stage~I, we employ a Domain-Expert Teacher to distill robust geospatial semantics into a lightweight student via masked feature alignment. In Stage~II, we inject FFL to compensate for the student's limited capacity. By synergizing spatial recalibration and frequency-domain detail restoration, GeoSAM-Lite effectively decouples general representation learning from task-specific adaptation, enabling robust onboard segmentation with minimal parameters.

\subsection{Geo-Init Pre-training Strategy}

The Geo-Init serves as the cornerstone of GeoSAM-Lite's representational capability. Unlike the SAMI \cite{54}, which distills from a general-purpose vision model, Geo-Init employs a Domain-Expert Teacher—RSAM-Seg (ViT-L) \cite{rsam_seg}, which has been extensively trained on large-scale RS datasets. This ensures that the knowledge transferred to the lightweight student contains rich, domain-specific priors.

\subsubsection{Masked Image Modeling}
We employ the Masked Image Modeling (MIM) paradigm to force the student encoder to learn holistic context reconstruction. Formally, we represent an input RS image as a patch-token matrix $\mathbf{X} \in \mathbb{R}^{N \times D}$, obtained by partitioning the image into $N$ non-overlapping patches and embedding each patch into a $D$-dimensional token. A random binary mask $\mathbf{M} \in \{0, 1\}^N$ splits the tokens into visible patches $\mathbf{X}_{\text{vis}}$ and masked patches $\mathbf{X}_{\text{mask}}$:
\begin{equation}
    \mathbf{X}_{\text{vis}} = \mathbf{X} \odot (\mathbf{1} - \mathbf{M}), \quad \mathbf{X}_{\text{mask}} = \mathbf{X} \odot \mathbf{M}
\end{equation}
where $\odot$ denotes element-wise multiplication and $\mathbf{1}$ is a matrix of ones. With a masking ratio (typically 75\%), the lightweight encoder $E_S$ processes only $\mathbf{X}_{\text{vis}}$. This sparse input forces the network to infer the missing $\mathbf{X}_{\text{mask}}$ based solely on global context, fostering feature abstraction for geospatial scenes.

\subsubsection{Teacher-Guided Knowledge Distillation}
While MIM drives context understanding, the semantic quality is supervised by the expert teacher. We use the frozen RSAM-Seg as the Domain-Expert Teacher ($E_T$). It receives the complete, unmasked image $\mathbf{X}$ to generate high-fidelity feature representations $\mathbf{F}_T = E_T(\mathbf{X})$. The fully trainable Lightweight Student ($E_S$) receives the masked input to generate latent features $\mathbf{F}_S = E_S(\mathbf{X}_{\text{vis}})$. The distillation objective compels the student to reconstruct the missing details and align its partial-view representation with the teacher's full-view feature space.

\subsection{Task-Specific Fine-tuning with Feature Fusion Layers}

In the second stage, the pre-trained lightweight encoder is augmented with FFL to adapt to specific downstream tasks. We introduce two specialized modules to compensate for the reduced capacity of the ViT-S backbone: FFL-S for spatial-domain recalibration and FFL-F for frequency-domain detail restoration.


\subsubsection{Spatial Recalibration with FFL-S}
RS targets span a wide range of spatial scales. To handle this, the FFL-S operates in the spatial-domain. It is embedded within the ViT blocks to dynamically recalibrate features based on their global significance.
Let $\mathbf{H}_l$ denote the input feature tokens at layer $l$. The FFL-S operation is formulated as a residual bottleneck structure:
\begin{equation}
    \text{FFL-S}(\mathbf{H}_l) = \mathbf{H}_l + \lambda_{\text{s}} \cdot \mathcal{U}\left(\sigma\left(\mathcal{D}(\mathbf{H}_l)\right)\right)
\end{equation}
where $\mathcal{D}(\cdot)$ and $\mathcal{U}(\cdot)$ represent down-projection and up-projection MLPs, respectively, shrinking and expanding the embedding dimension to capture scale-invariant semantics. $\sigma$ denotes the ReLU function, and $\lambda_{\text{s}}$ is a learnable scaling factor initialized to zero to ensure stable training startup.

\subsubsection{High-Frequency Injection with FFL-F}
Lightweight models often act as low-pass filters, blurring fine boundaries—a critical issue for tasks like cloud edge detection. We propose FFL-F to explicitly restore these lost details in the frequency-domain.
Given an input feature map $\mathbf{I} \in \mathbb{R}^{H \times W \times C}$, we first convert it to the frequency domain via the Fast Fourier Transform (FFT). We then apply a high-pass filtering mask $\mathbf{M}_{\text{h}}$ to isolate high-frequency components (HFC).
Let $(u, v)$ be the frequency coordinates relative to the center $(u_0, v_0)$. The mask effectively suppresses low-frequency signals:
\begin{equation}
    \mathbf{M}_{\text{h}}(u, v) = \mathbb{I}\left( \frac{|(u-u_0)(v-v_0)|}{HW} > \tau \right)
\end{equation}
where $\tau$ is a bandwidth threshold and $\mathbb{I}(\cdot)$ is the indicator function. The HFC features $\mathbf{I}_{\text{hfc}}$ are reconstructed via the Inverse FFT:
\begin{equation}
    \mathbf{I}_{\text{hfc}} = \text{IFFT}\left(\text{FFT}(\mathbf{I}) \odot \mathbf{M}_{\text{h}}\right)
\end{equation}
These high-frequency cues, rich in boundary information, are then injected back into the encoder layers, ensuring the model retains crisp segmentation edges even after significant compression.

\subsection{Training Strategy and Optimization Objectives}

We maximize the efficacy of GeoSAM-Lite through a decoupled two-stage training protocol, ensuring that the general capabilities learned from Geo-Init are effectively specialized by FFL.

\subsubsection{Stage~I Representation Learning (Geo-Init)}
The objective of this stage is General Initialization. The lightweight encoder is trained on broad RS datasets without task-specific labels. The composite loss function $\mathcal{L}_{\text{pre}}$ combines reconstruction and feature alignment objectives:
\begin{equation}
\begin{aligned}
    \mathcal{L}_{\text{pre}} = \lambda_{\text{rec}} \mathcal{L}_{\text{rec}} + \lambda_{\text{dist}} \sum_{l \in \Omega} \left\| \phi(\mathbf{F}_S^l) - \mathbf{F}_T^l \right\|_2^2
\end{aligned}
\end{equation}
Here, $\mathcal{L}_{\text{rec}}$ represents the Mean Squared Error (MSE) loss on masked patches, and the second term enforces the feature alignment between student and teacher. $\phi(\cdot)$ is a projection that aligns the student's feature dimension with that of teacher’s.

\subsubsection{Stage~II Task-Specific Fine-Tuning}

The objective of this stage is Targeted Adaptation. We load the pre-trained weights into the GeoSAM-Lite backbone and insert the FFL-S and FFL-F modules. To achieve prompt-free inference, SAM's original prompt encoder and prompt-dependent mask decoder are replaced with a lightweight convolutional decoder that directly maps encoder features to dense segmentation masks. The entire network is then fine-tuned on the specific dataset using full supervision. For binary segmentation tasks, we employ the Binary Cross-Entropy (BCE) loss:
\begin{equation}
    \mathcal{L}_{\text{seg}} = - \frac{1}{N} \sum_{i=1}^{N} \left[ y_i \log(p_i) + (1-y_i) \log(1-p_i) \right]
\end{equation}
where $y_i$ is the ground truth and $p_i$ is the predicted probability.

\section{Experiment Results}

\subsection{Datasets}

To evaluate the proposed GeoSAM-Lite framework, several representative datasets are utilized to verify both its segmentation fidelity and generalization robustness. We strategically select cloud detection as our primary evaluation scenario because it rigorously tests the model against two core challenges: extreme scale variations and the blurring of high-frequency boundaries inherent in lightweight models. To further demonstrate the model's domain-agnostic generalization capability for broader remote sensing target extraction, we also incorporate a farmland scenario. 

Specifically, the 38-Cloud dataset \cite{38clouddataset} collected by Landsat-8 consists of 826 large scenes cropped to $1024 \times 1024$ patches. It is divided into an 8:1:1 ratio, providing 660 training, 83 validation, and 83 testing patches. The CloudSEN12 dataset \cite{cloudsen12_dataset} contains 9,880 regions of interest with scenes categorized across five cloud coverage levels. The images are cropped to $512 \times 512$ pixels, yielding 1,500 training, 187 validation, and 187 testing images. The SPARCS dataset \cite{sparcs_dataset} contains $1000 \times 1000$ pixel subsets of Landsat 8 scenes, annotated as a three-class segmentation task (cloud, cloud shadow, and background), where shadows are merged into the background for binary cloud detection evaluation. It includes 80 training images, 10 validation images, and 10 testing images following an 8:1:1 split format.

\subsection{Implementation Details}
Our experiments are implemented using PyTorch. In Stage~I, the Geo-Init pre-training is conducted on two NVIDIA A40 GPUs (48GB). The model is trained for 800 epochs with a base learning rate of $1.5 \times 10^{-4}$, employing a cosine decay scheduler and a warm-up period. In Stage~II, the fine-tuning is performed on a single NVIDIA RTX 4090 GPU (24GB). We fine-tune the model for 60 epochs on each downstream dataset. The AdamW optimizer is utilized throughout both stages to ensure stable convergence.

\subsection{Performance Comparison}

As shown in Table \ref{tab:RSAMRES}, GeoSAM-Lite consistently outperforms other lightweight baselines across three cloud detection datasets. On the complex CloudSEN12 dataset, it improves the Jaccard index by approximately 15\% over its lightweight competitors. Although a slight performance drop is observed compared to the heavyweight RSAM-Seg, this represents an acceptable trade-off given the significant reduction in parameters. Crucially, our compact model maintains high Recall and Precision, validating its practicality for resource-constrained onboard processing. Visual results in Fig. \ref{fig:qualitative_results} demonstrate that while generic lightweight models blur complex boundaries, GeoSAM-Lite preserves fine details and sharp edges via its Frequency-Spatial Adaptation mechanism.

\begin{table}[!t]
    \centering
    \caption{Quantitative Comparison of Segmentation Performance on Cloud Detection Scenarios}
    \label{tab:RSAMRES}
    \setlength{\tabcolsep}{2.4pt} 
    \renewcommand{\arraystretch}{0.9} 
    \scriptsize
    \begin{tabular}{llccccc}
        \toprule
        \textbf{Dataset} & \textbf{Method} & \textbf{Jaccard} & \textbf{F1 Score} & \textbf{OA}& \textbf{Precision} & \textbf{Recall} \\ 
        \midrule
        \multirow{4}{*}{38-Cloud} 
        & EfficientSAM & 0.6761 & 0.7699 & 0.8729 & 0.7841 & 0.8130 \\
        & MobileSAM & 0.6820 & 0.7687 & 0.8736 & 0.7375 & \textbf{0.8811} \\
        & RSAM-Seg & \textbf{0.7310} & \textbf{0.8152} & \textbf{0.9197} & \textbf{0.8301} & \underline{0.8396} \\
        & GeoSAM-Lite & \underline{0.6840} & \underline{0.7740} & \underline{0.8907} & \underline{0.7837} & 0.8266 \\ 
        \midrule
        \multirow{4}{*}{CloudSEN12} 
        & EfficientSAM & 0.6517 & 0.7765 & 0.8825 & 0.7561 & 0.8039 \\
        & MobileSAM & 0.6397 & 0.7634& 0.8821 & 0.7374 & 0.7978 \\
        & RSAM-Seg & \textbf{0.8375} & \textbf{0.8724} & \textbf{0.9689} & \textbf{0.8744} & \textbf{0.8759} \\
        & GeoSAM-Lite & \underline{0.7974} & \underline{0.8453} & \underline{0.9535} & \underline{0.8470} & \underline{0.8510} \\
        \midrule
        \multirow{4}{*}{SPARCS} 
        & EfficientSAM & 0.6662 & 0.7794 & 0.9141 & 0.7863 & 0.7816 \\
        & MobileSAM & 0.6248 & 0.7570 & 0.8833 & 0.7685 & 0.7562 \\
        & RSAM-Seg & \textbf{0.7820} & \textbf{0.8596} & \textbf{0.9721} & \textbf{0.8666} & \textbf{0.8535} \\
        & GeoSAM-Lite & \underline{0.7127} & \underline{0.8024} & \underline{0.9603} & \underline{0.7903} & \underline{0.8195} \\ 
        \bottomrule
        \multicolumn{5}{l}{\textit{Note: Best results are in bold; second best are underlined.}}
    \end{tabular}
\end{table}

\begin{figure}[!t]
    \centering
    \newcommand{\smallimg}[1]{\includegraphics[width=0.145\columnwidth]{#1}}
    \setlength{\tabcolsep}{1.2pt} 
    \renewcommand{\arraystretch}{0.6} 
    \scriptsize 

    \begin{tabular}{c cccccc}
        & \textbf{(a)} & \textbf{(b)} & \textbf{(c)}  & \textbf{(d)} & \textbf{(e)} & \textbf{(f)}\\
        
        \multirow{3}{*}{\rotatebox{90}{\textbf{38-Cloud}}} 
        & \smallimg{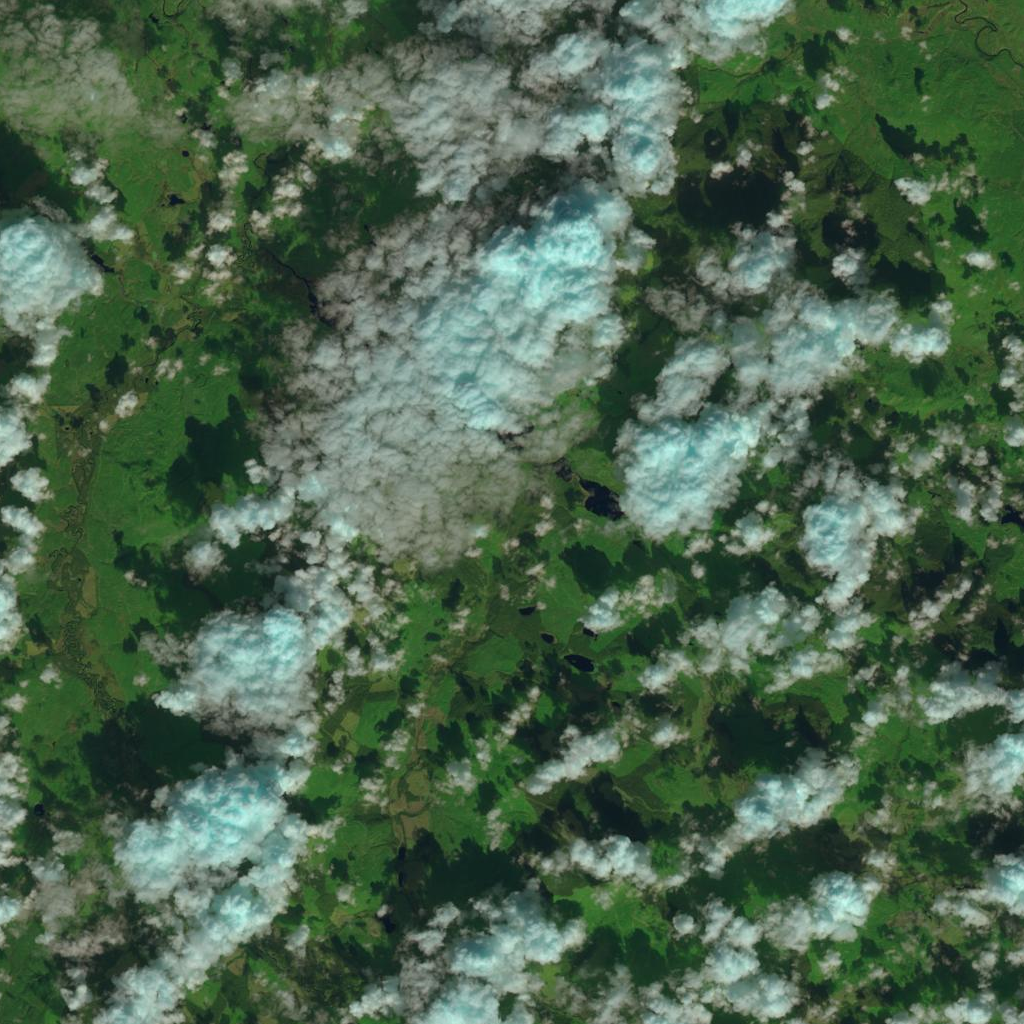} & \smallimg{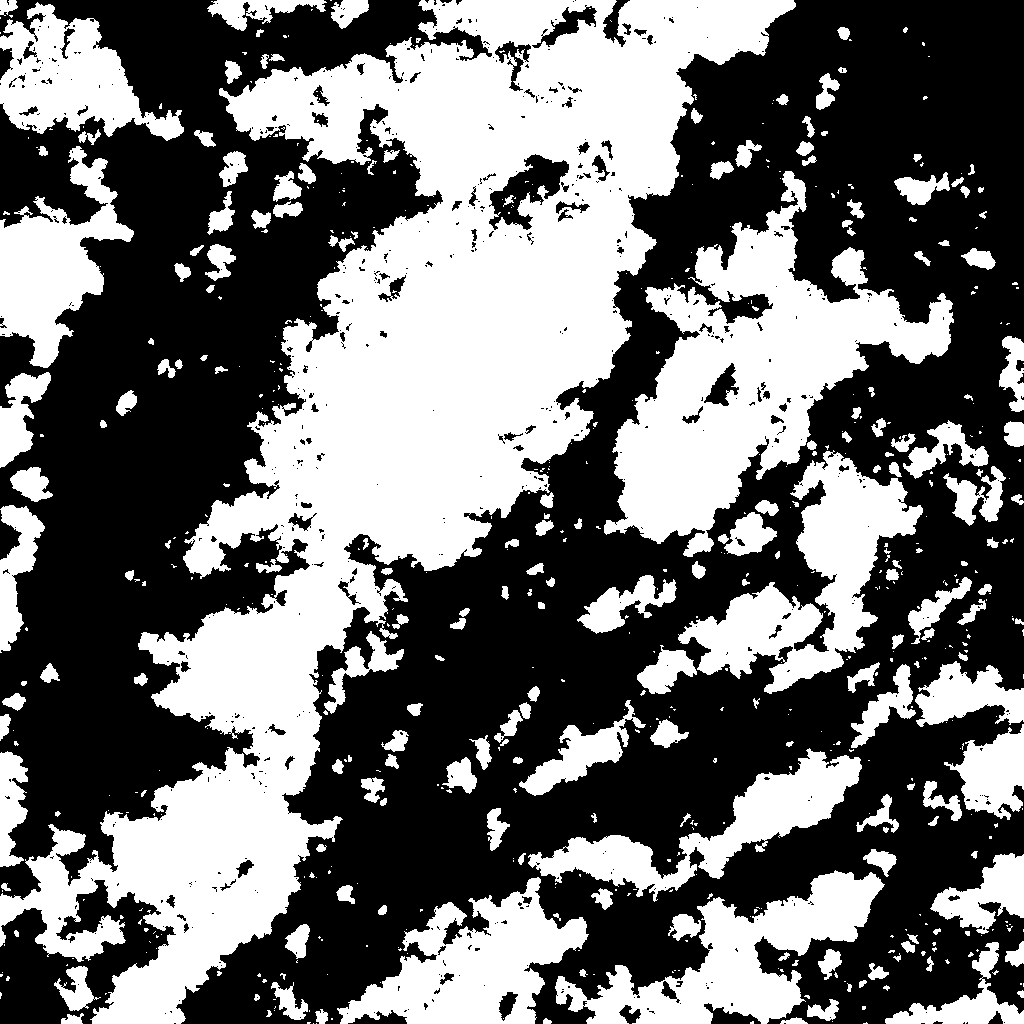} & \smallimg{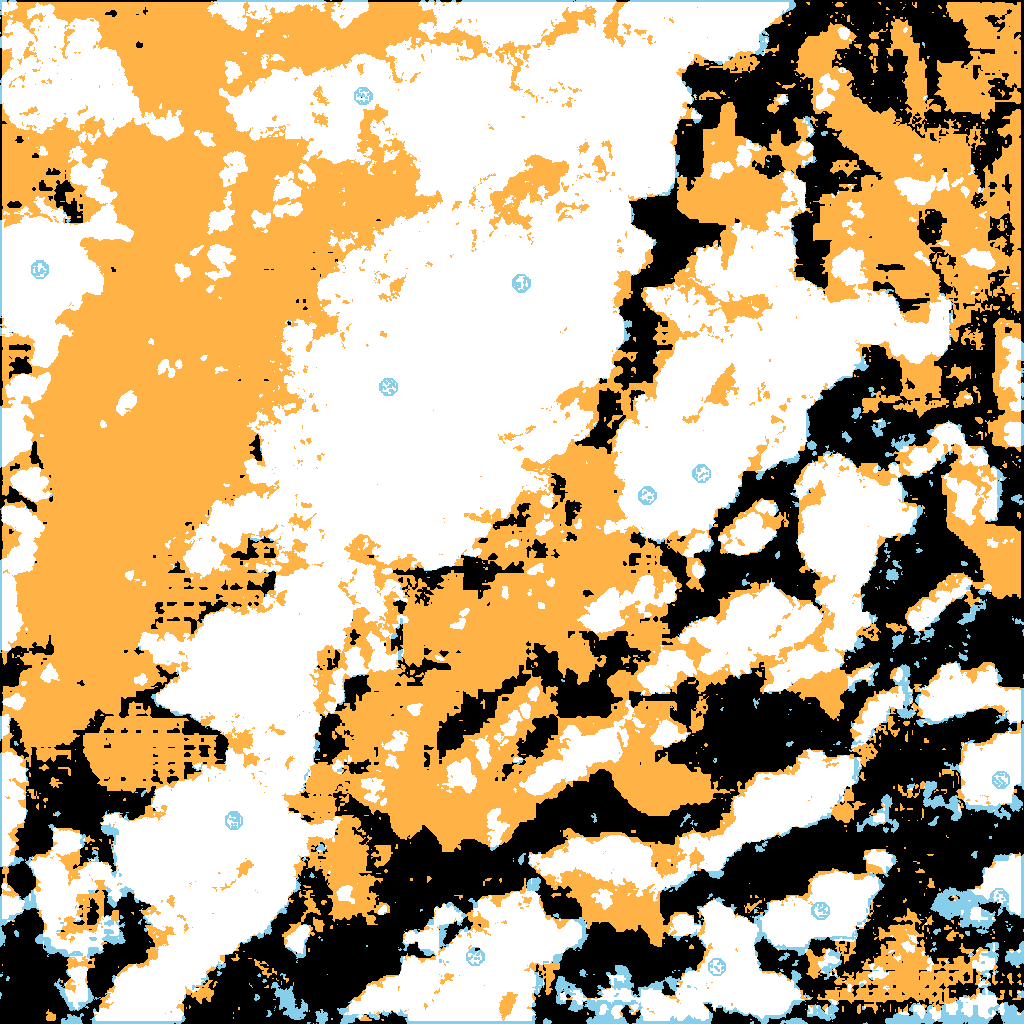} & \smallimg{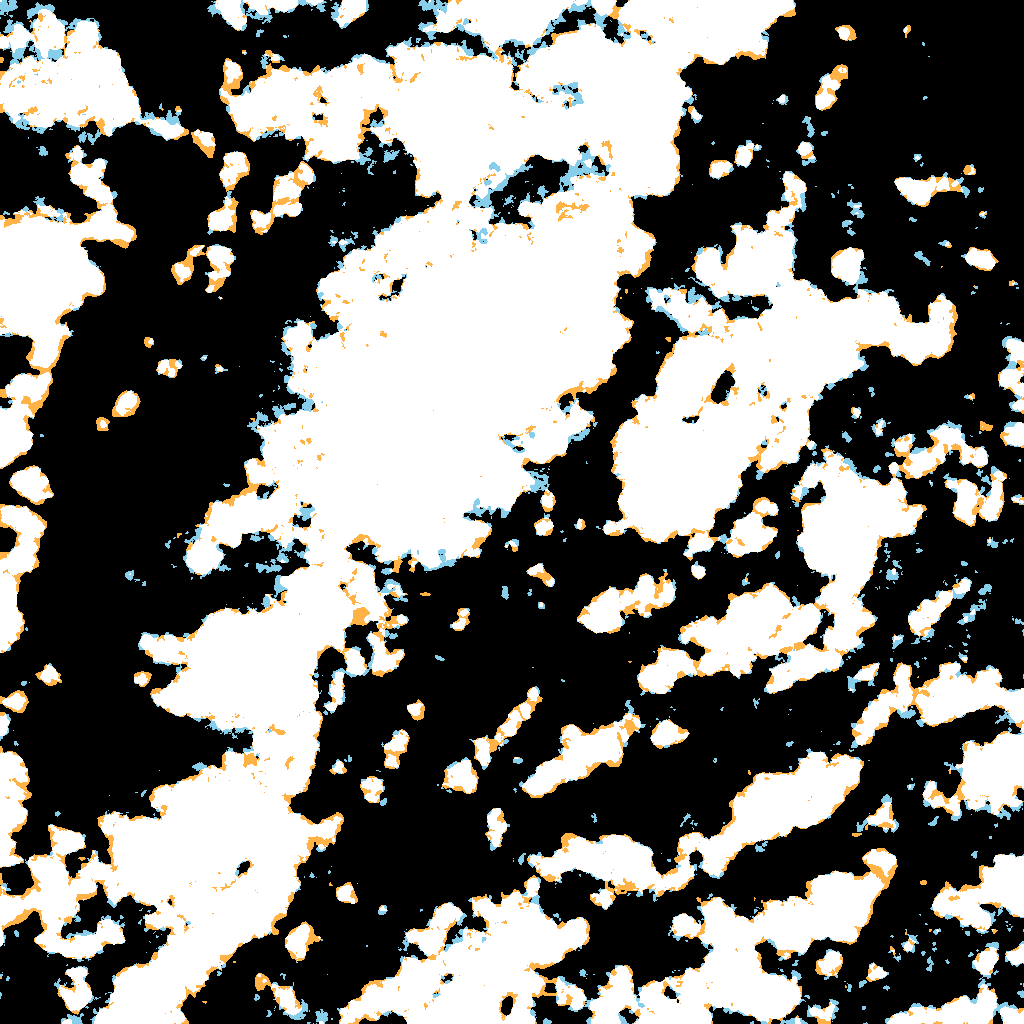} & \smallimg{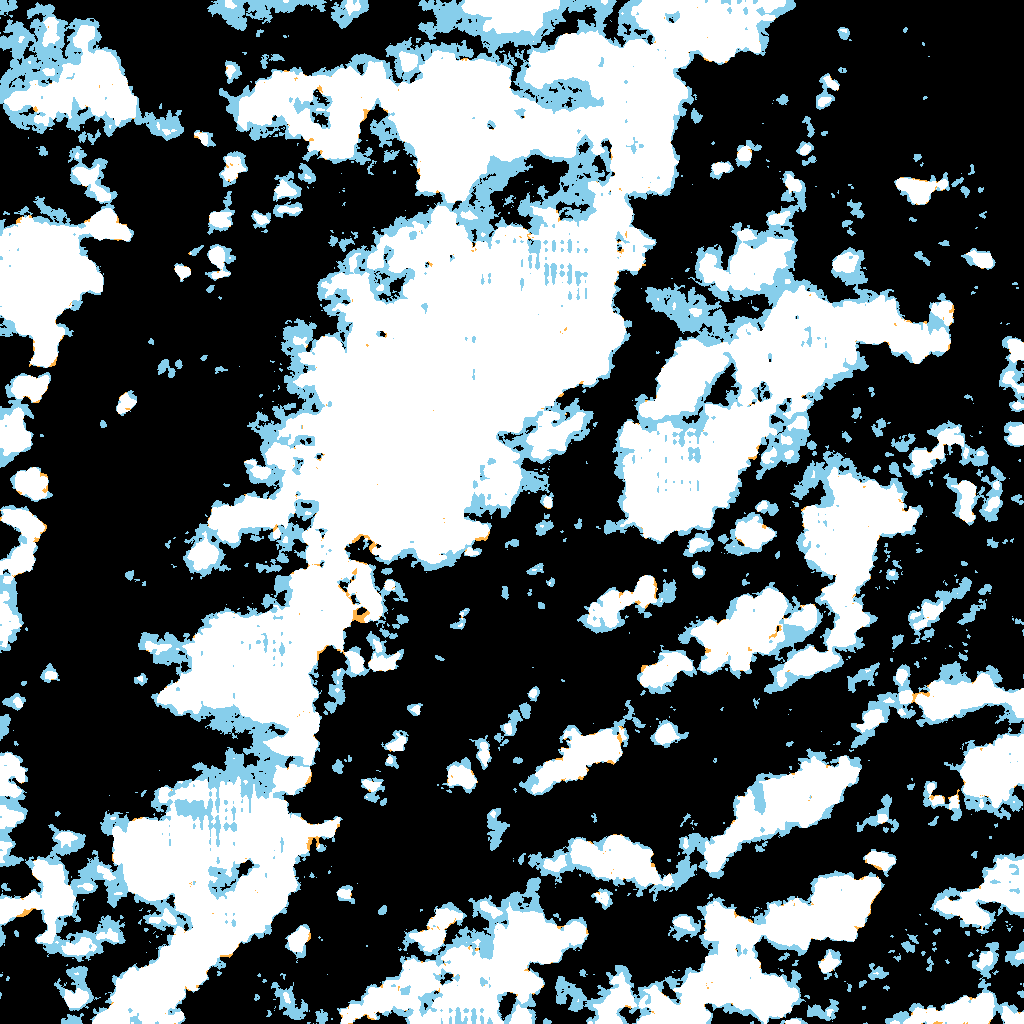} &
        \smallimg{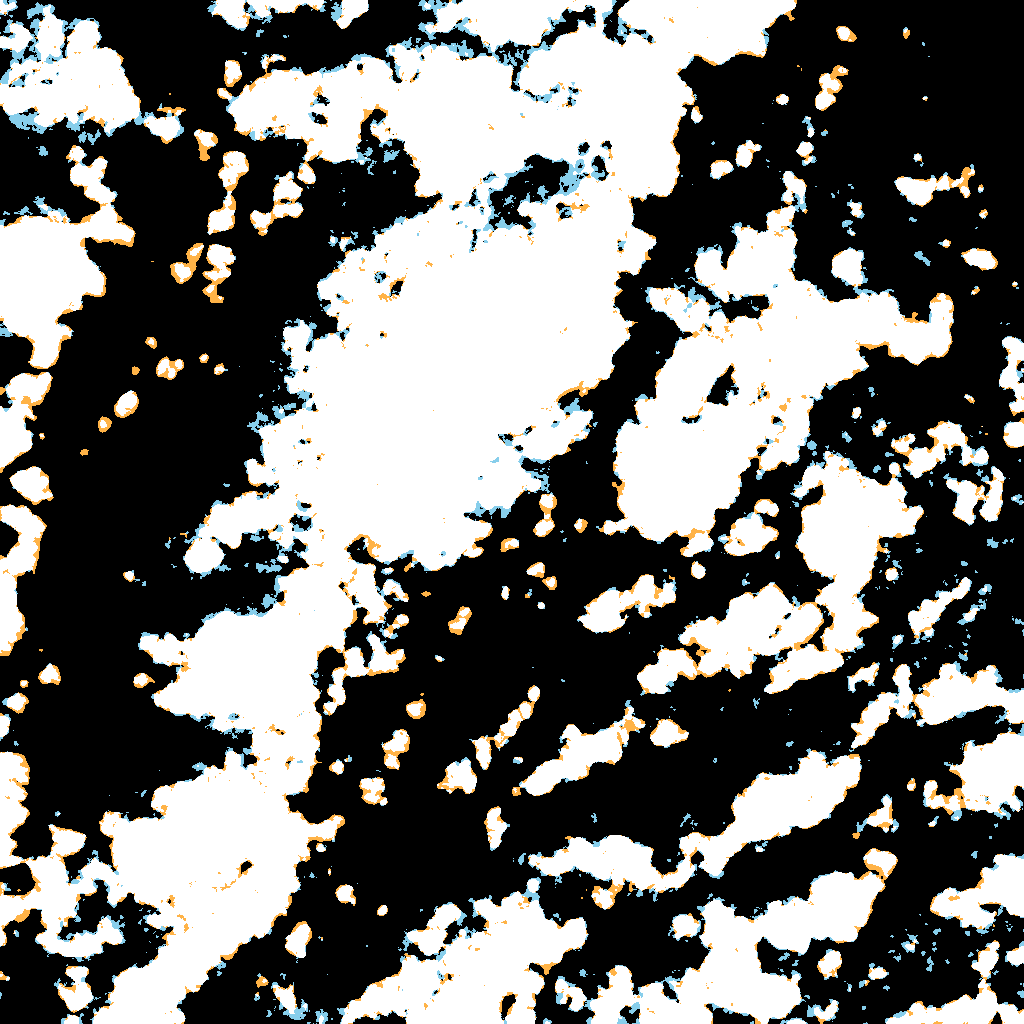}  \\
        & \smallimg{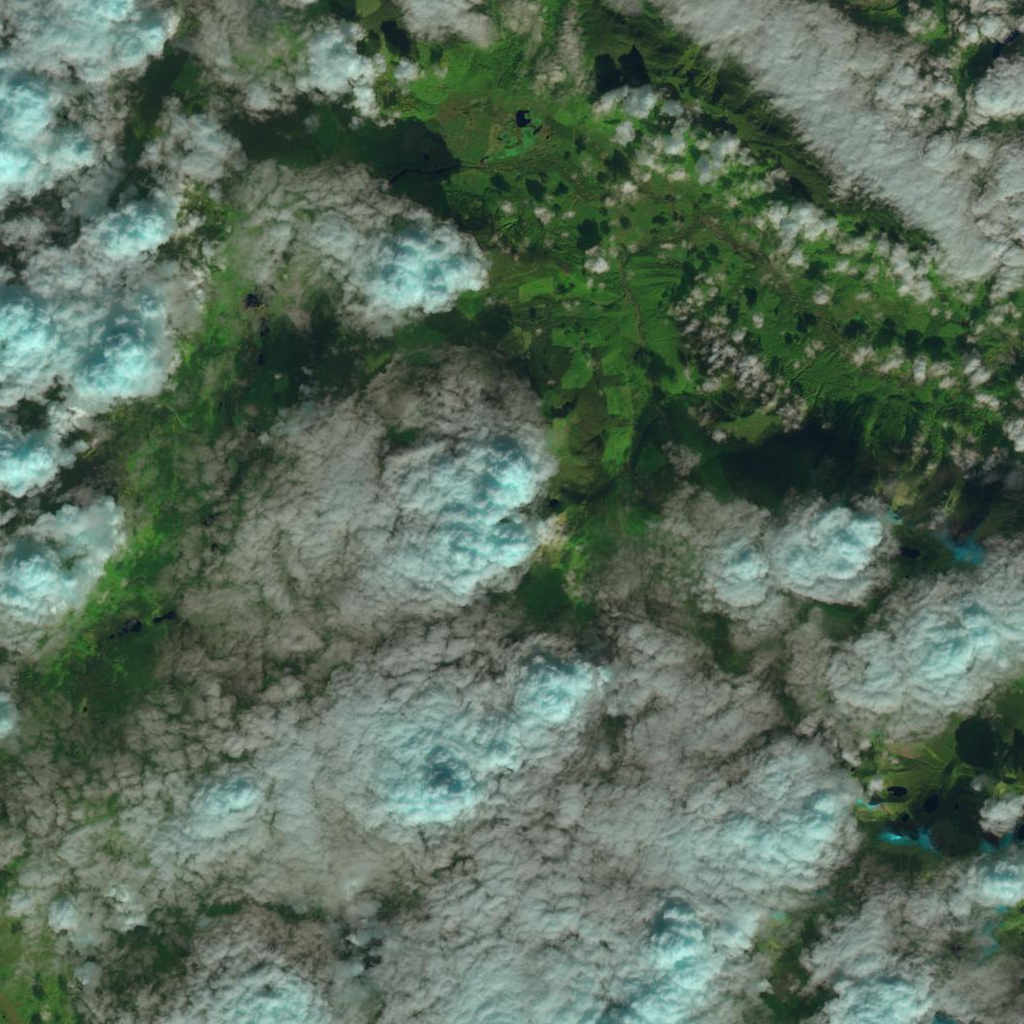} & \smallimg{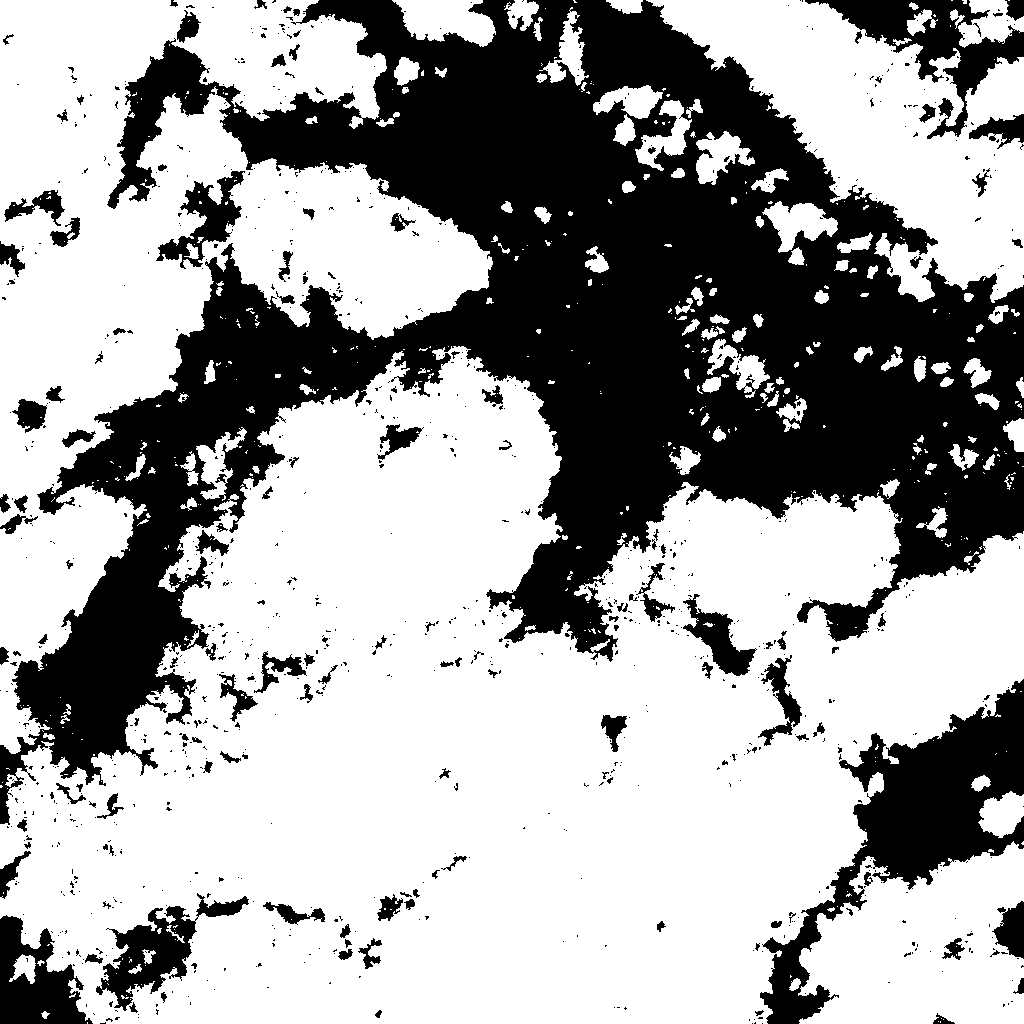} & \smallimg{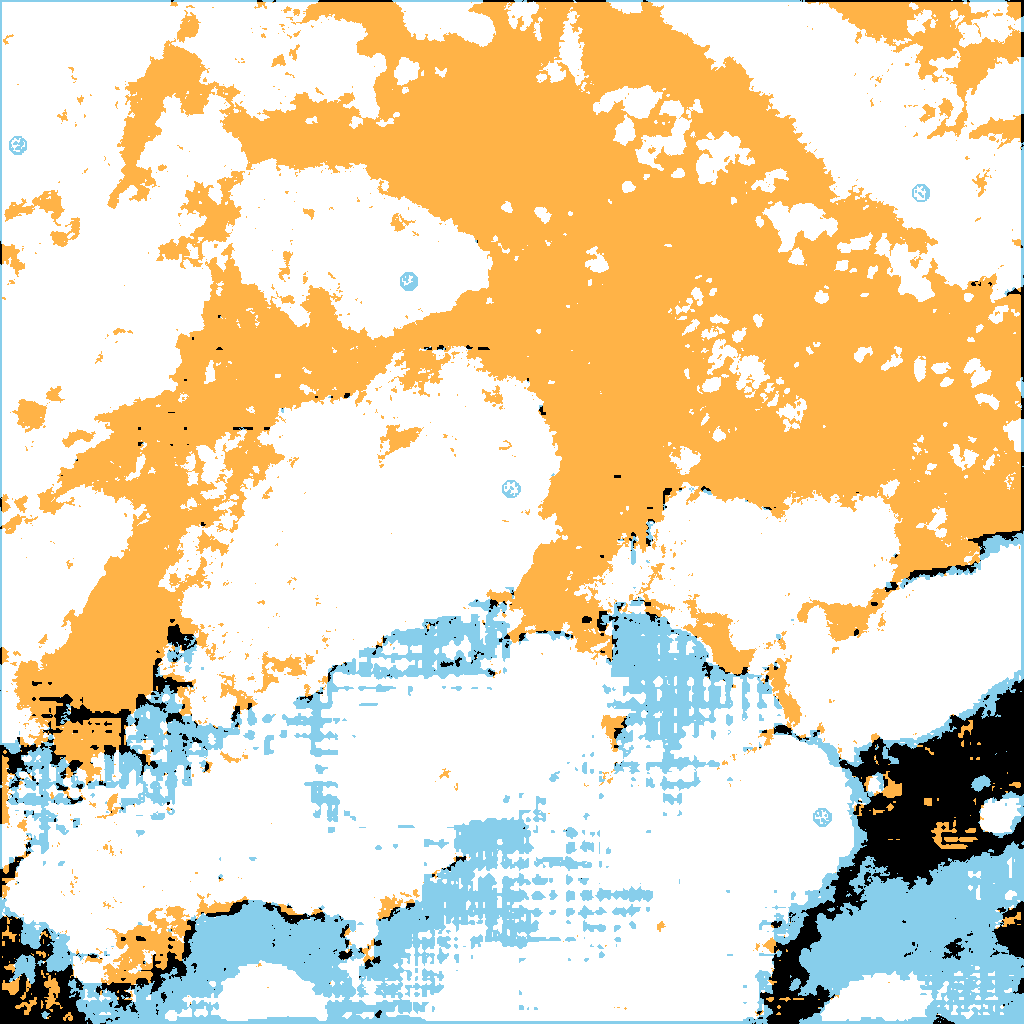} & \smallimg{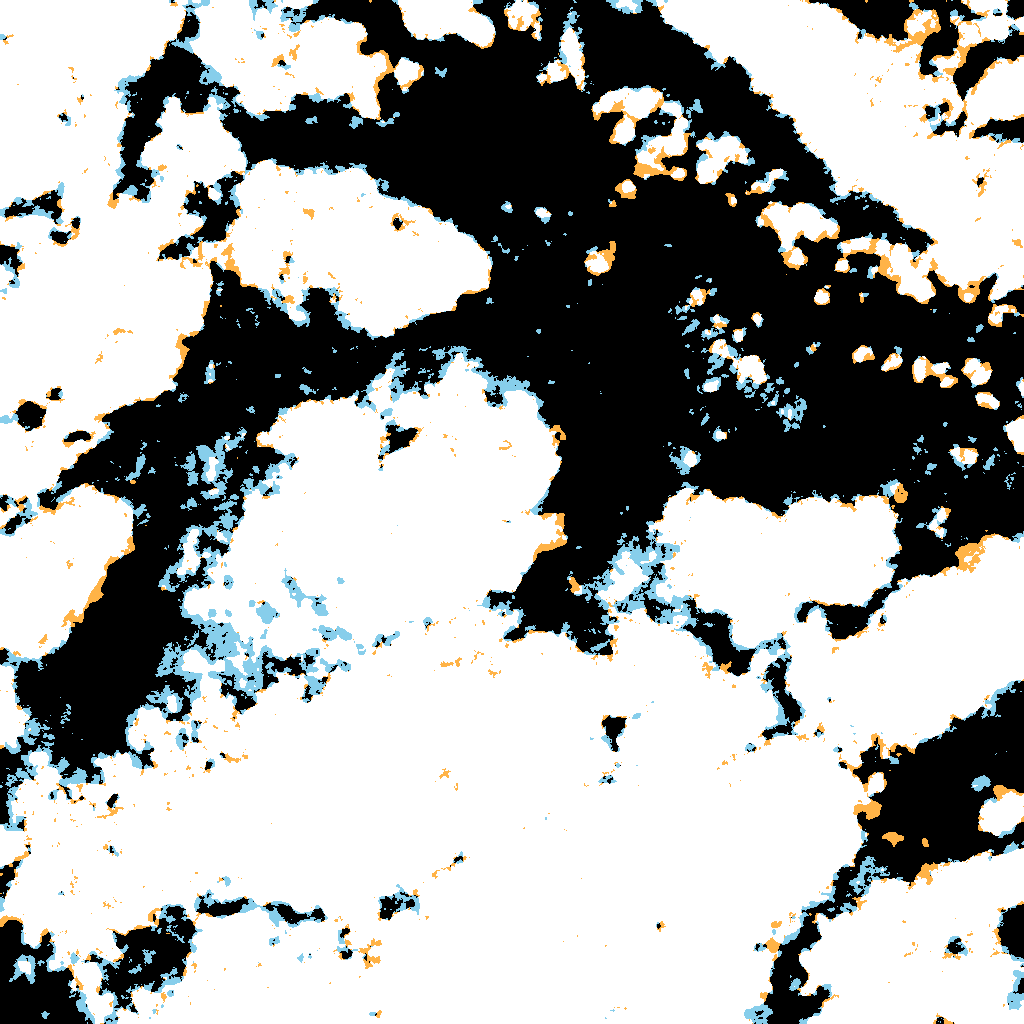} & \smallimg{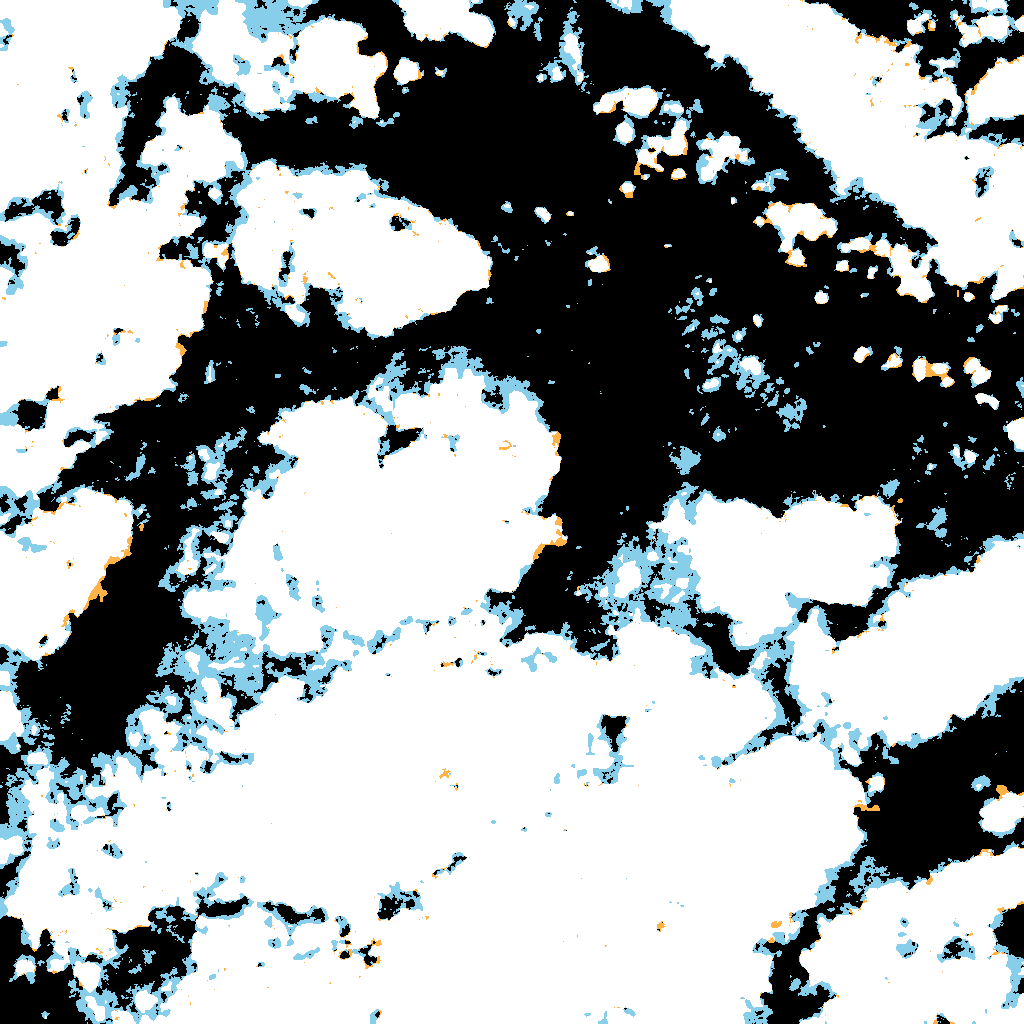} &
        \smallimg{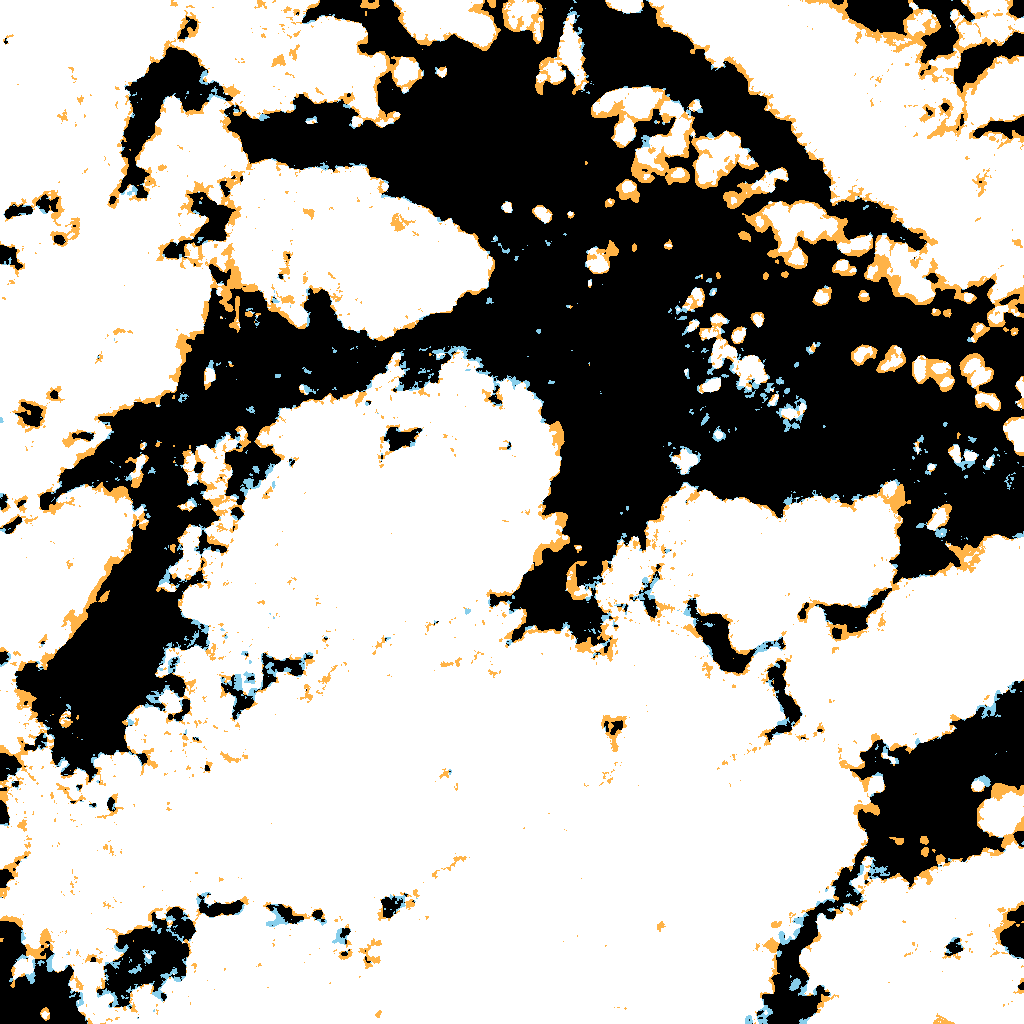} \\
        & \smallimg{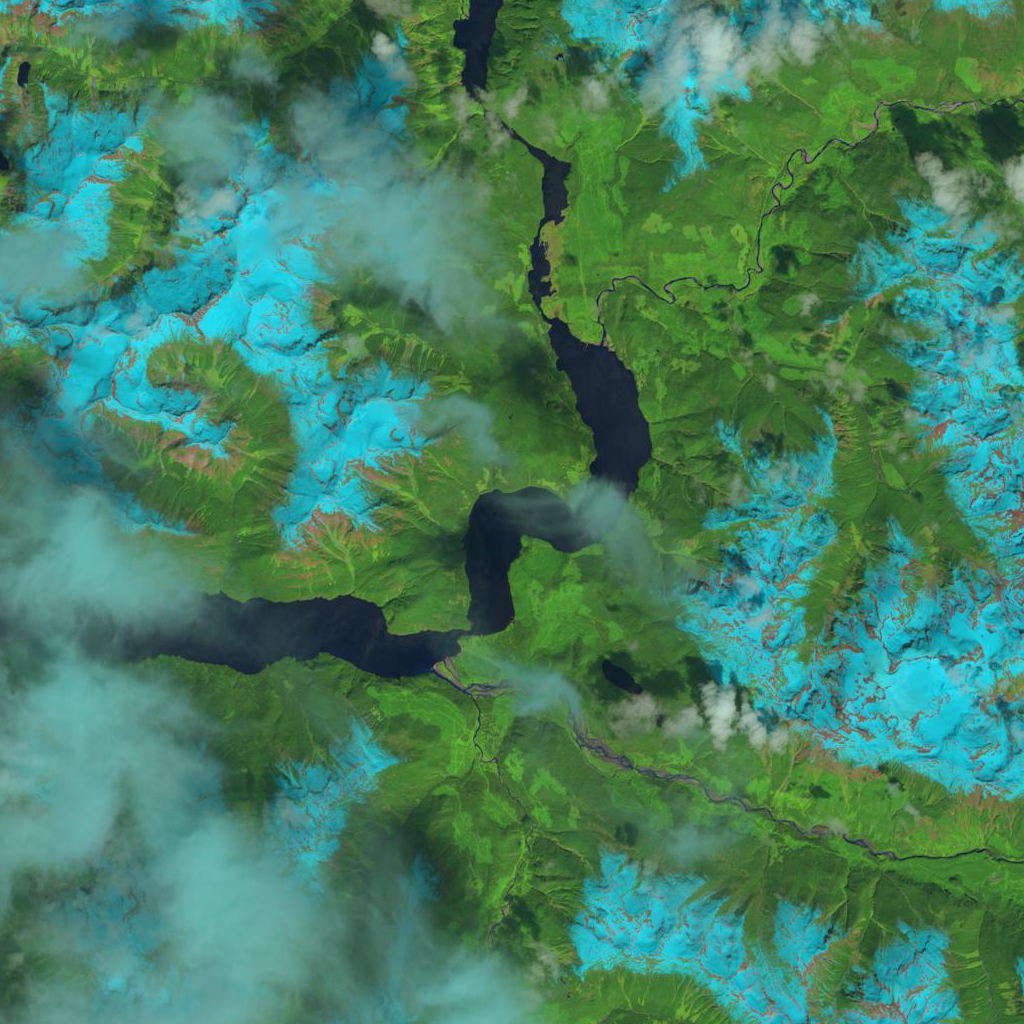} & \smallimg{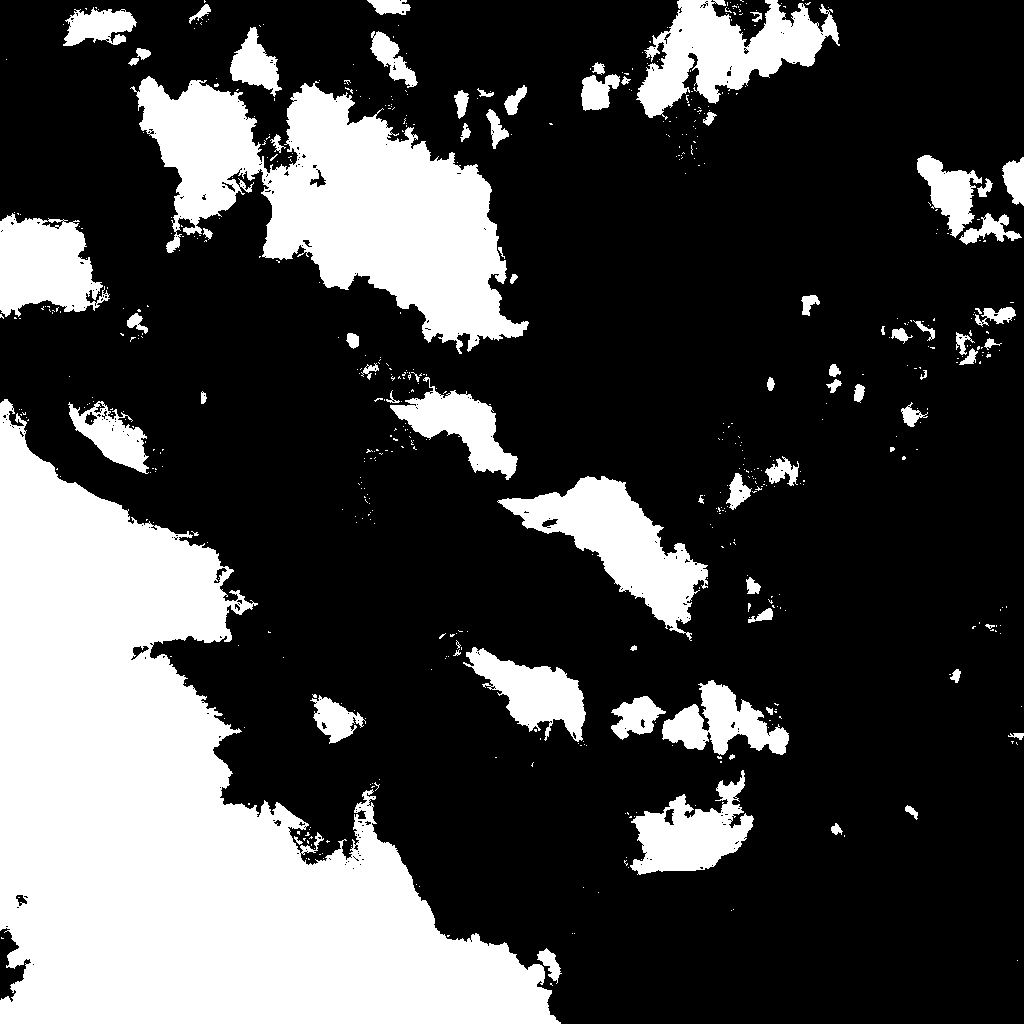} & \smallimg{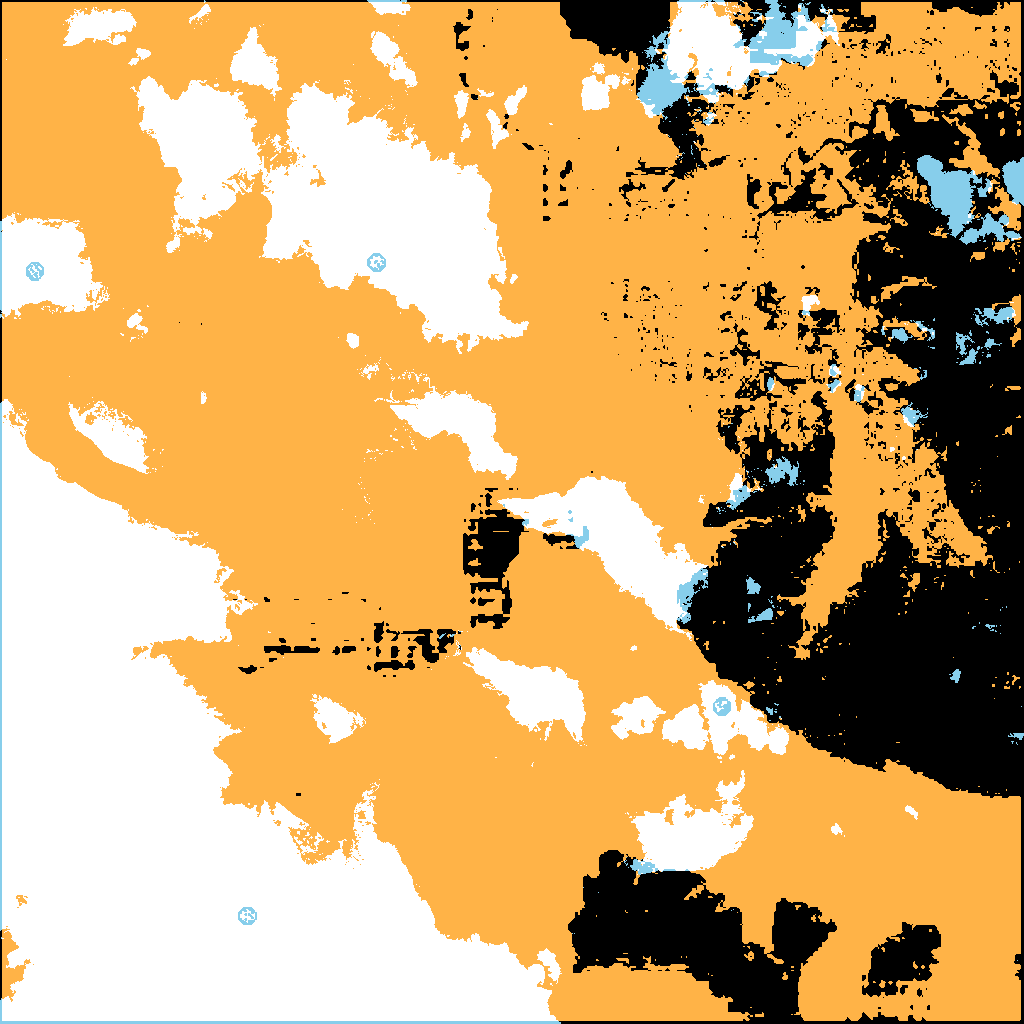} & \smallimg{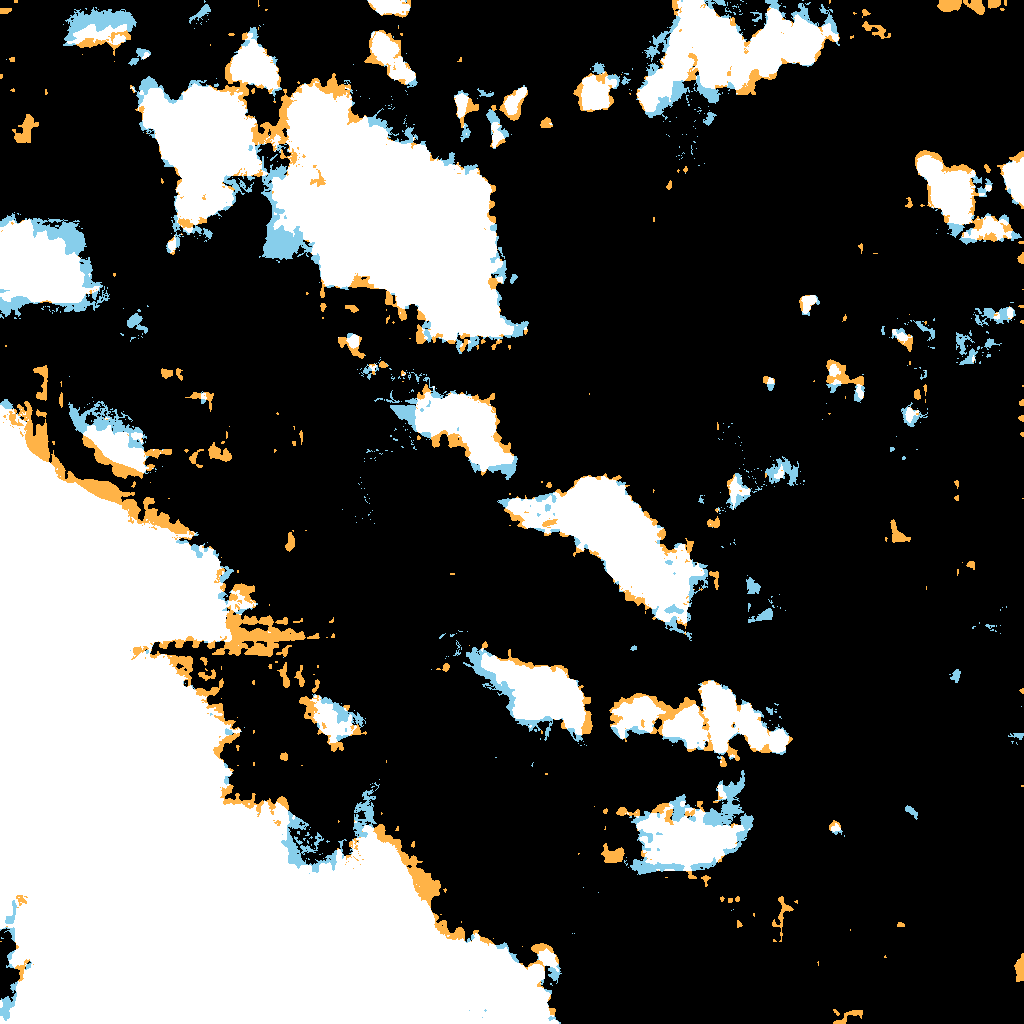} & \smallimg{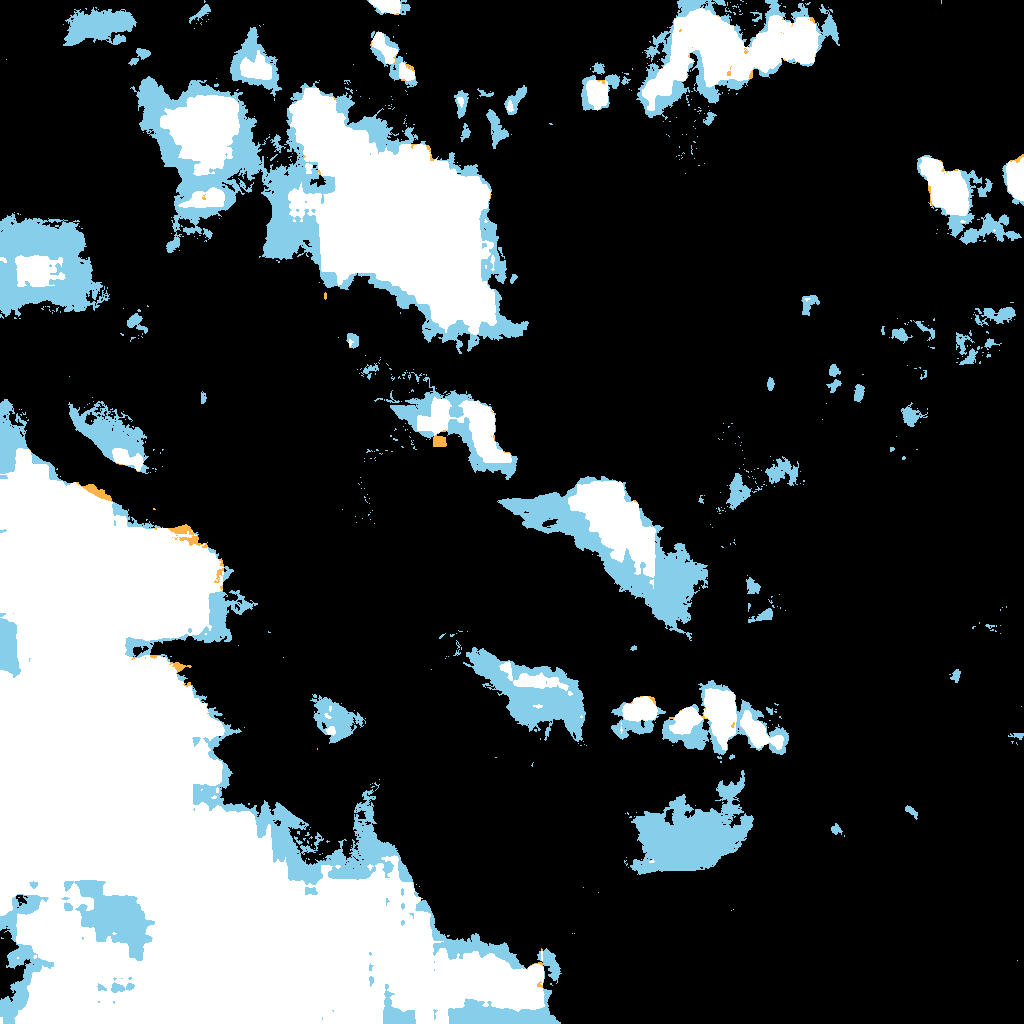} &
        \smallimg{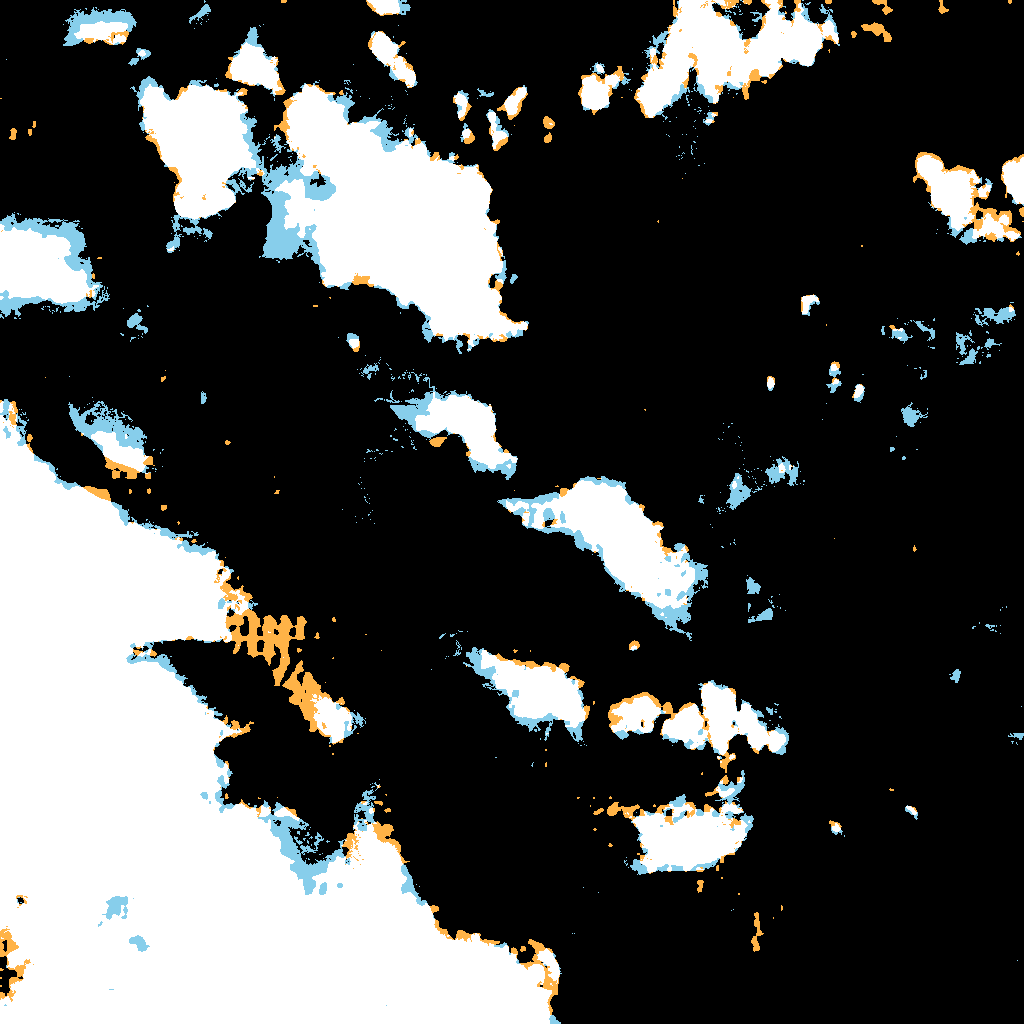} \\
        
        \noalign{\vspace{2pt}}

        \multirow{3}{*}{\rotatebox{90}{\textbf{CloudSEN12}}} 
        & \smallimg{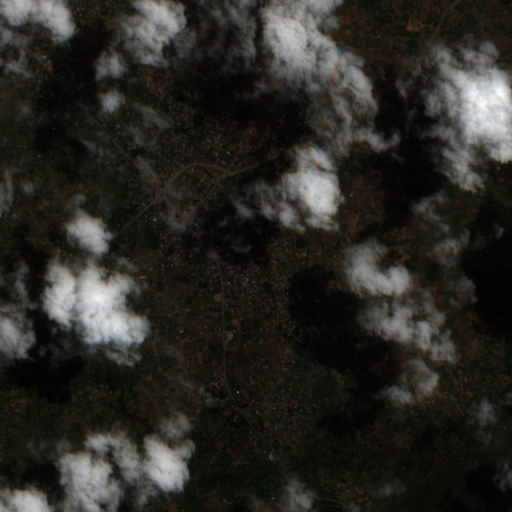} & \smallimg{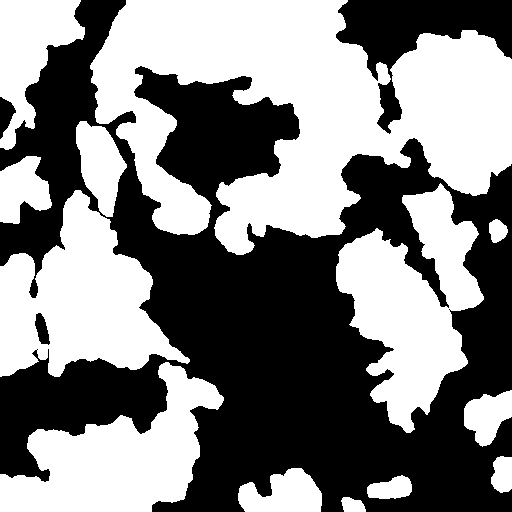} & \smallimg{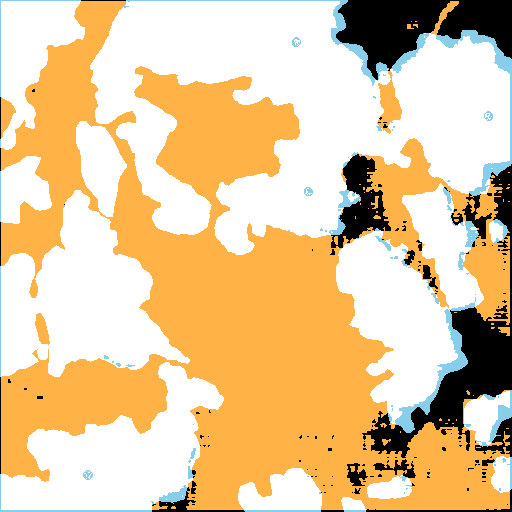} & \smallimg{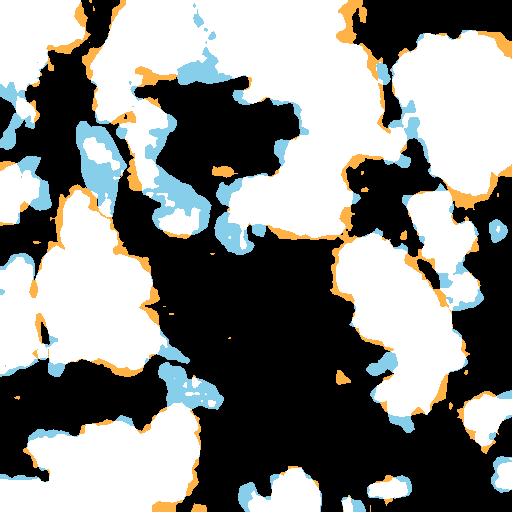} & \smallimg{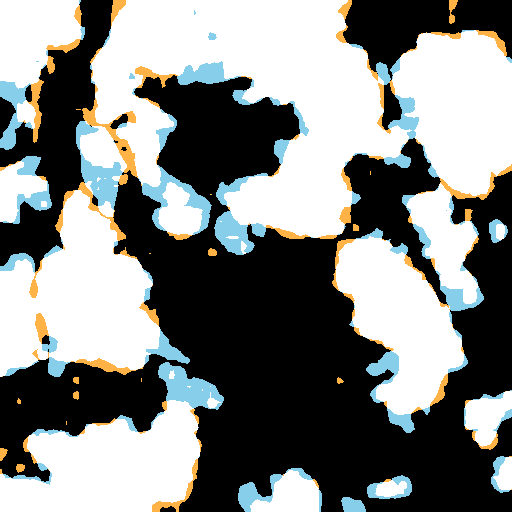} &
        \smallimg{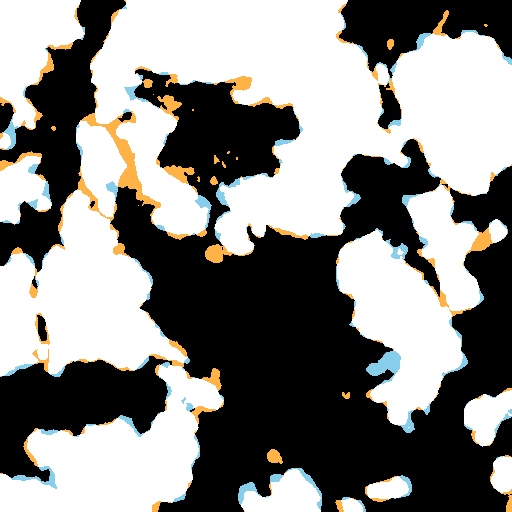} \\
        & \smallimg{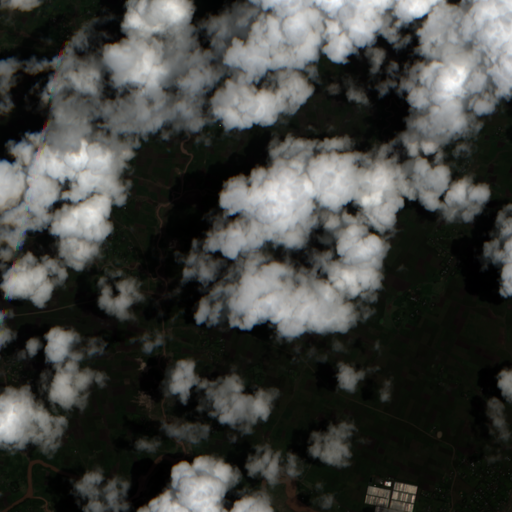} & \smallimg{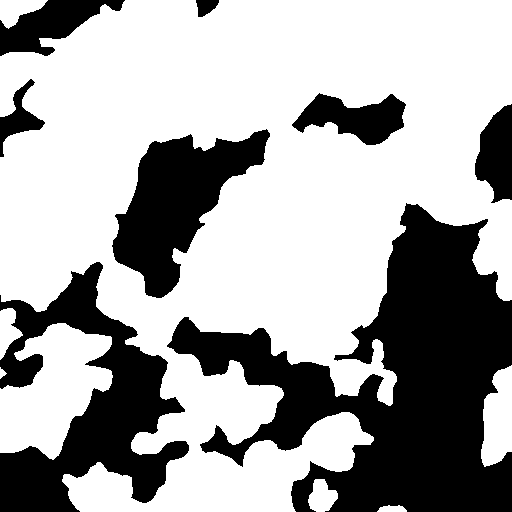} & \smallimg{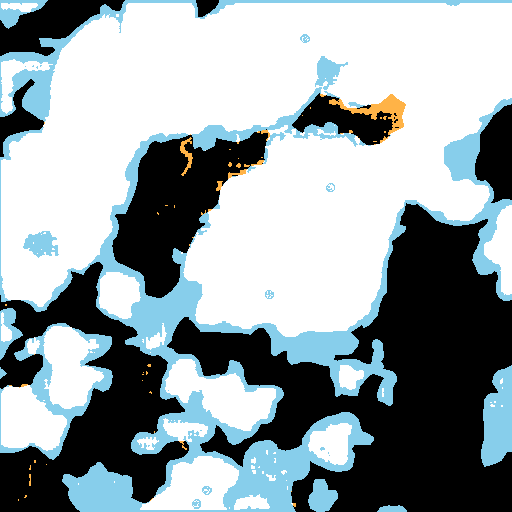} & \smallimg{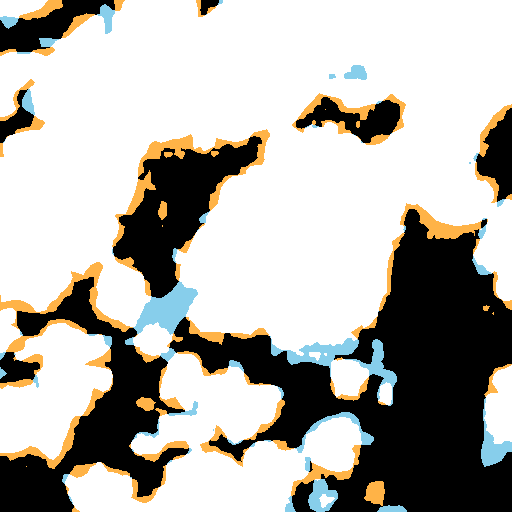} & \smallimg{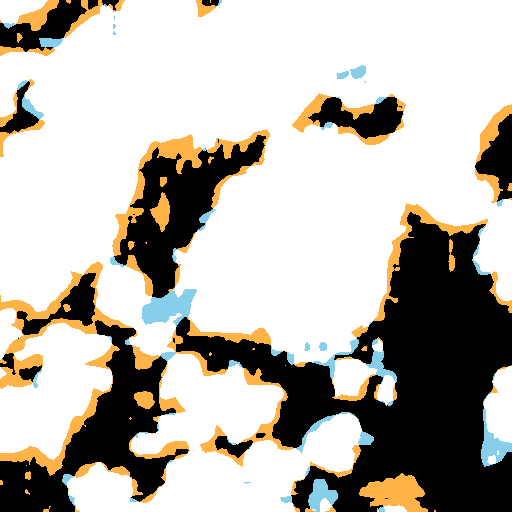} &
        \smallimg{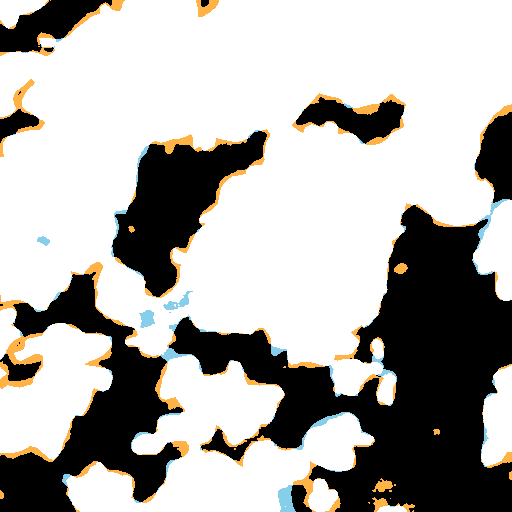} \\
        & \smallimg{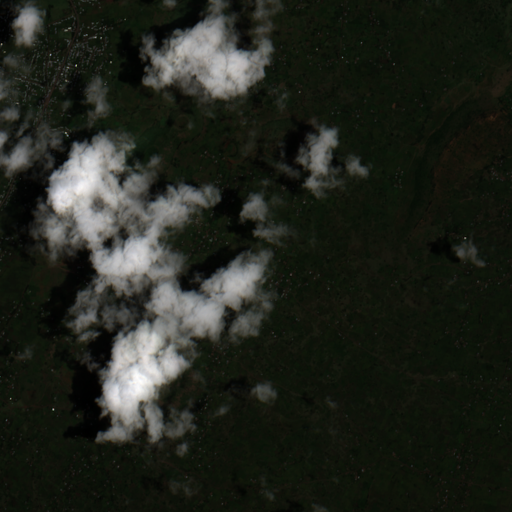} & \smallimg{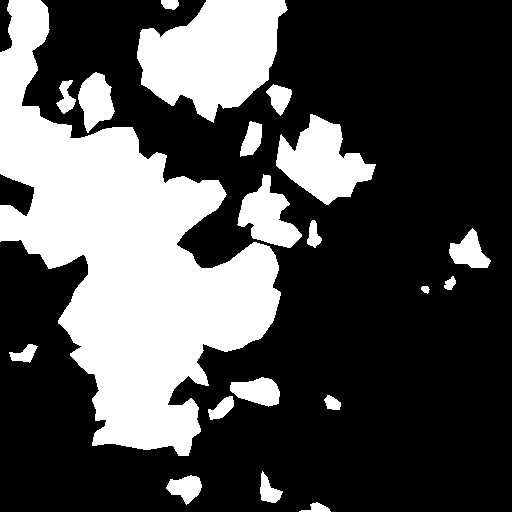} & \smallimg{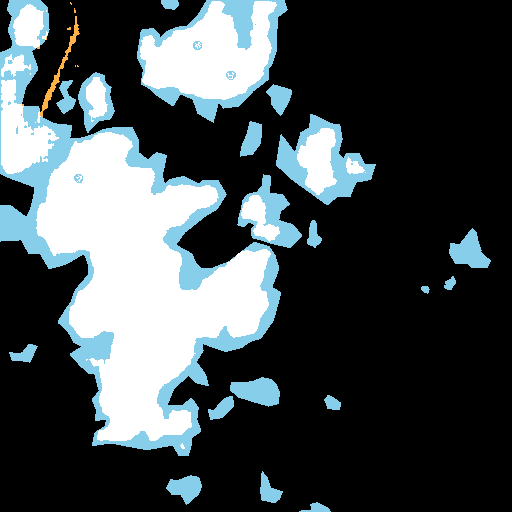} & \smallimg{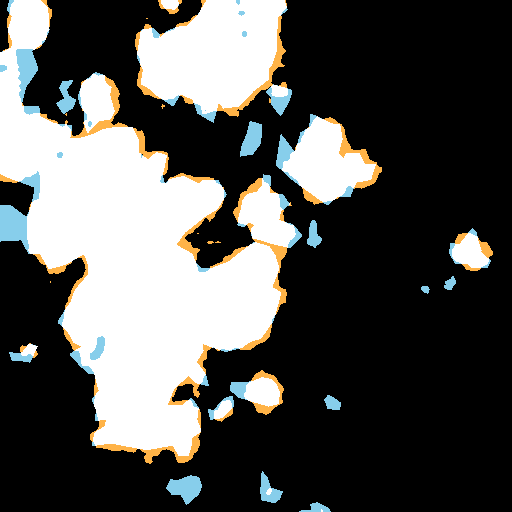} & \smallimg{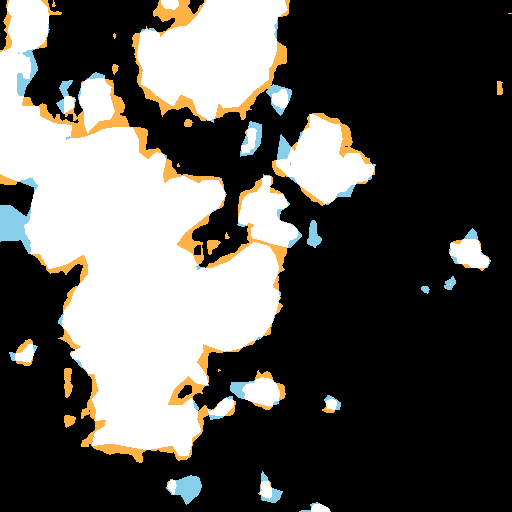} &
        \smallimg{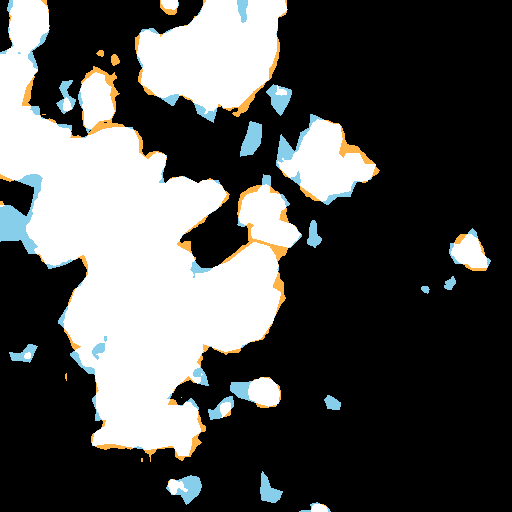} \\

        \noalign{\vspace{2pt}}
        \multirow{3}{*}{\rotatebox{90}{\textbf{SPARCS}}} 
        & \smallimg{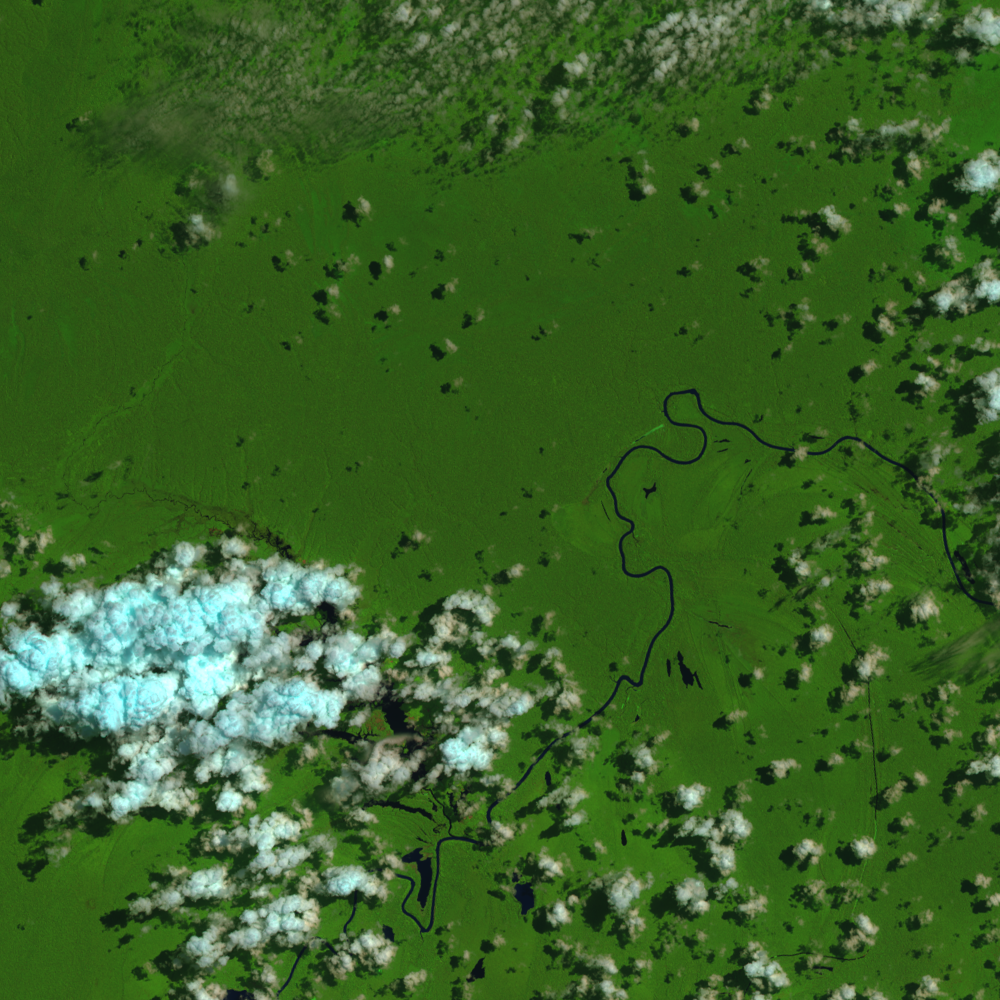} & \smallimg{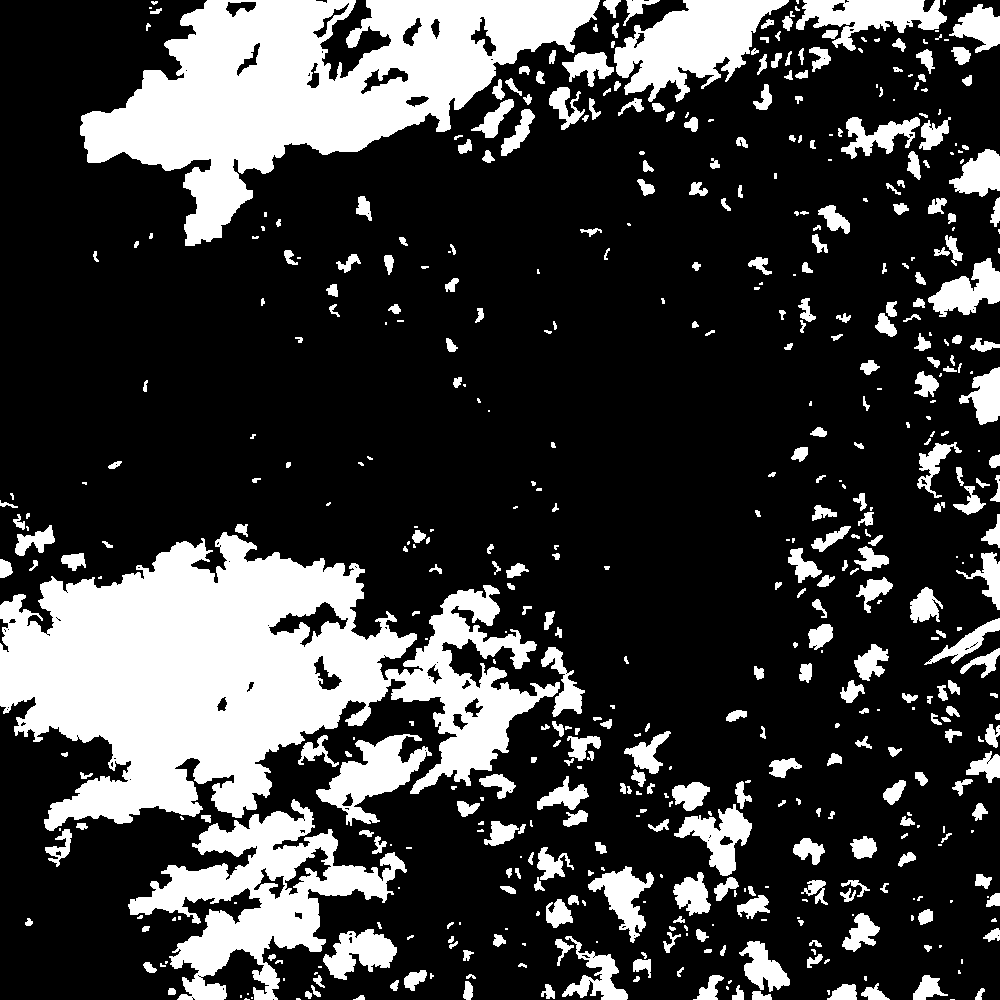} &  \smallimg{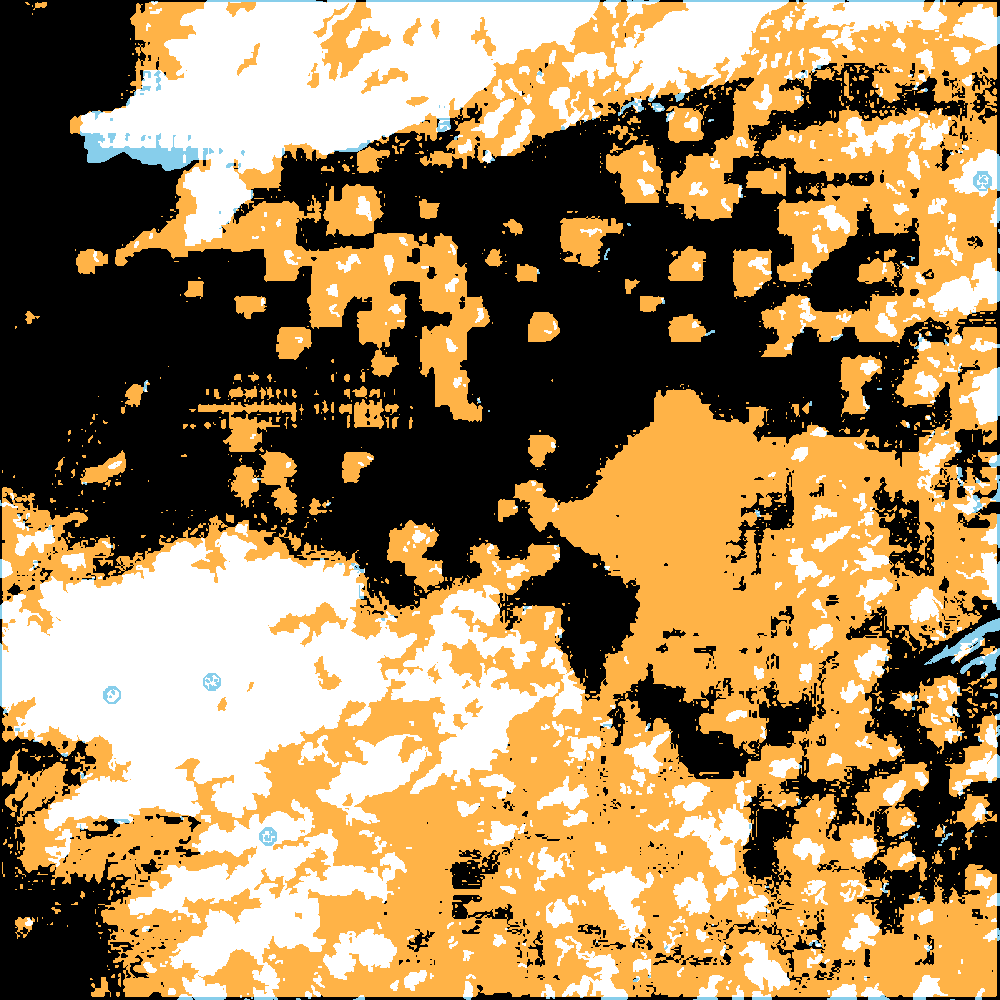} & \smallimg{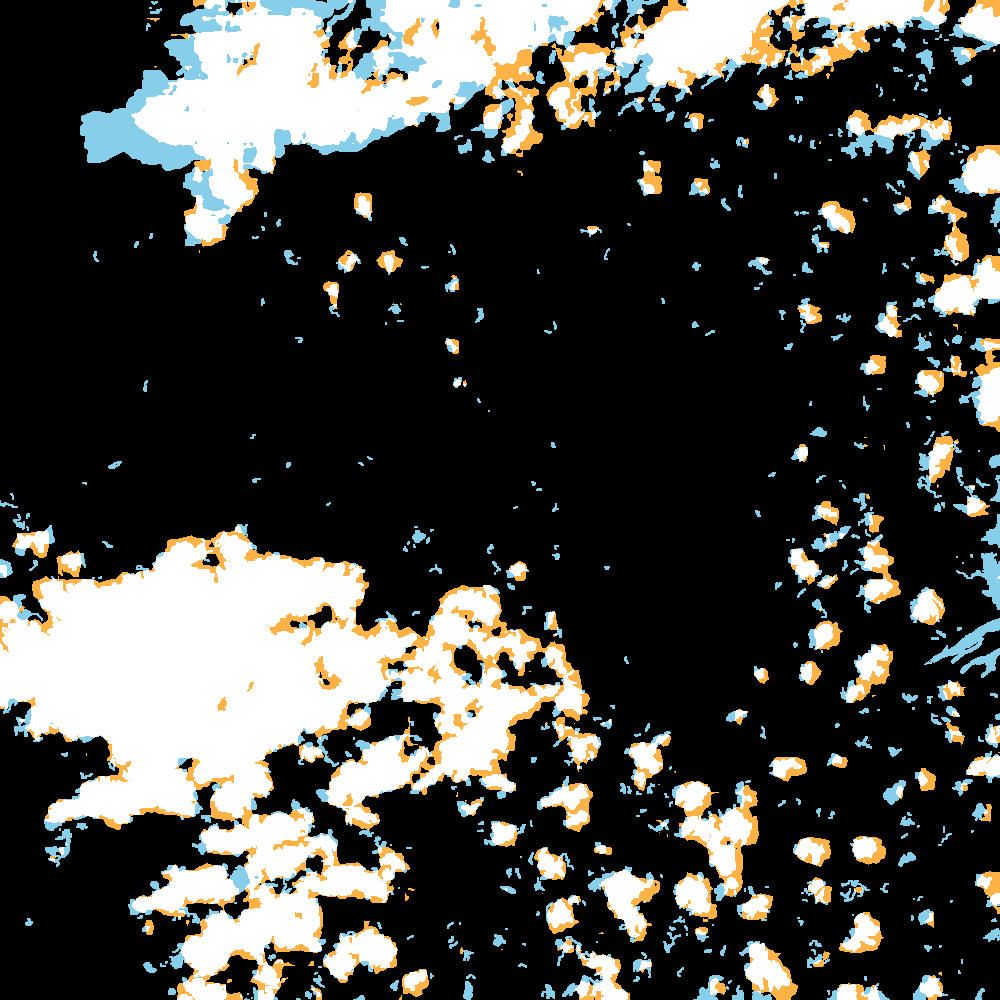} & \smallimg{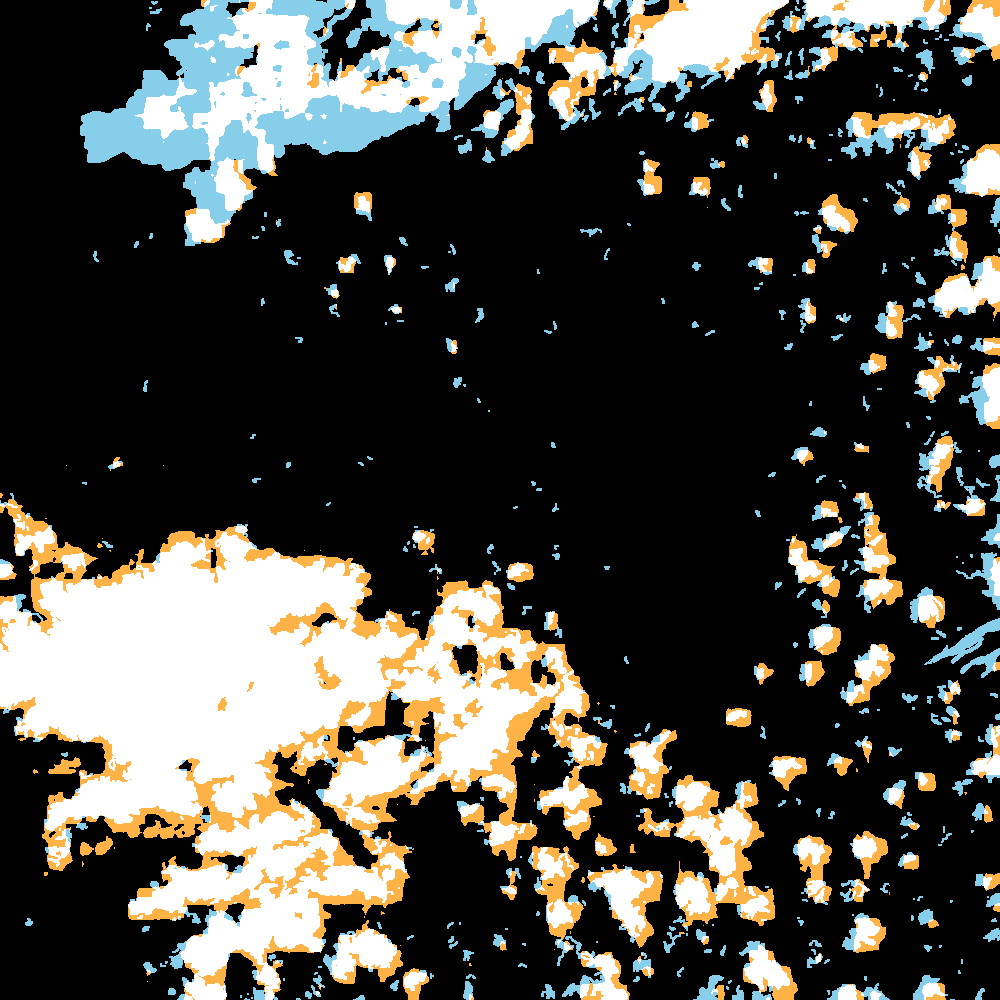} & 
        \smallimg{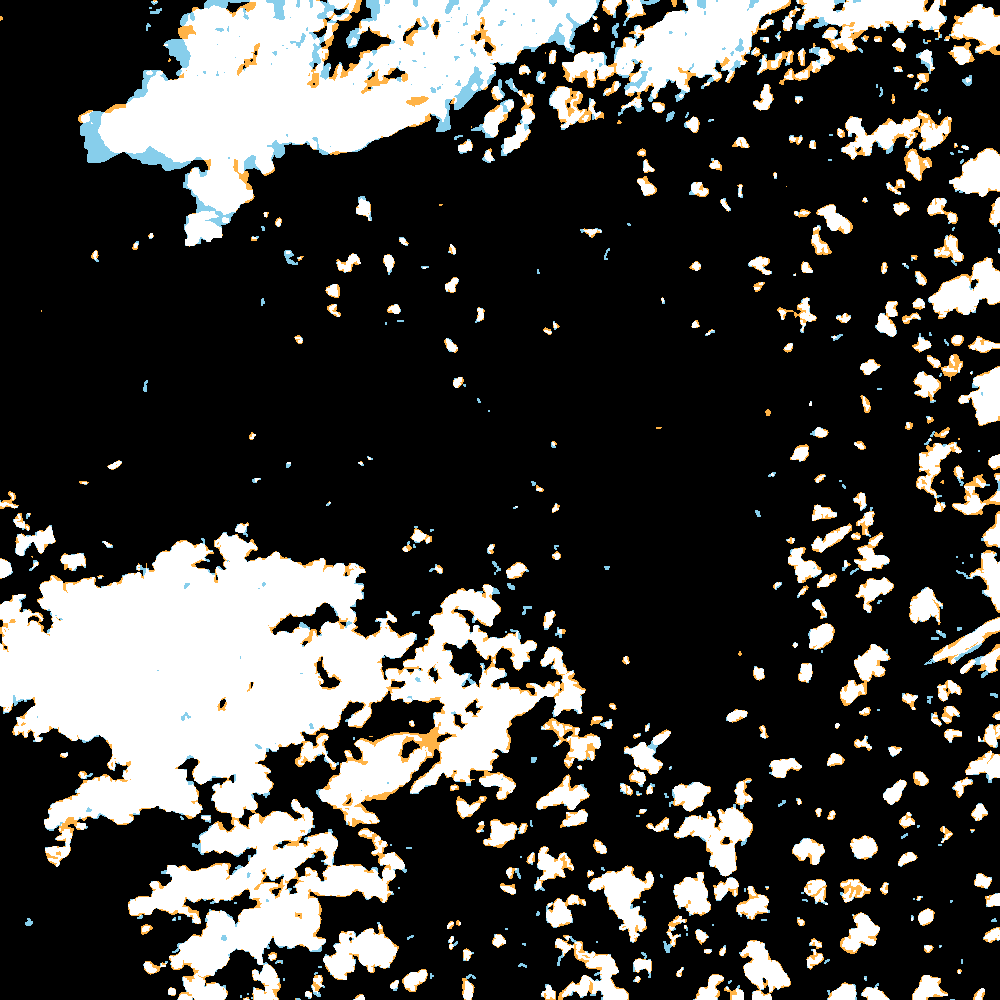} \\
        & \smallimg{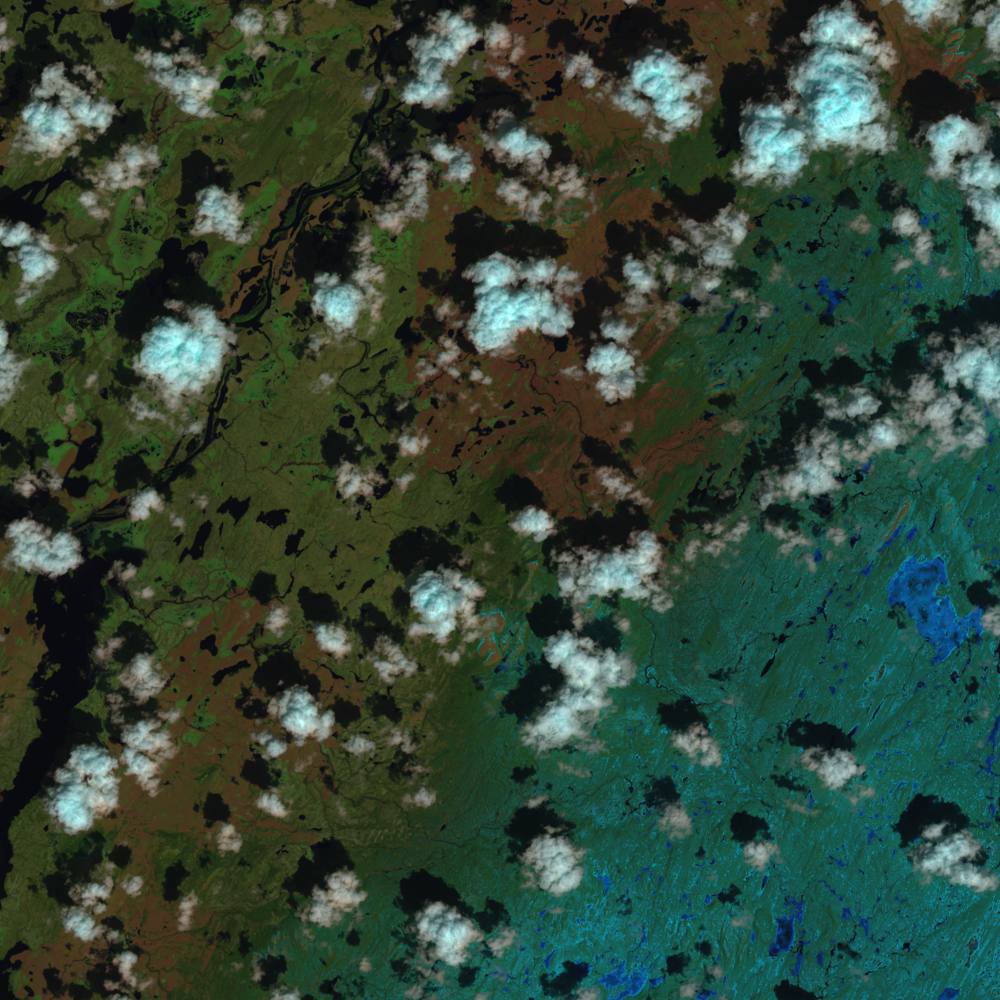} & \smallimg{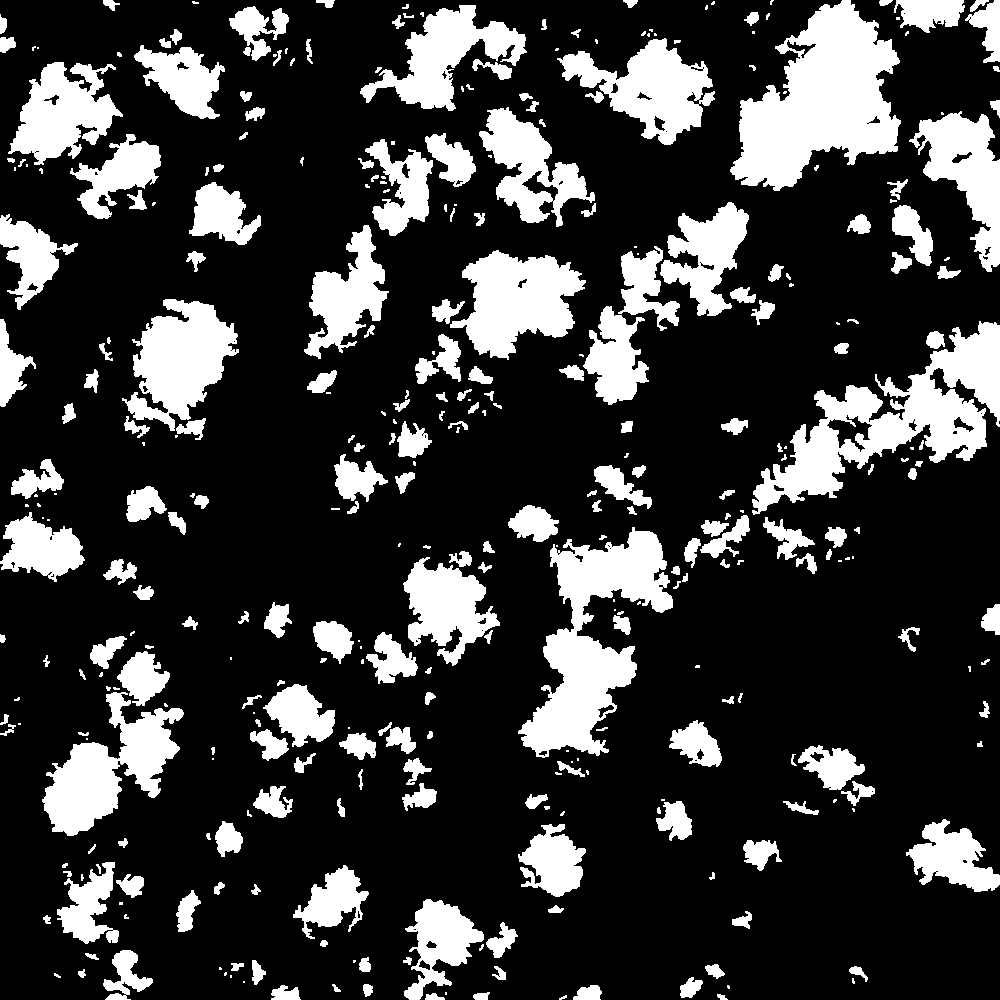} & \smallimg{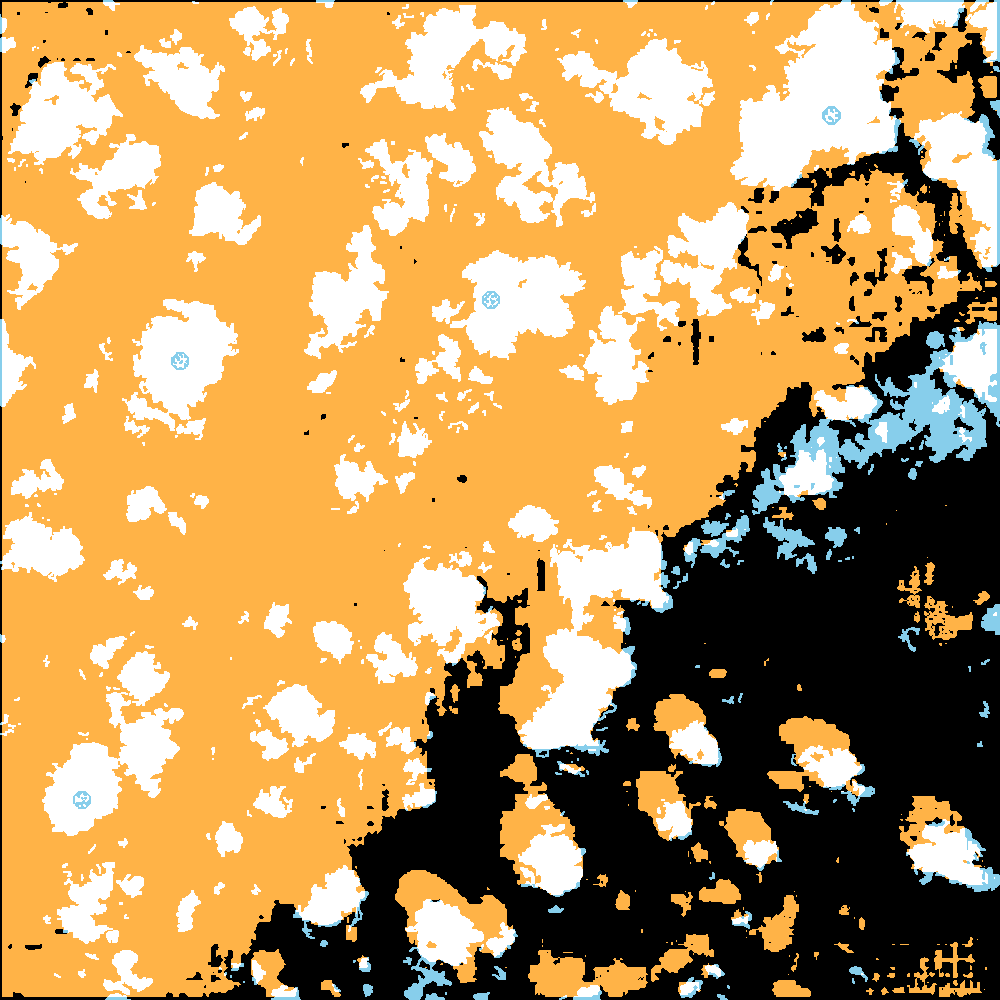} & \smallimg{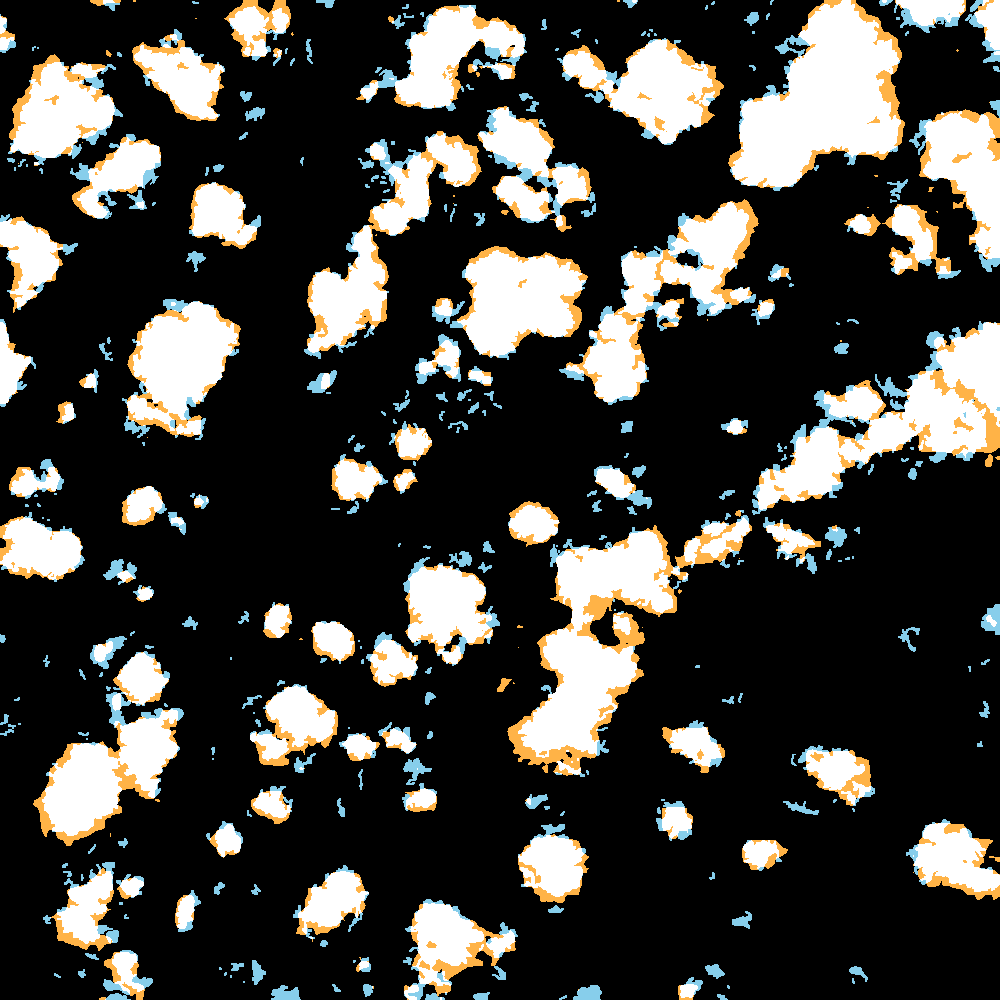} & \smallimg{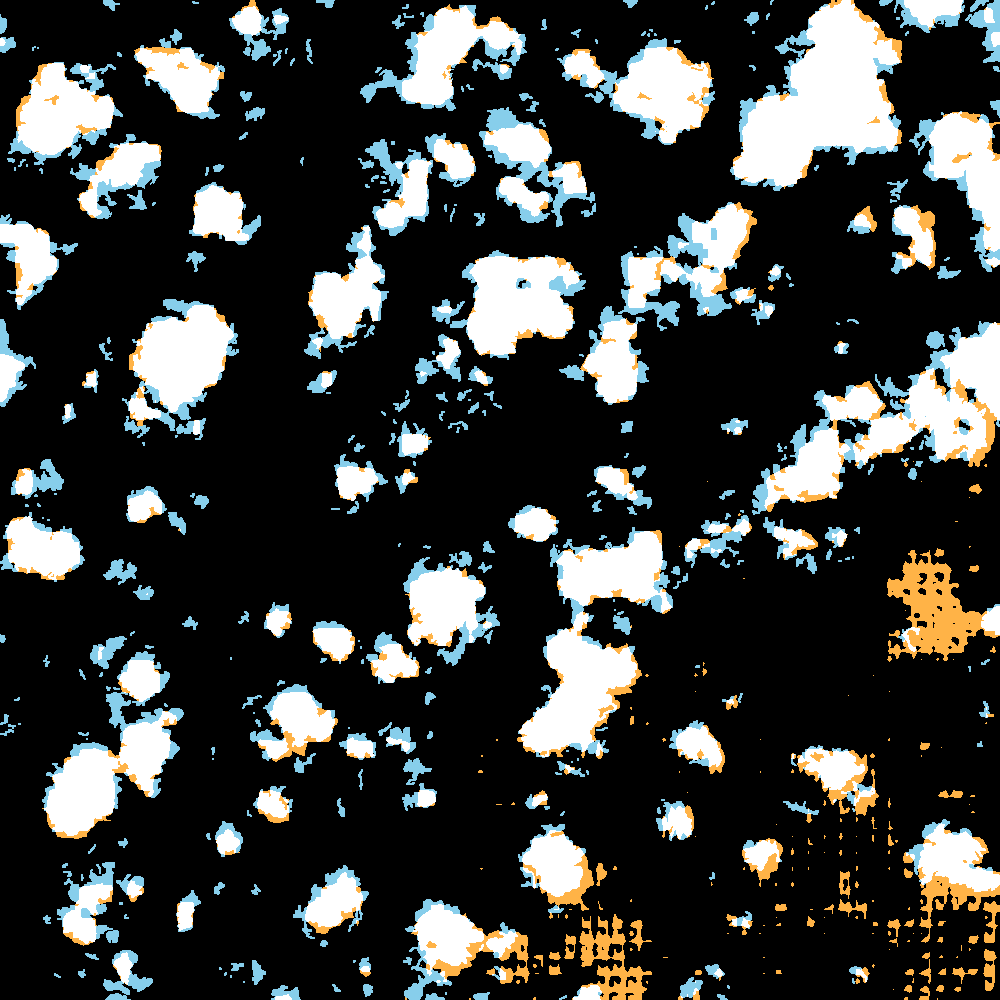} &
        \smallimg{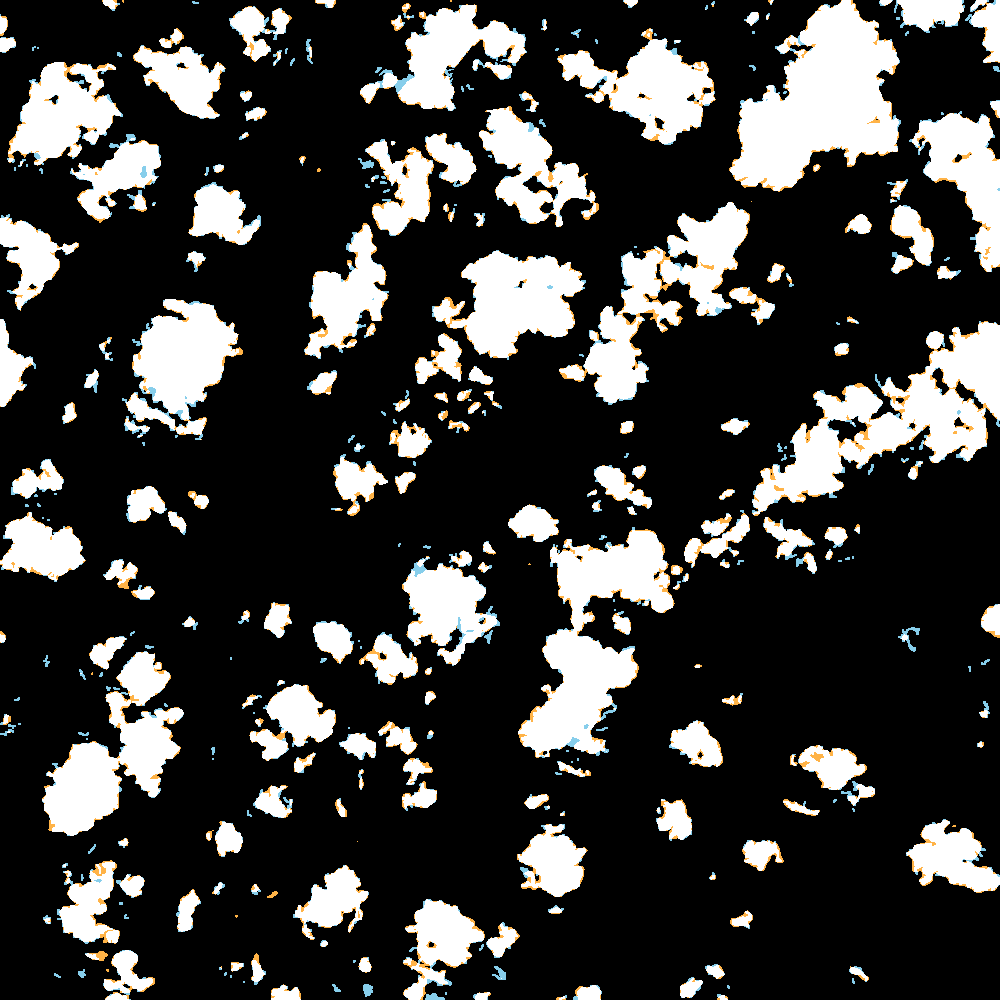} \\
        & \smallimg{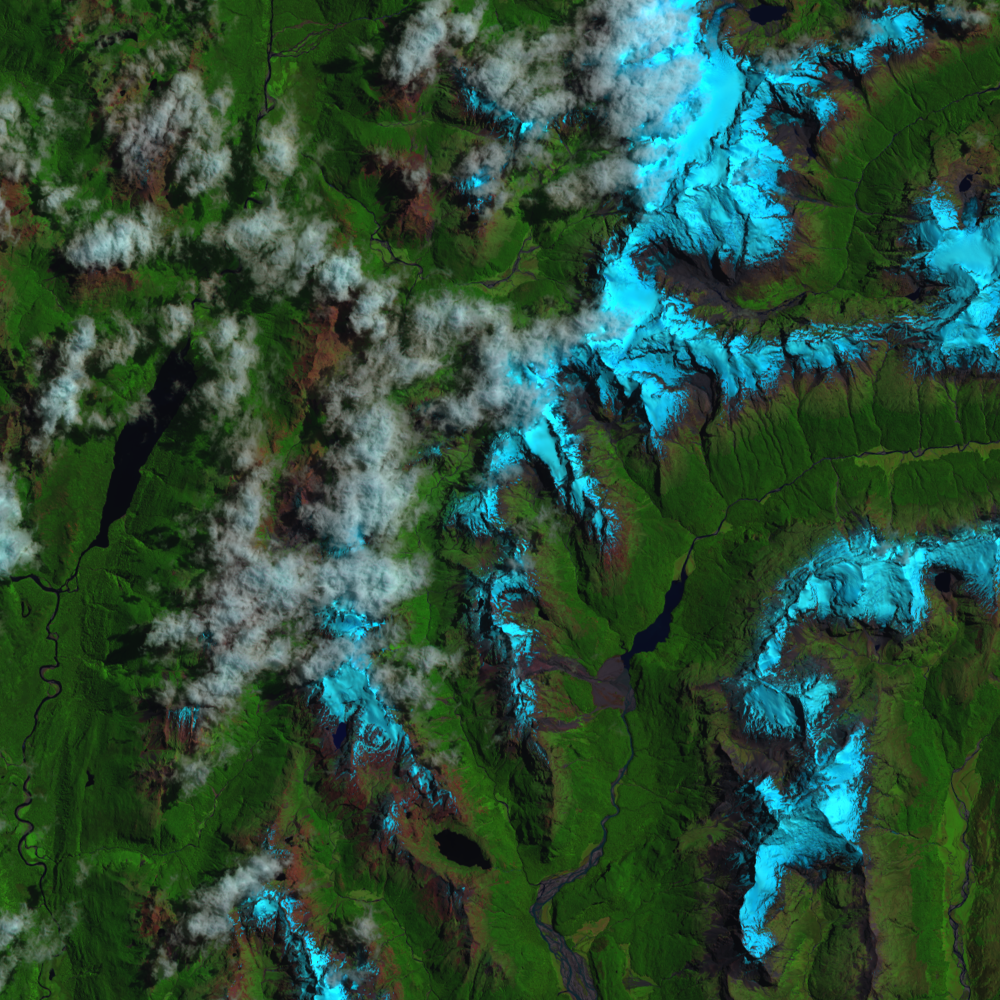} & \smallimg{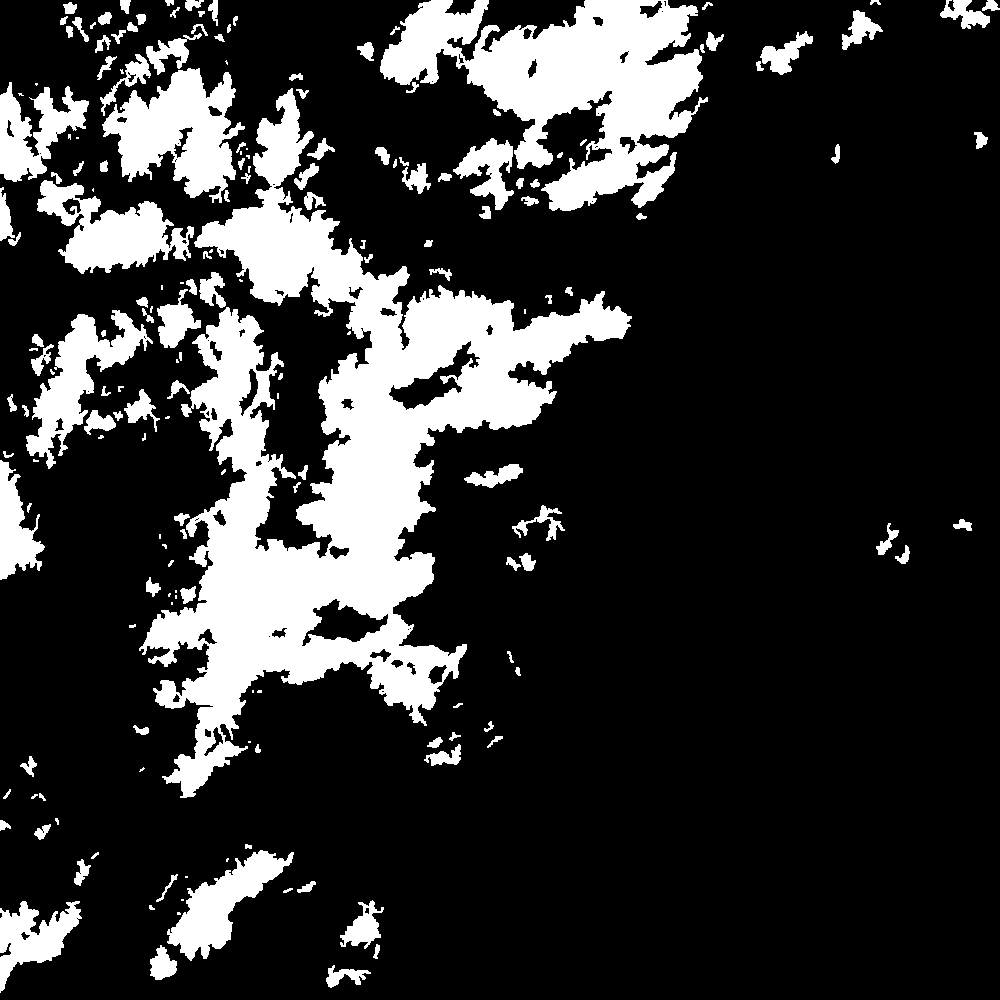} &  \smallimg{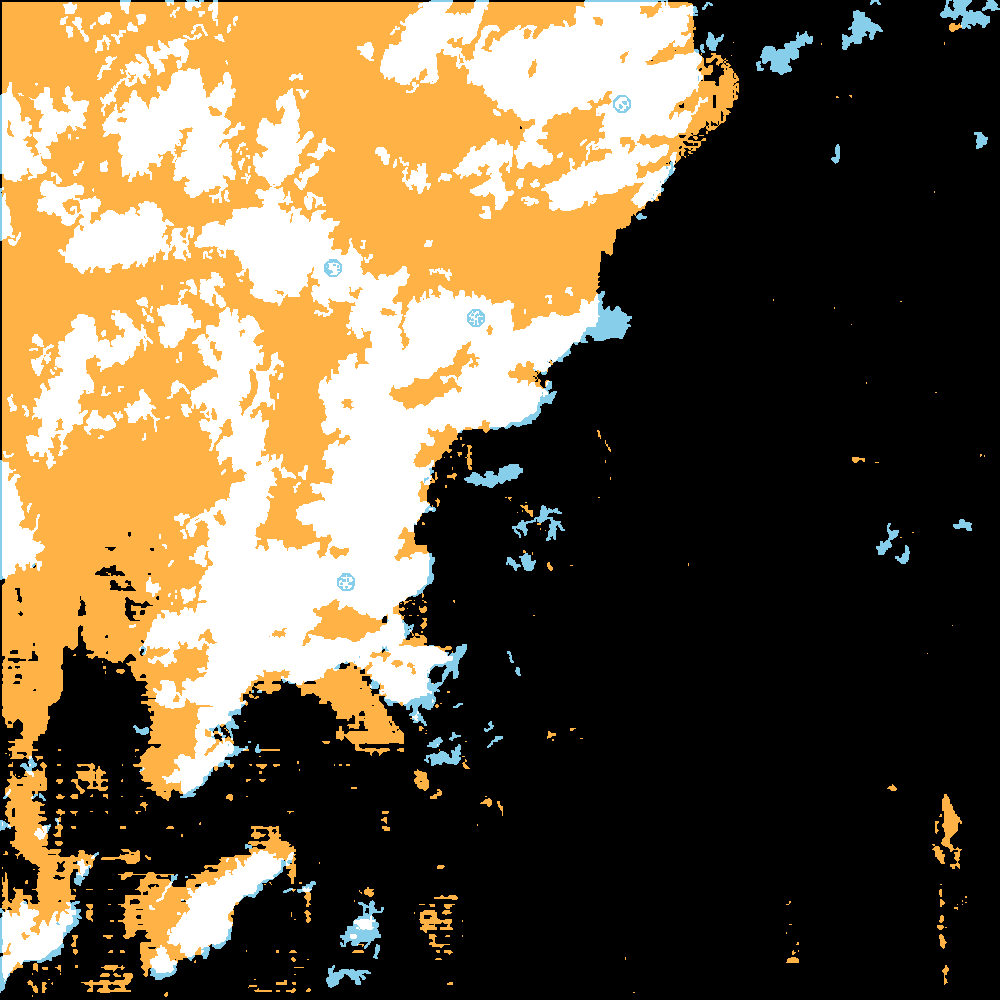} & \smallimg{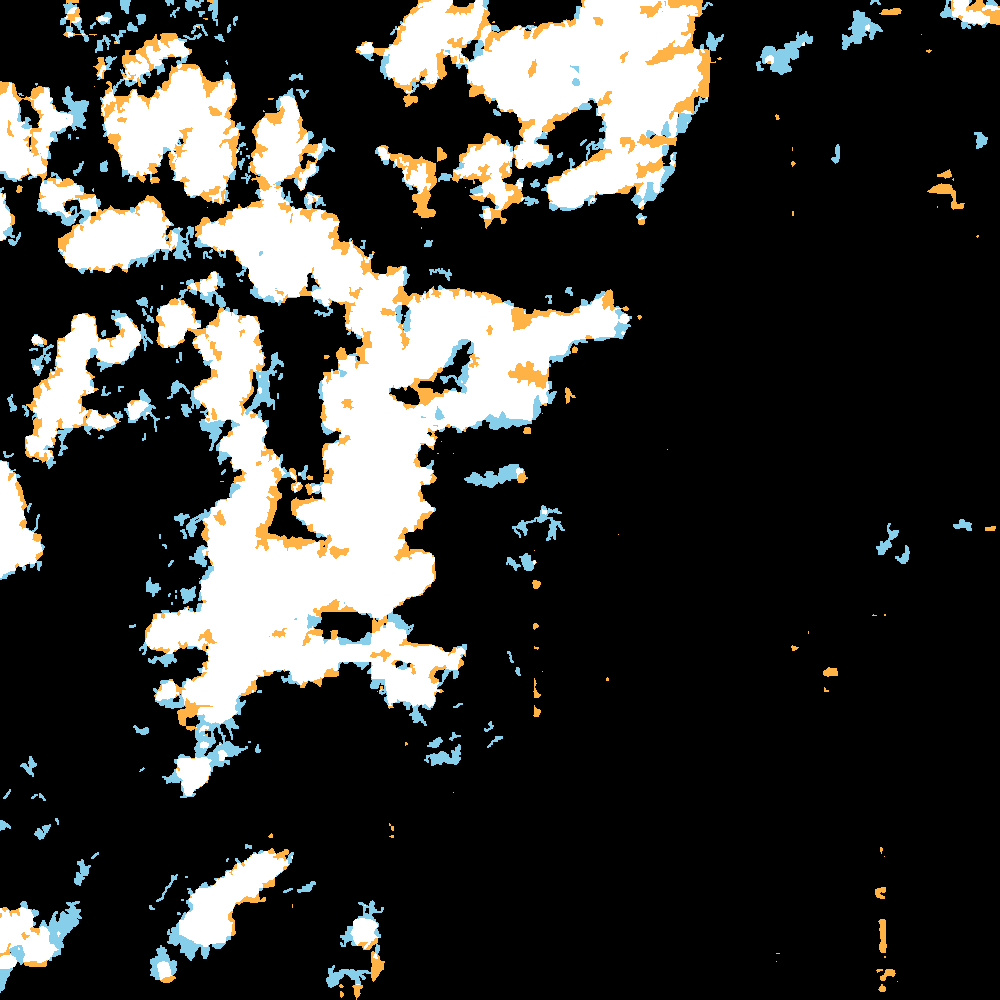} & \smallimg{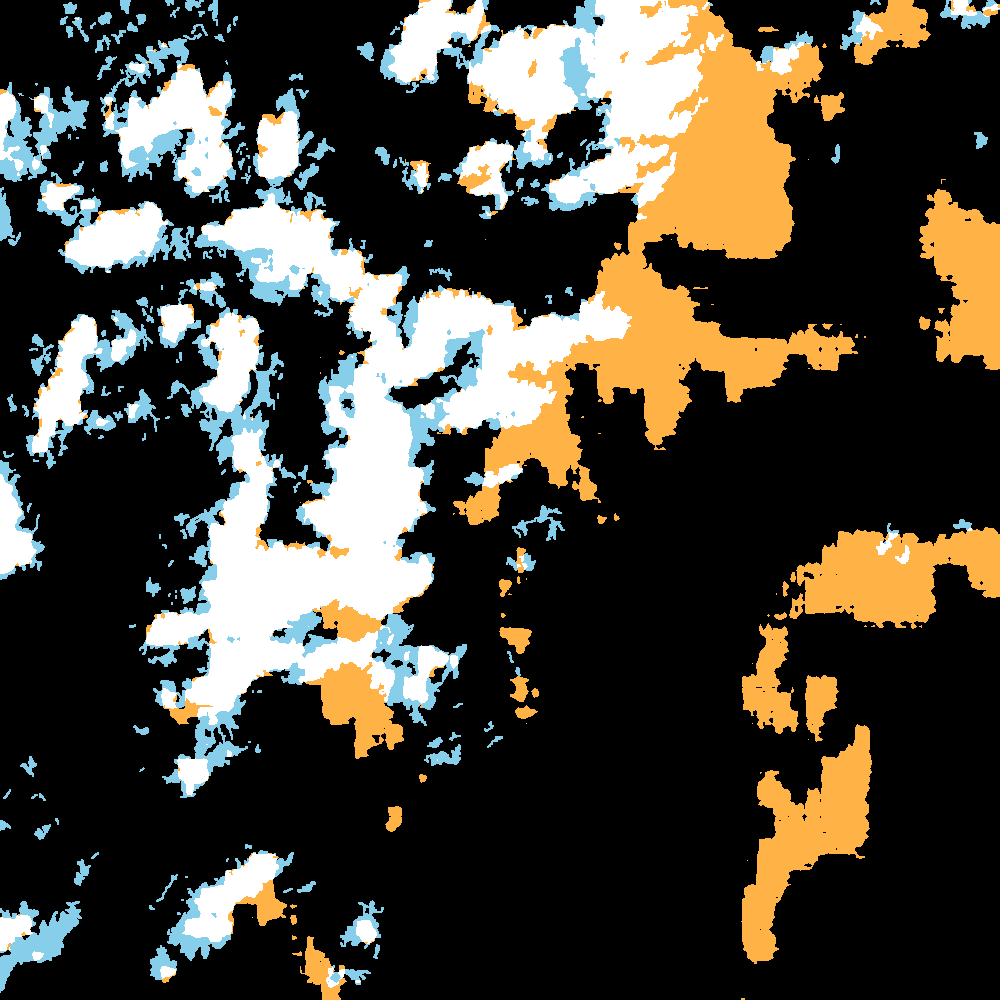} &
        \smallimg{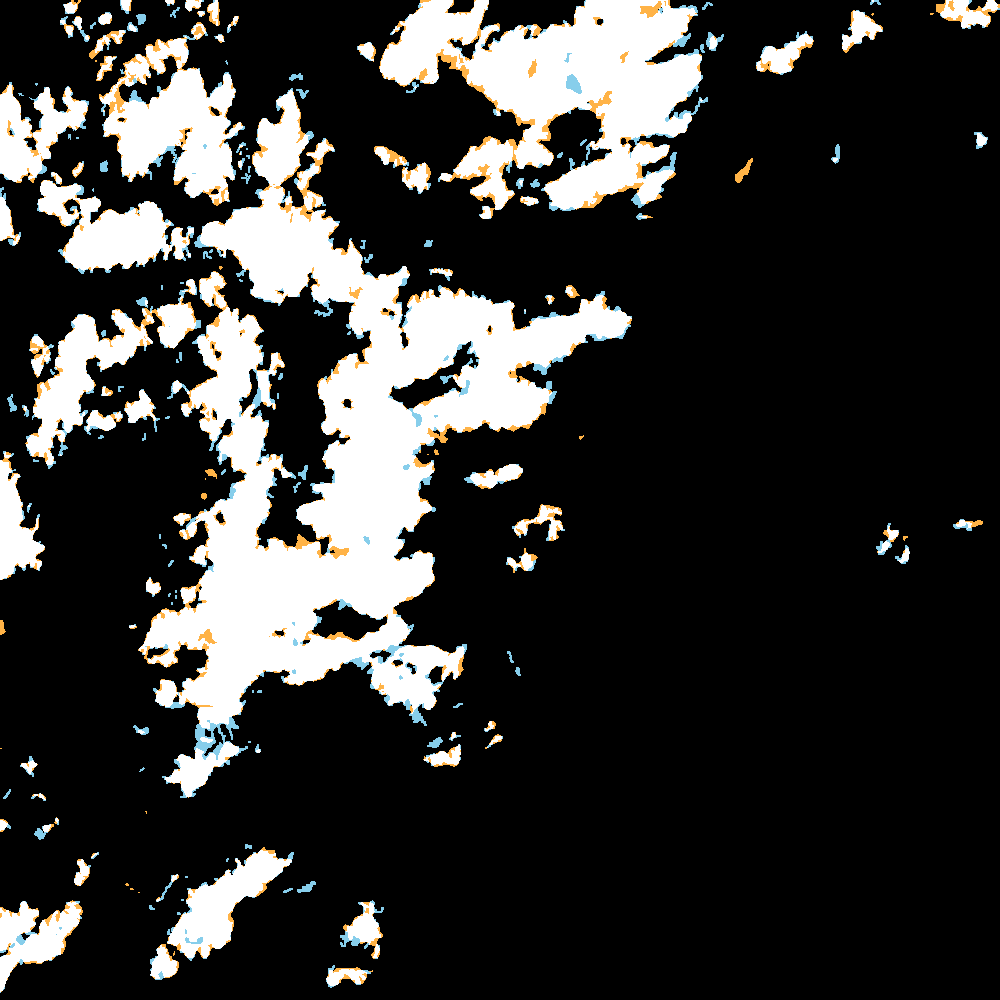} \\

    \end{tabular}

    \caption{Visual comparison of cloud segmentation results. (a) Original image, (b) ground truth, (c) SAM (manual), (d) EfficientSAM, (e) MobileSAM, (f) GeoSAM-Lite. \textcolor[rgb]{1.0, 0.7, 0.28}{Orange} regions indicate False Positives (background misclassified as cloud), and \textcolor[rgb]{0.53, 0.81, 0.92}{Blue} regions indicate False Negatives (missed cloud areas).}
    \label{fig:qualitative_results}
\end{figure}

\subsection{Multi-class Segmentation on SPARCS}
\begin{figure}[!t]
    \centering
    \newcommand{\smallimg}[1]{\includegraphics[width=0.145\columnwidth]{#1}}
    \setlength{\tabcolsep}{1.2pt} 
    \renewcommand{\arraystretch}{0.6} 
    \scriptsize 

    \begin{tabular}{c cccccc}
        & \textbf{(a)} & \textbf{(b)} & \textbf{(c)}  & \textbf{(d)} & \textbf{(e)} & \textbf{(f)}\\

        \multirow{3}{*}{\rotatebox{90}{\textbf{SPARCS}}} 
        & \smallimg{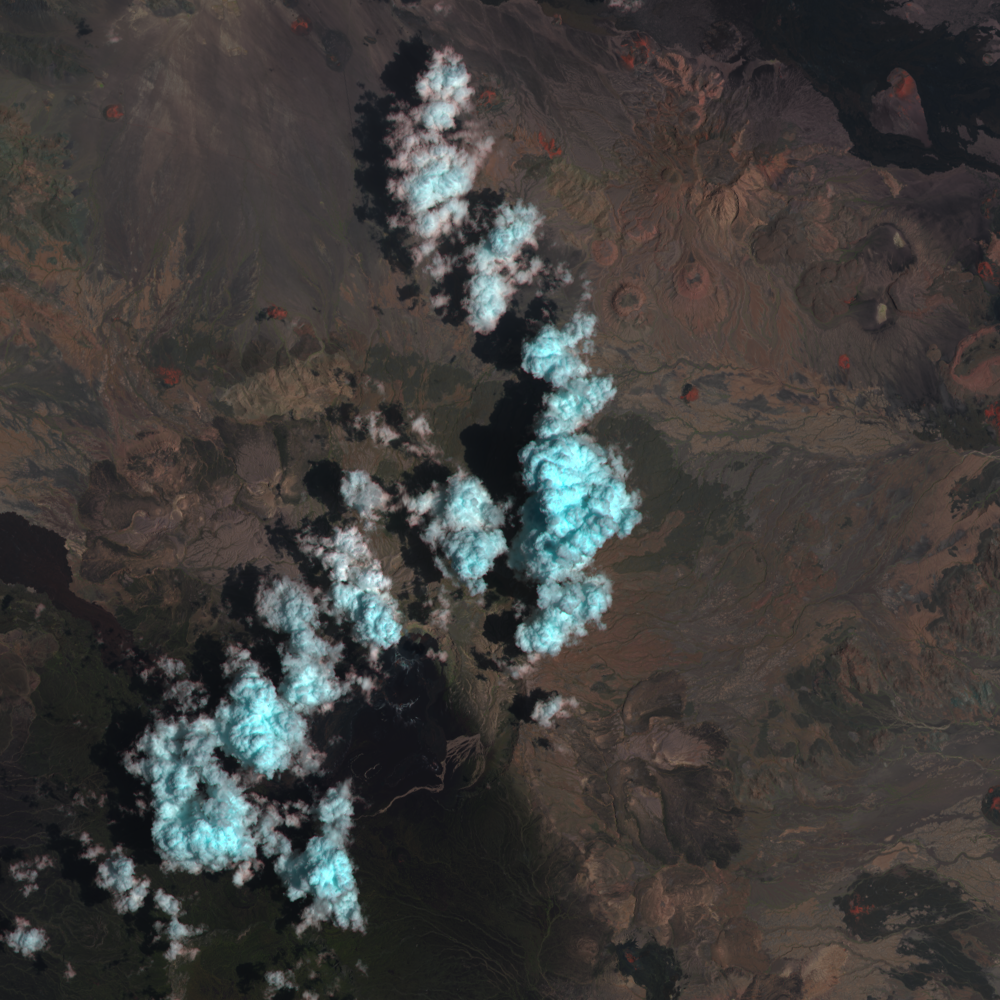} & \smallimg{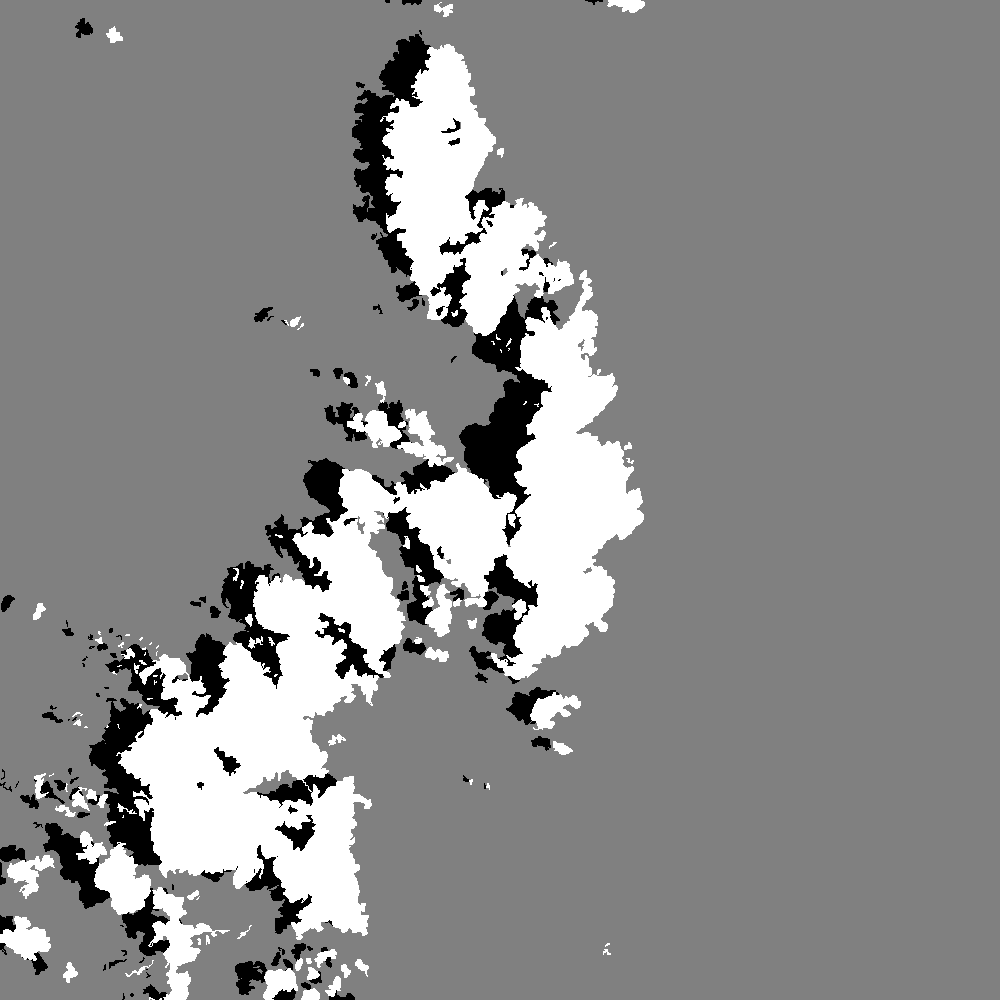} 
         
        & \smallimg{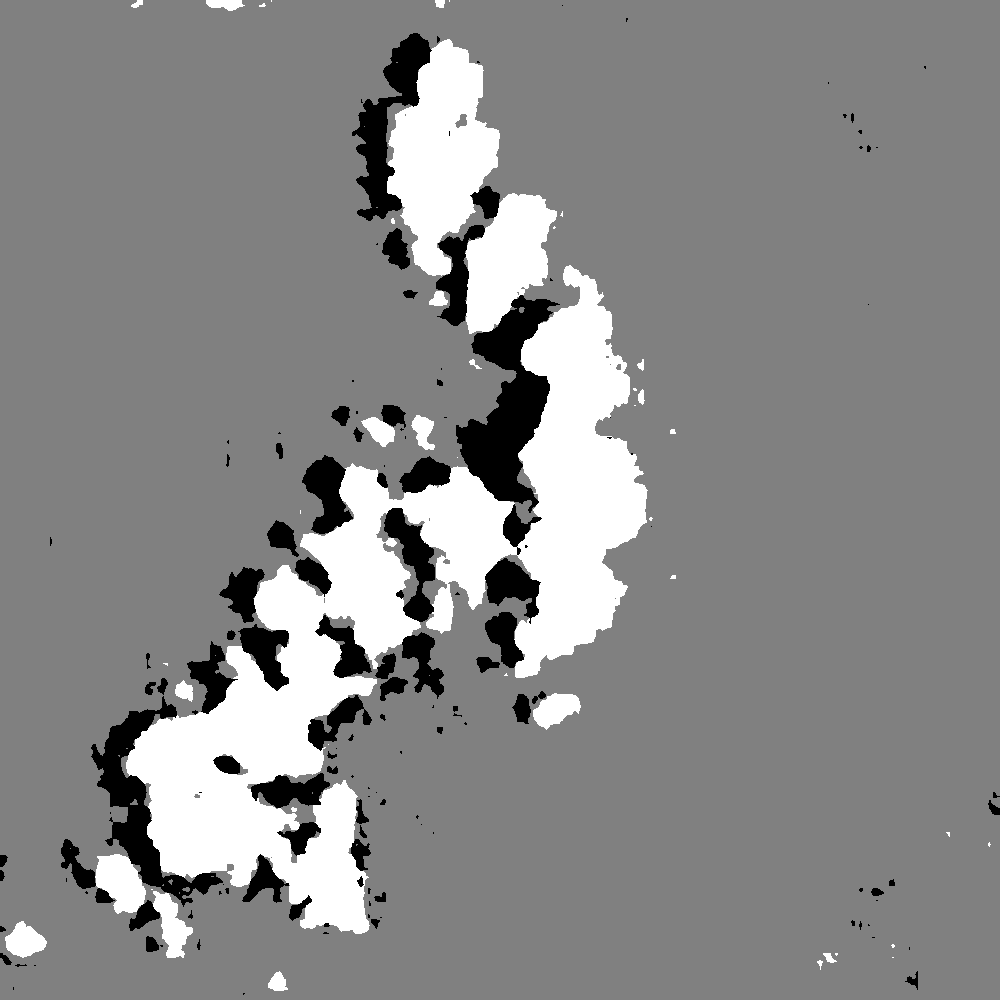} 
        & \smallimg{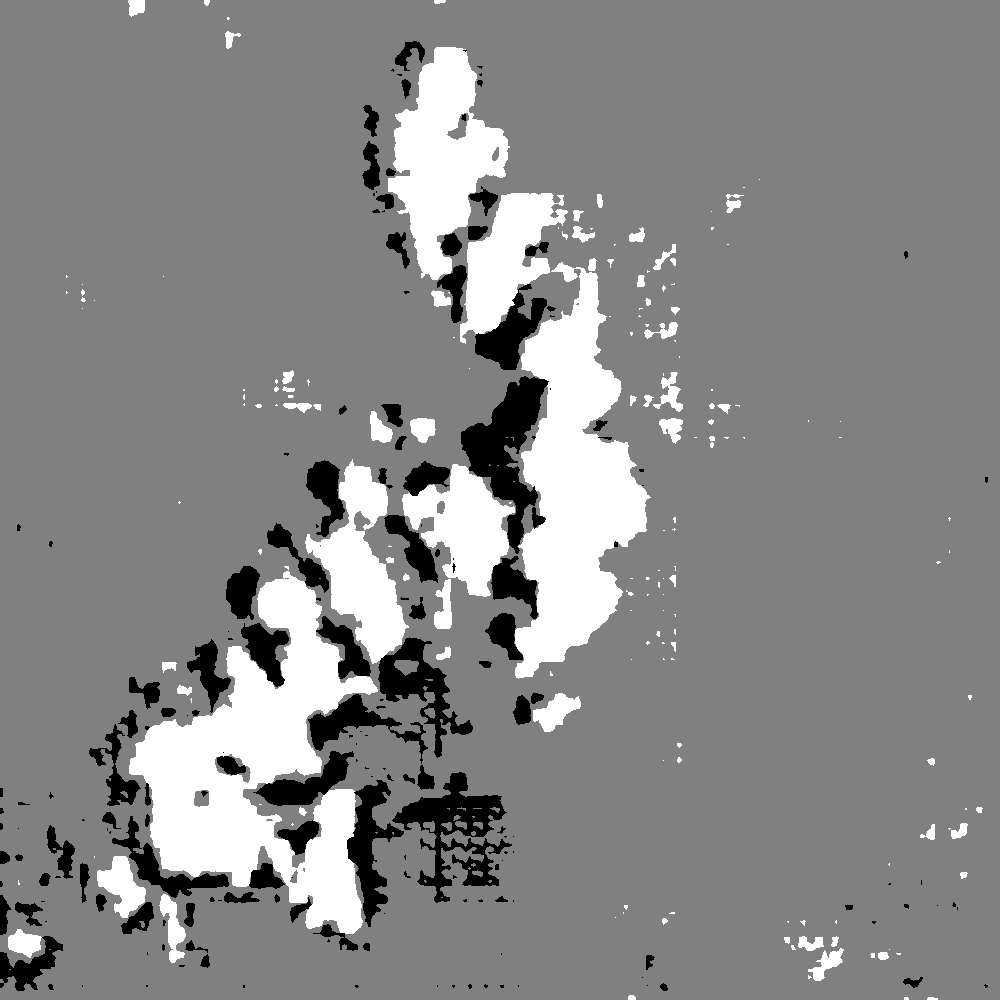}
        & \smallimg{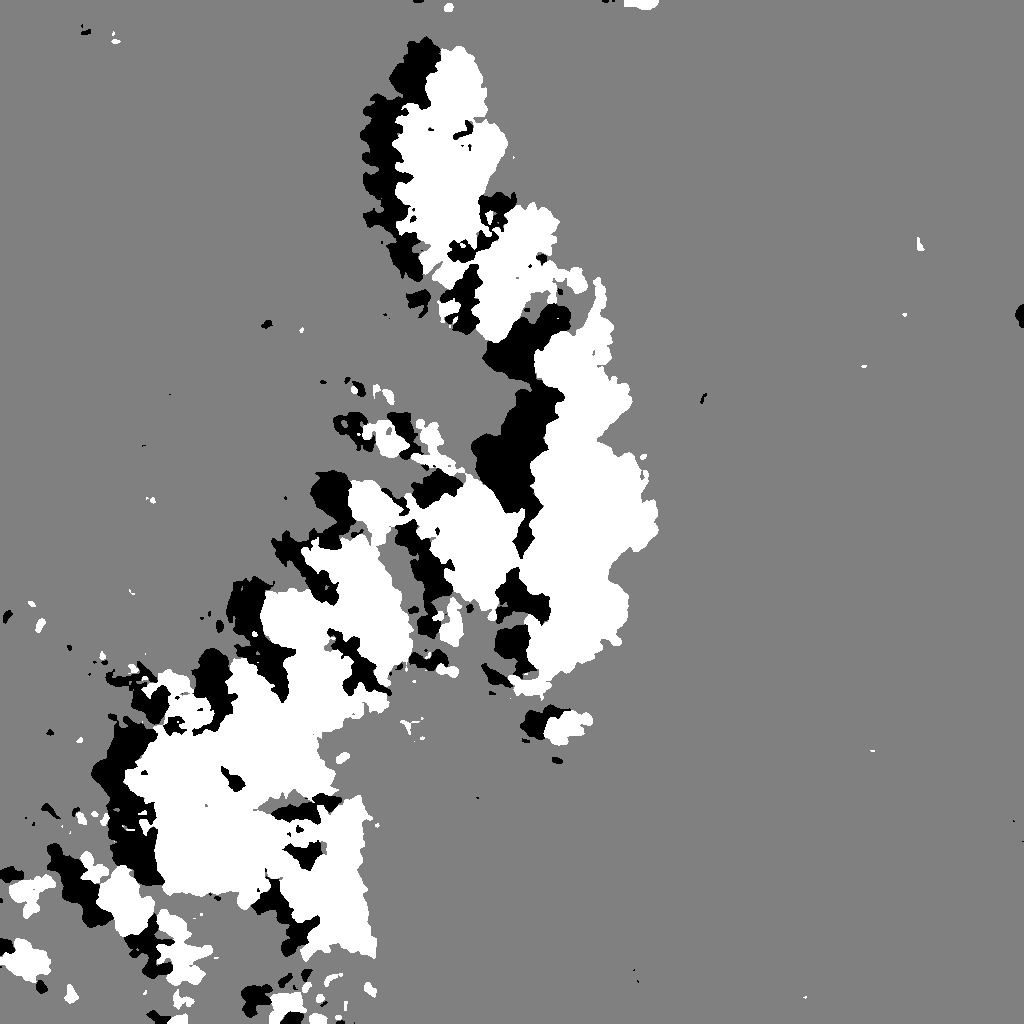} 
        & \smallimg{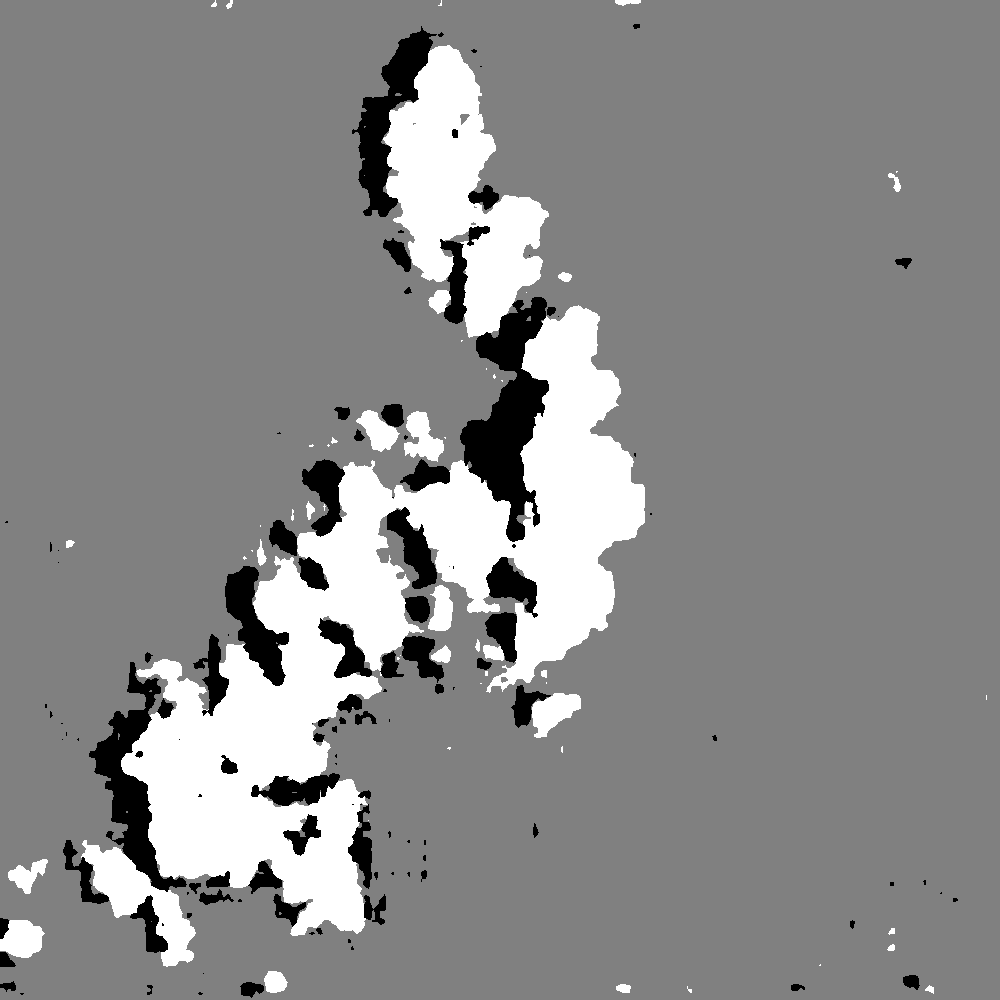} \\
        & \smallimg{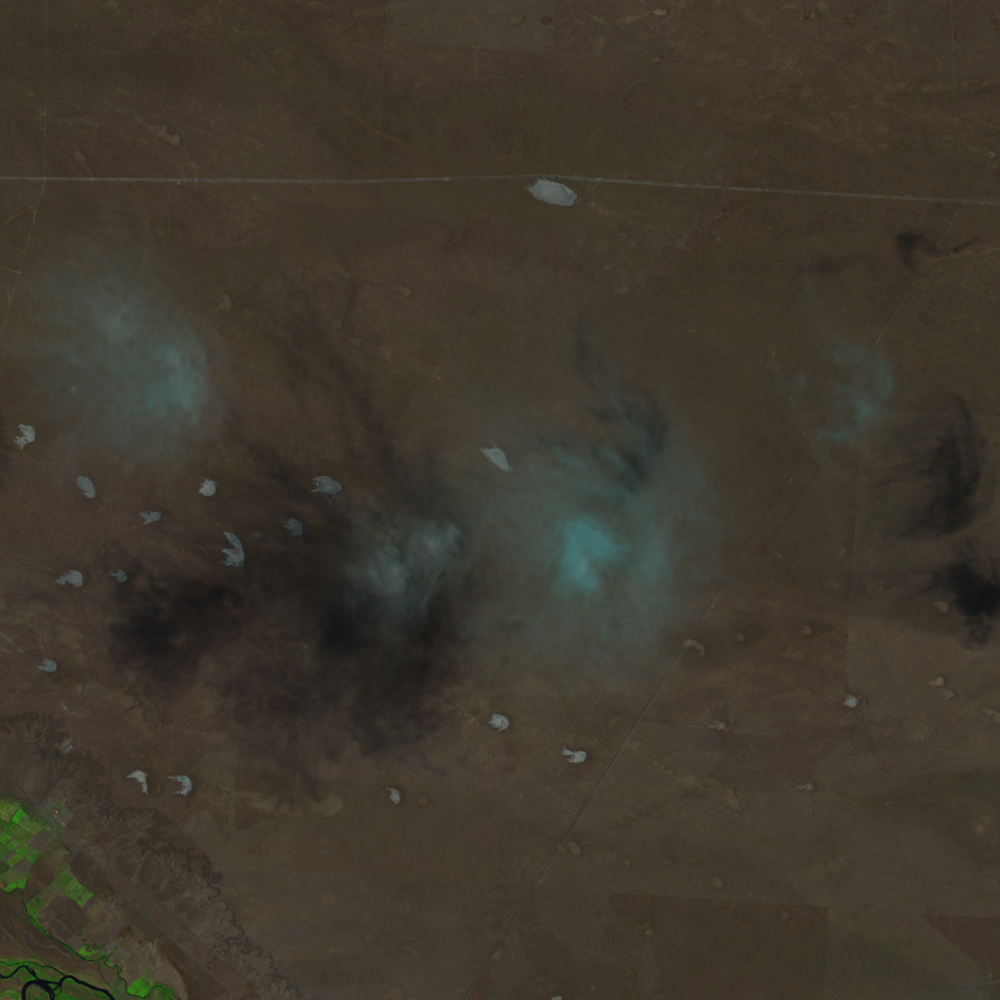} & \smallimg{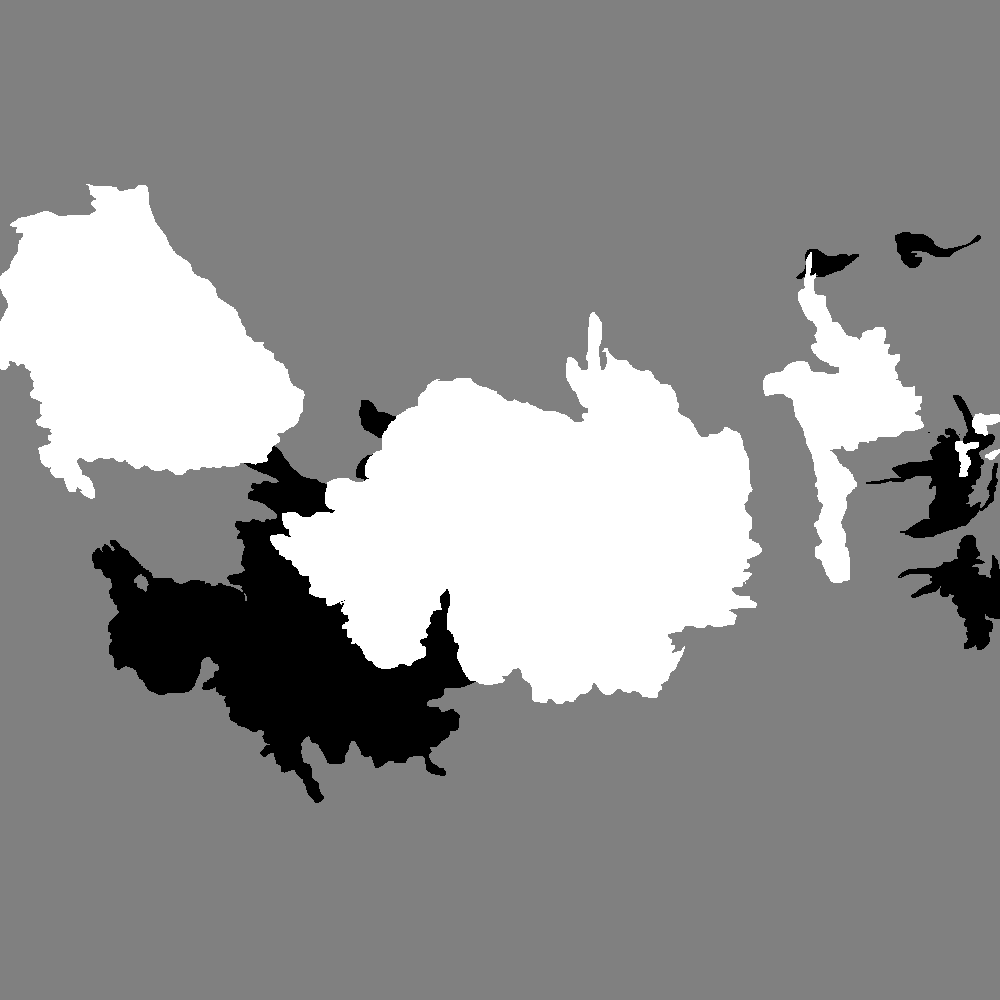}  & \smallimg{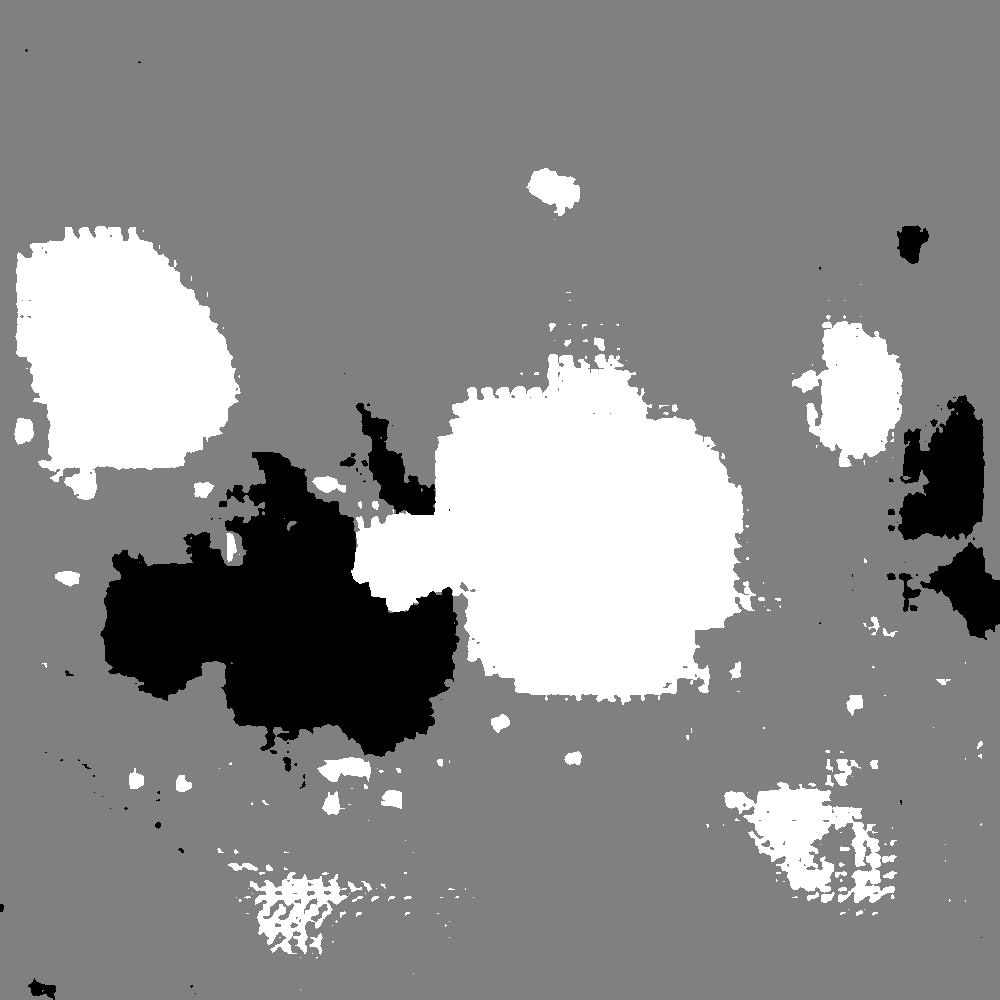} & \smallimg{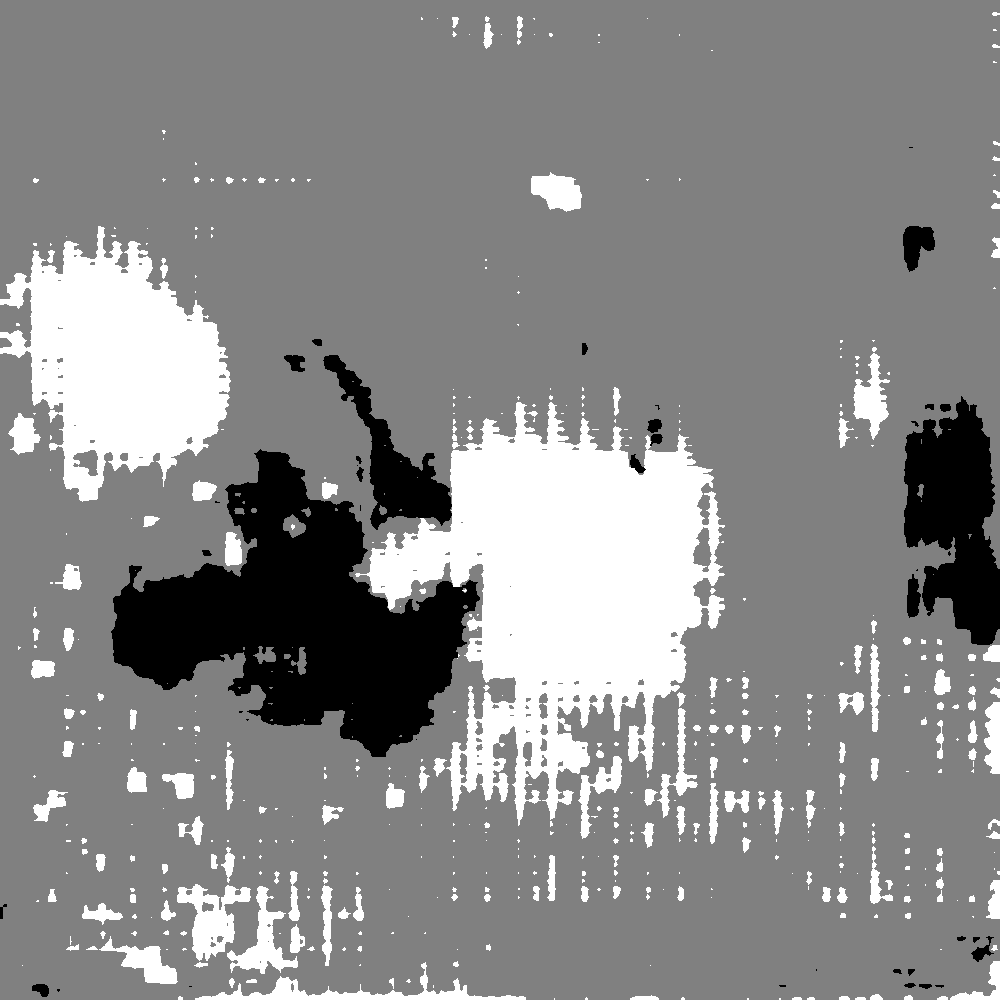} 
        & \smallimg{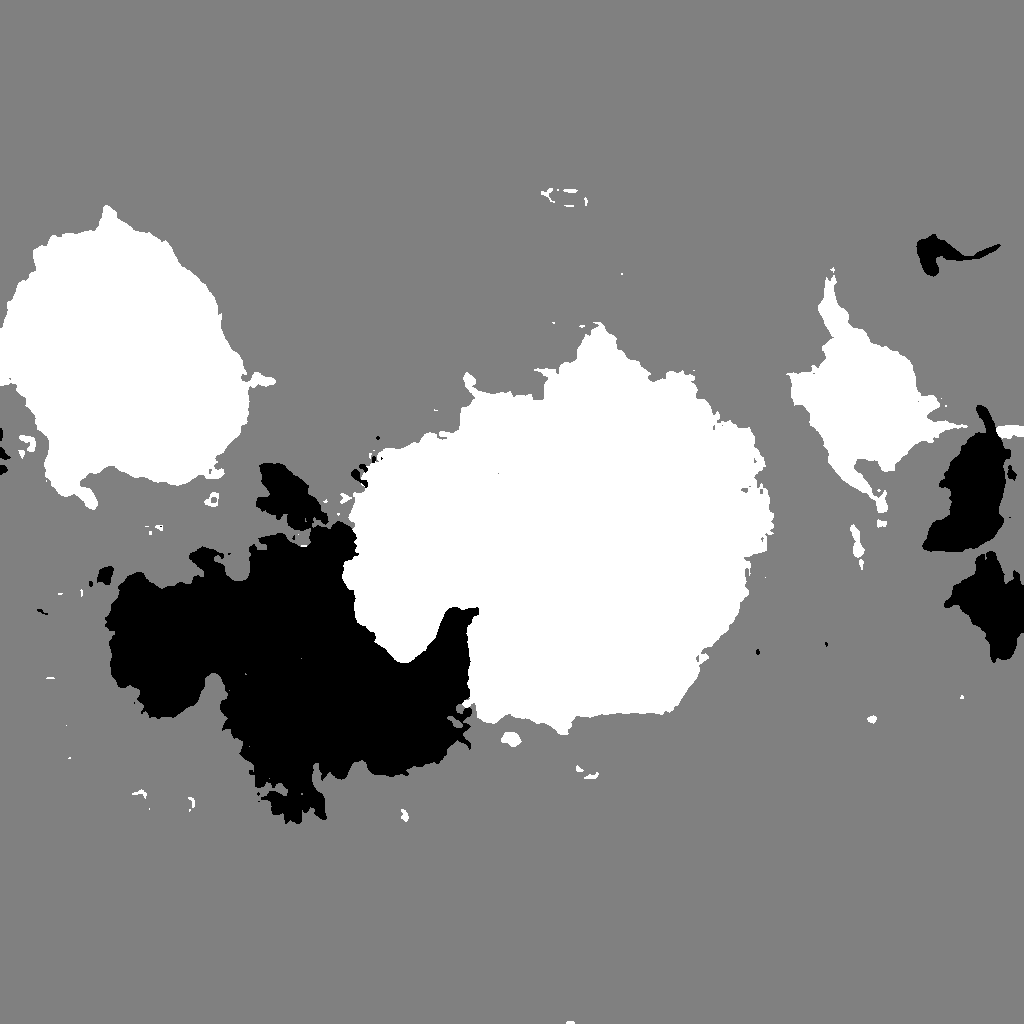}
        & \smallimg{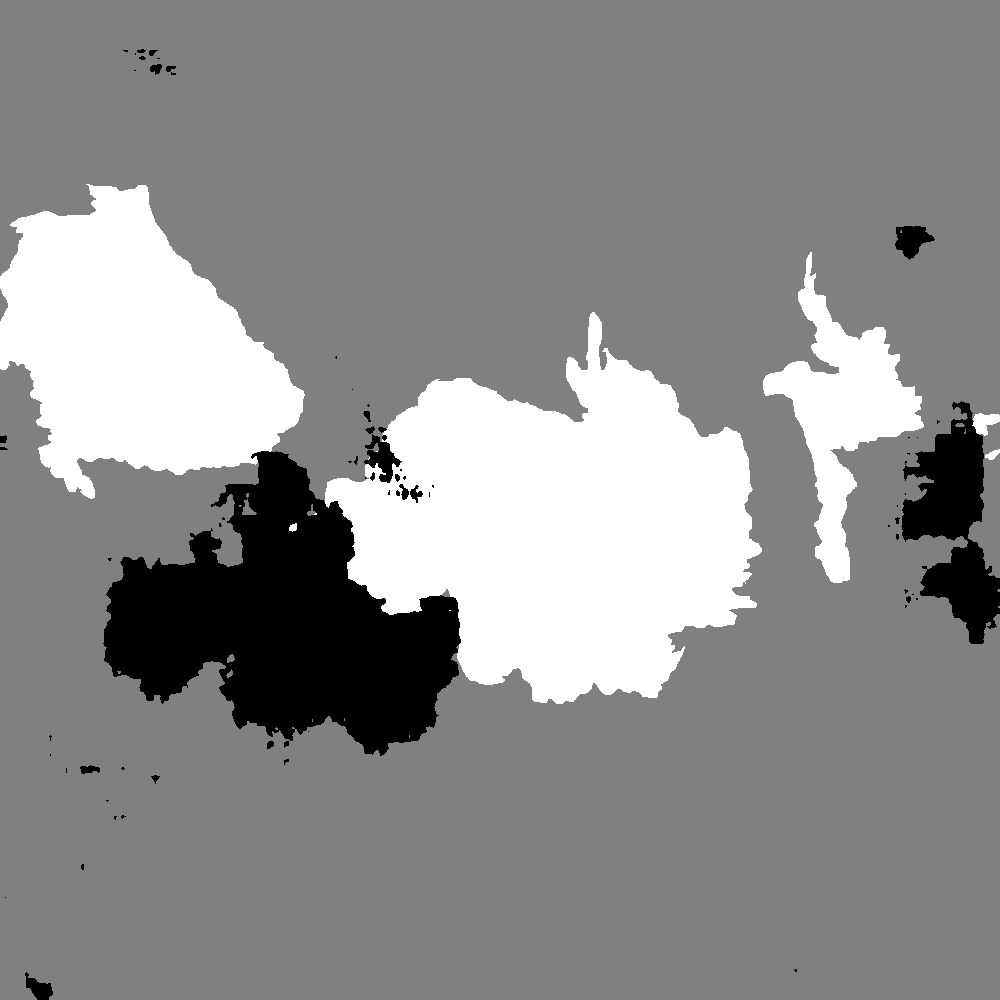} \\
        & \smallimg{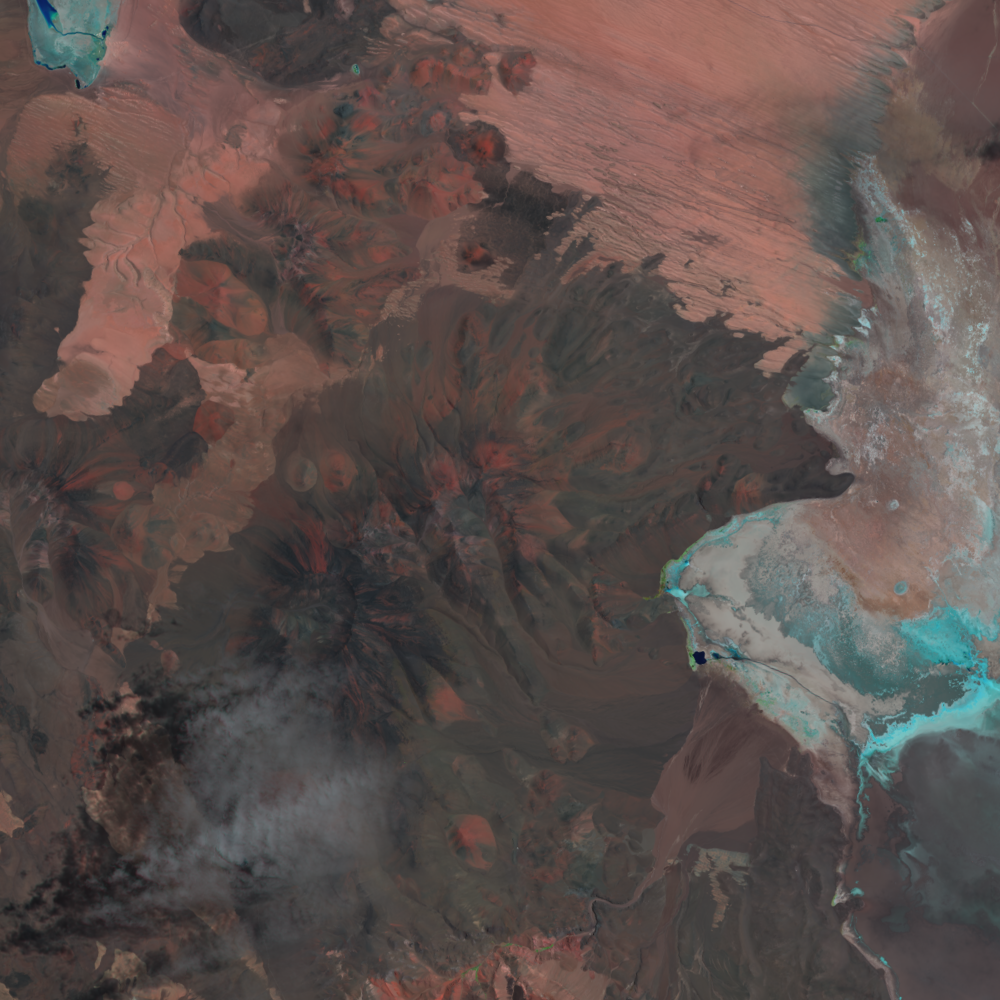} & \smallimg{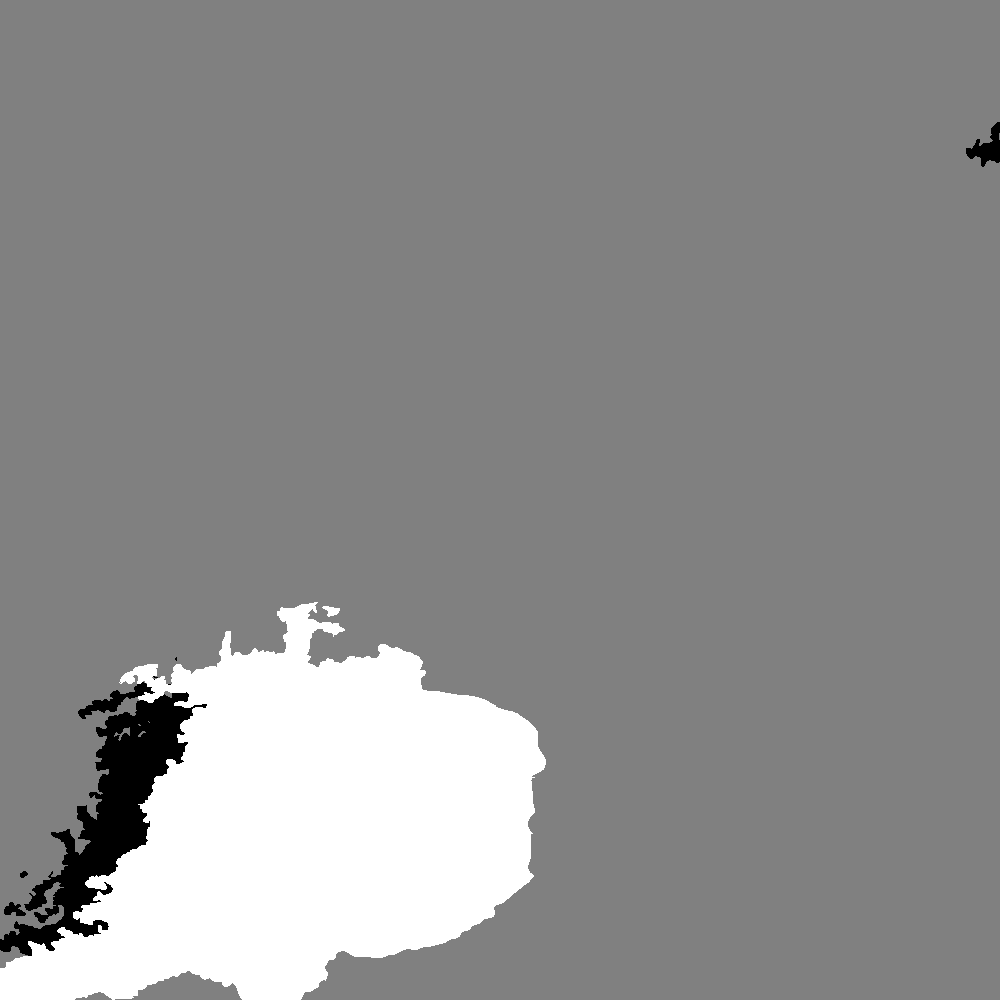}  & \smallimg{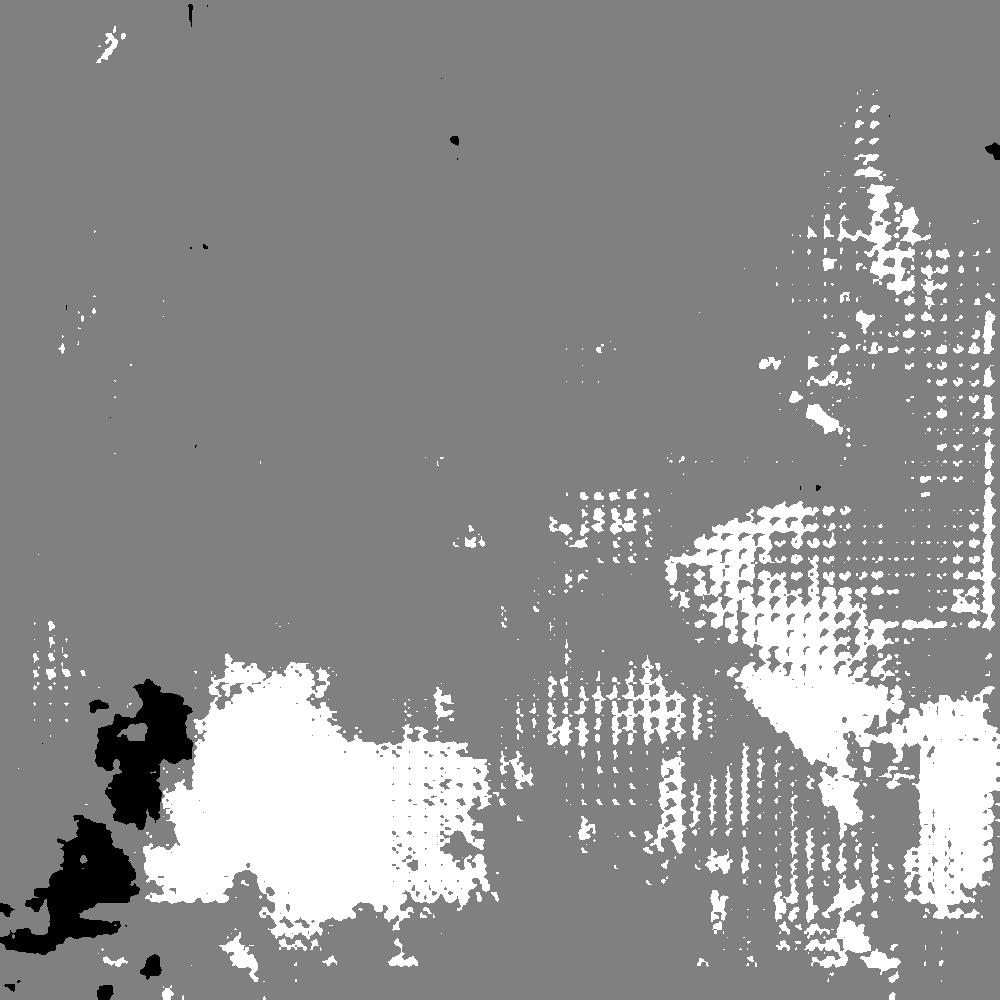} & \smallimg{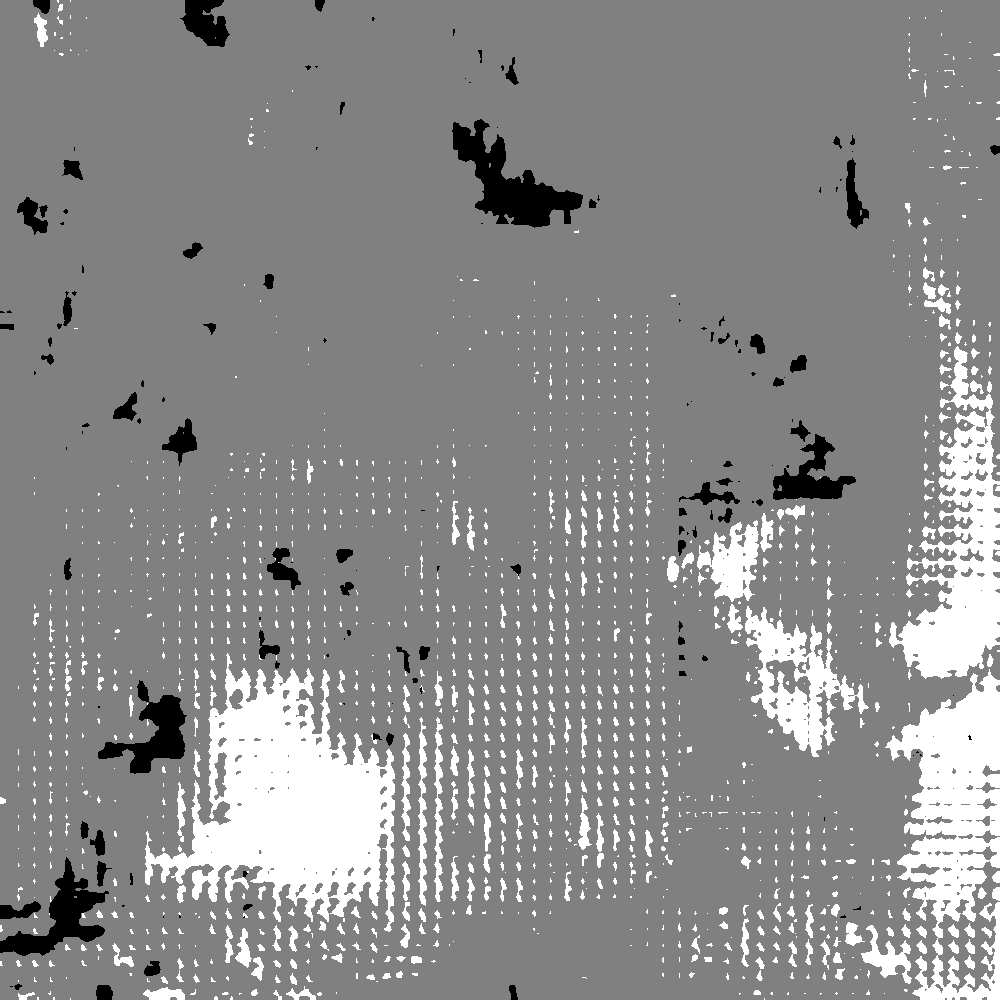} 
        & \smallimg{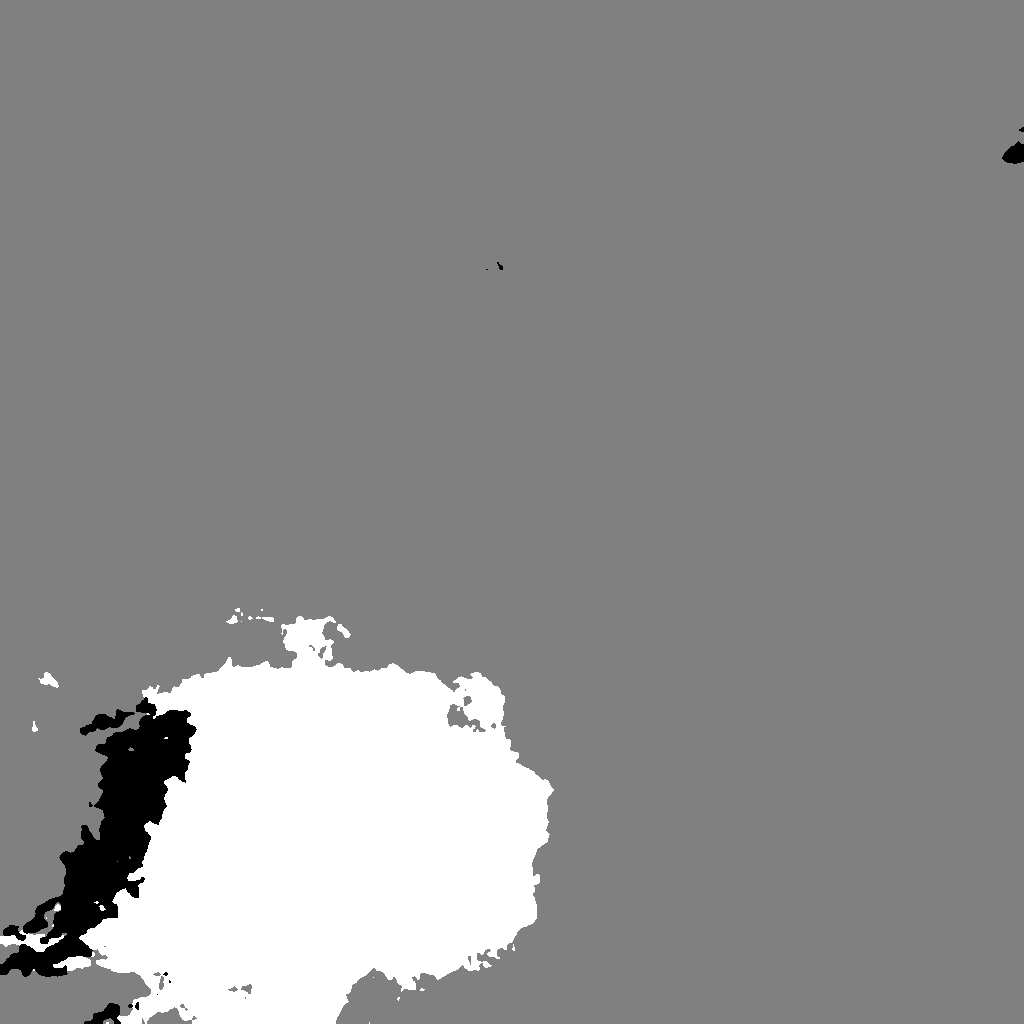}
        & \smallimg{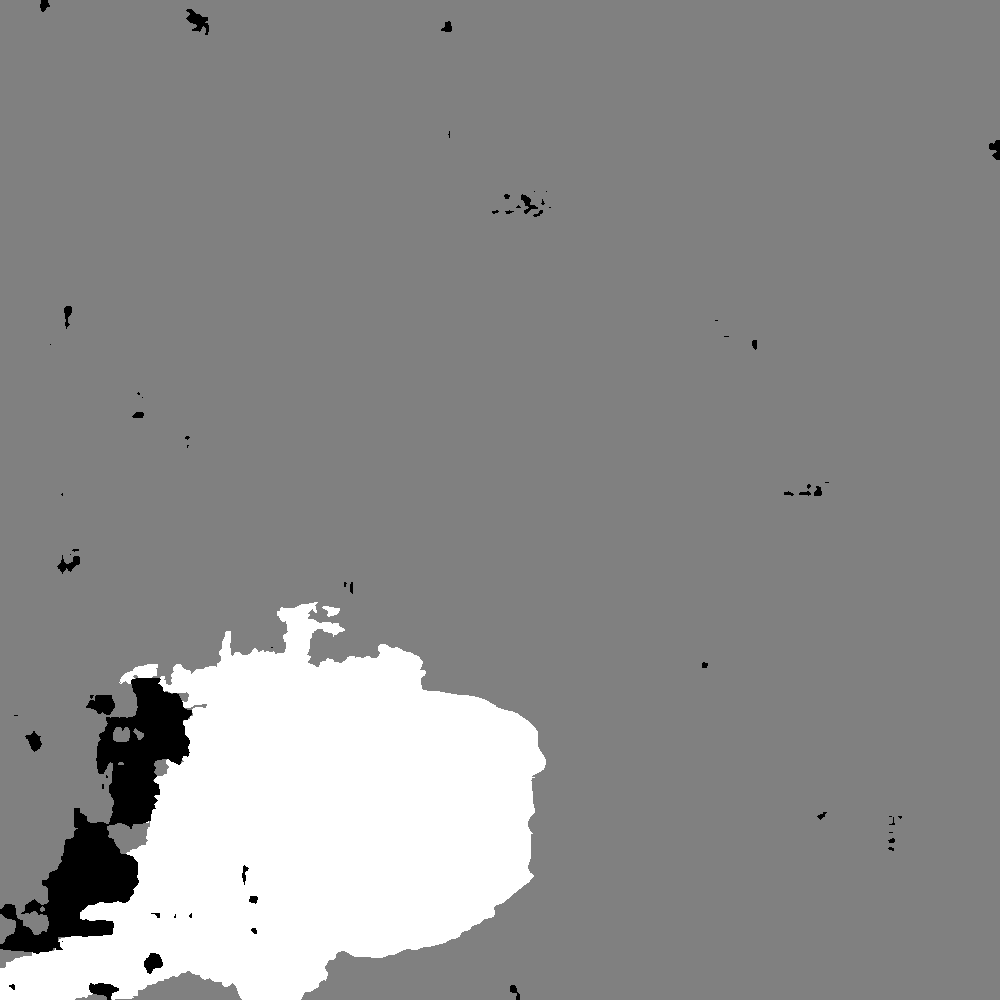} \\
    \end{tabular}
    
    \caption{Multi-class segmentation results in cloud and shadow scenes. (a) Original image, (b) Ground truth, (c) EfficientSAM, (d) MobileSAM, (e) RSAM-Seg, (f) GeoSAM-Lite. White, black, and gray regions represent cloud, shadow, and background, respectively.}
    \label{fig:sparcs_multi}
\end{figure}

To further evaluate the model's capability in complex multi-class cloud and shadow scenarios, we present the segmentation results for cloud and shadow on the SPARCS dataset in Fig.~\ref{fig:sparcs_multi}. Despite its lightweight architecture, GeoSAM-Lite accurately distinguishes clouds, shadows, and background, preserving more precise morphological features and boundary details than EfficientSAM and MobileSAM. While its boundary precision is marginally lower than the heavyweight RSAM-Seg, GeoSAM-Lite maintains highly competitive overall segmentation quality, successfully balancing performance and efficiency for resource-constrained remote sensing applications.

\subsection{Ablation Study}
We conduct an ablation study on the 38-Cloud dataset to verify the contribution of each GeoSAM-Lite component. The results in Table \ref{tab:ablation_table} indicate that removing either FFL-F or FFL-S leads to a consistent degradation in segmentation performance, with FFL-F being particularly critical for preserving high-frequency boundary details. Furthermore, the exclusion of the Geo-Init pre-training strategy causes a significant drop across four key metrics, confirming that distilling knowledge from a Domain-Expert Teacher is essential for bridging the domain gap. The full GeoSAM-Lite framework yields the best overall balance, confirming the synergy between all components.

\begin{table}[!t]
\centering
\caption{Ablation Study on the 38-Cloud Dataset}
\label{tab:ablation_table}
\renewcommand{\arraystretch}{0.9}
\setlength{\tabcolsep}{5pt}
\scriptsize
\begin{tabular}{lccccc}
\toprule
\textbf{Method} & \textbf{Jaccard} & \textbf{F1} & \textbf{OA} & \textbf{Precision} 
& \textbf{Recall} \\ 
\midrule
GeoSAM-Lite & \textbf{0.6840} & \textbf{0.7740} & \textbf{0.8907} & \textbf{0.7837} & 0.8266 \\ 
\midrule
w/o FFL-F ($F_{\text{embedding}}$) & 0.6823 & 0.7723 & 0.8850 & 0.7640 & 0.8462 \\
w/o FFL-F ($F_{\text{hfc}}$) & 0.6605 & 0.7526 & 0.8734 & 0.7526 & 0.8256 \\
w/o FFL-S & 0.6804 & 0.7715 & 0.8823 & 0.7615 & 0.8411 \\
w/o Geo-Init & 0.6781 & 0.7675 & 0.8854 & 0.7566 & \textbf{0.8483} \\ 
\bottomrule
\end{tabular}
\end{table}

\subsection{Model Efficiency and Lightweight Analysis}
\begin{figure}[!t]
    \centering
    \setlength{\tabcolsep}{1pt} 
    \scriptsize
    \begin{tabular}{@{}cccc@{}}

        \includegraphics[width=0.16\columnwidth]{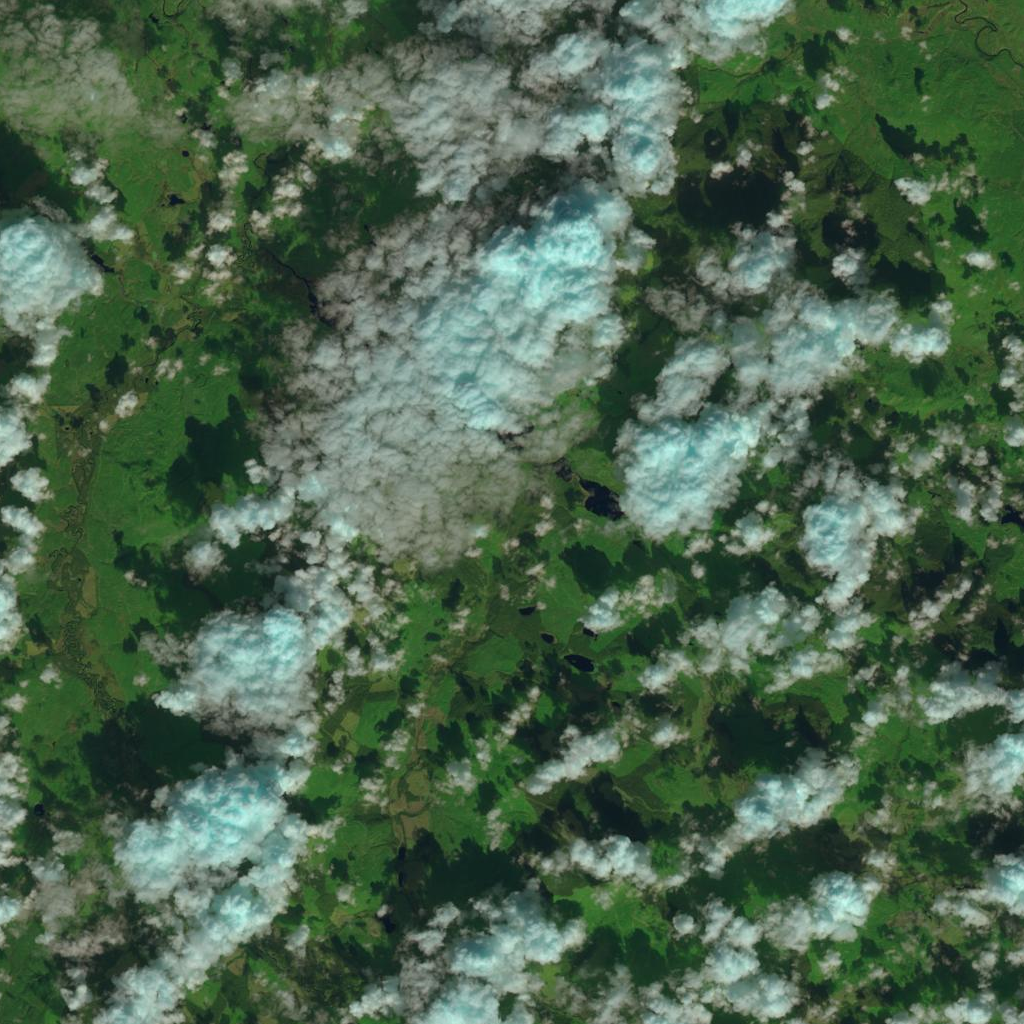} & 
        \includegraphics[width=0.16\columnwidth]{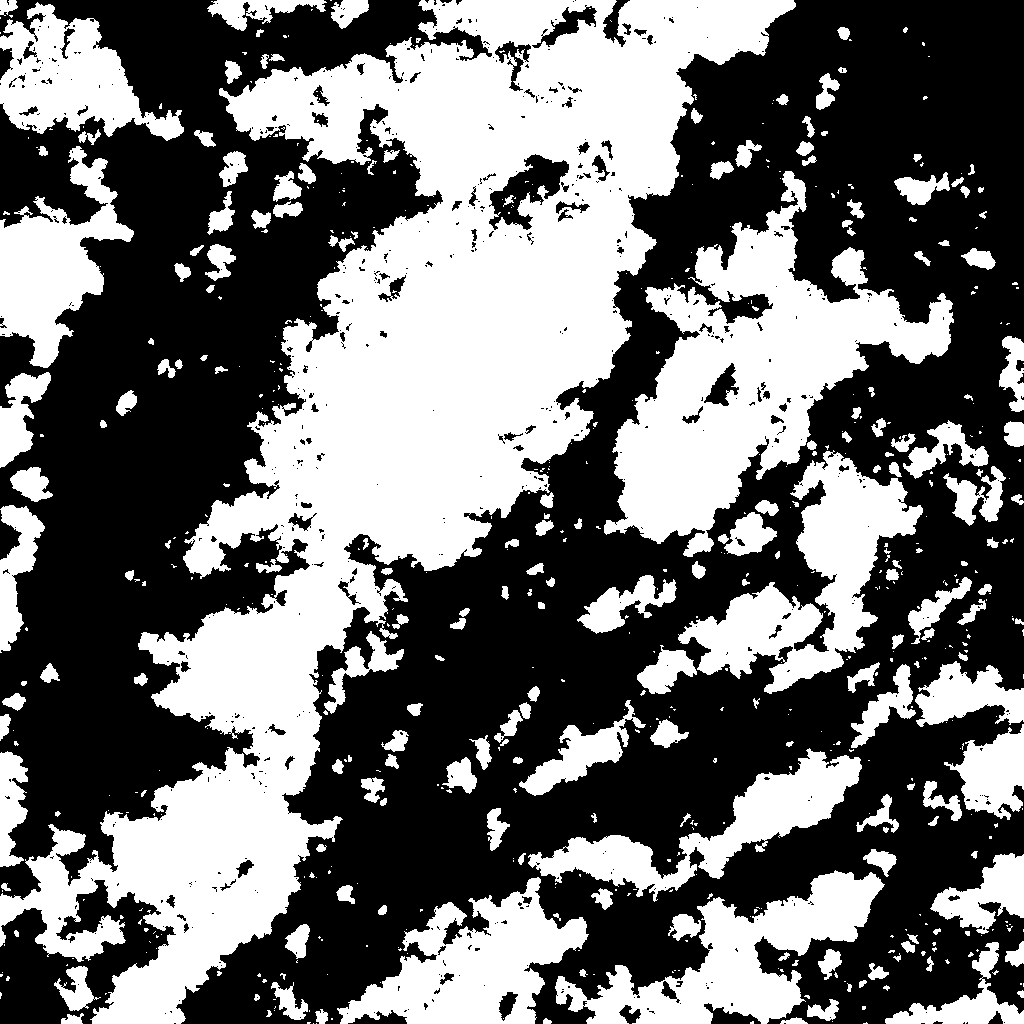} & 
        \includegraphics[width=0.16\columnwidth]{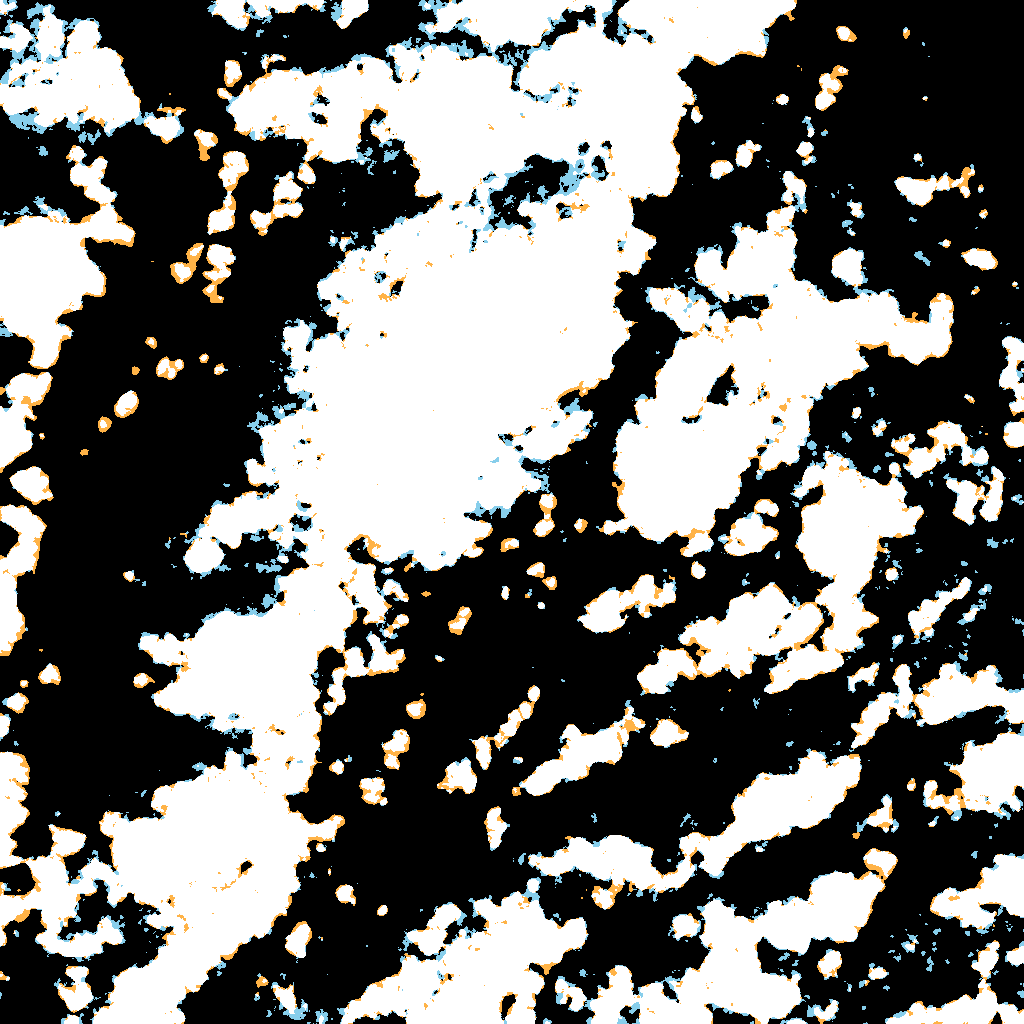} & 
        \includegraphics[width=0.16\columnwidth]{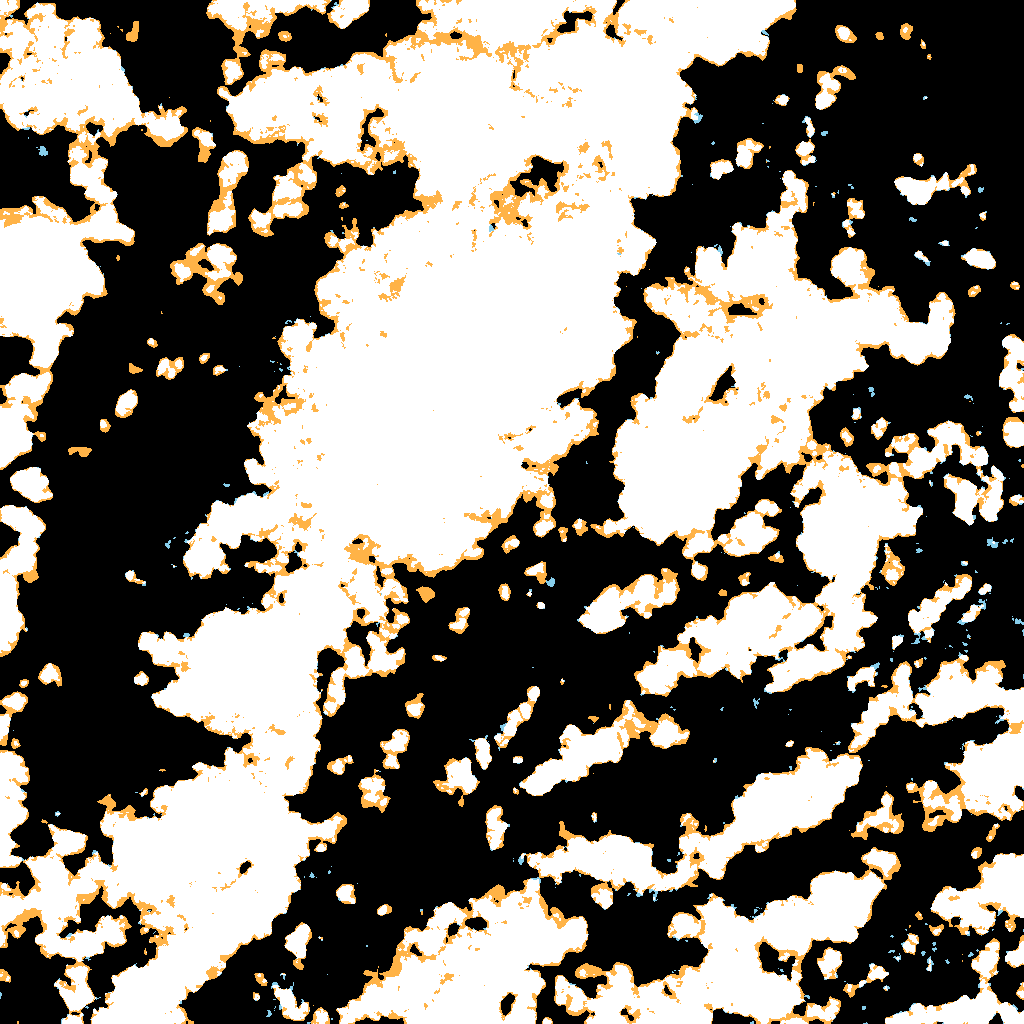} \\[-3pt] 
        
        \includegraphics[width=0.16\columnwidth]{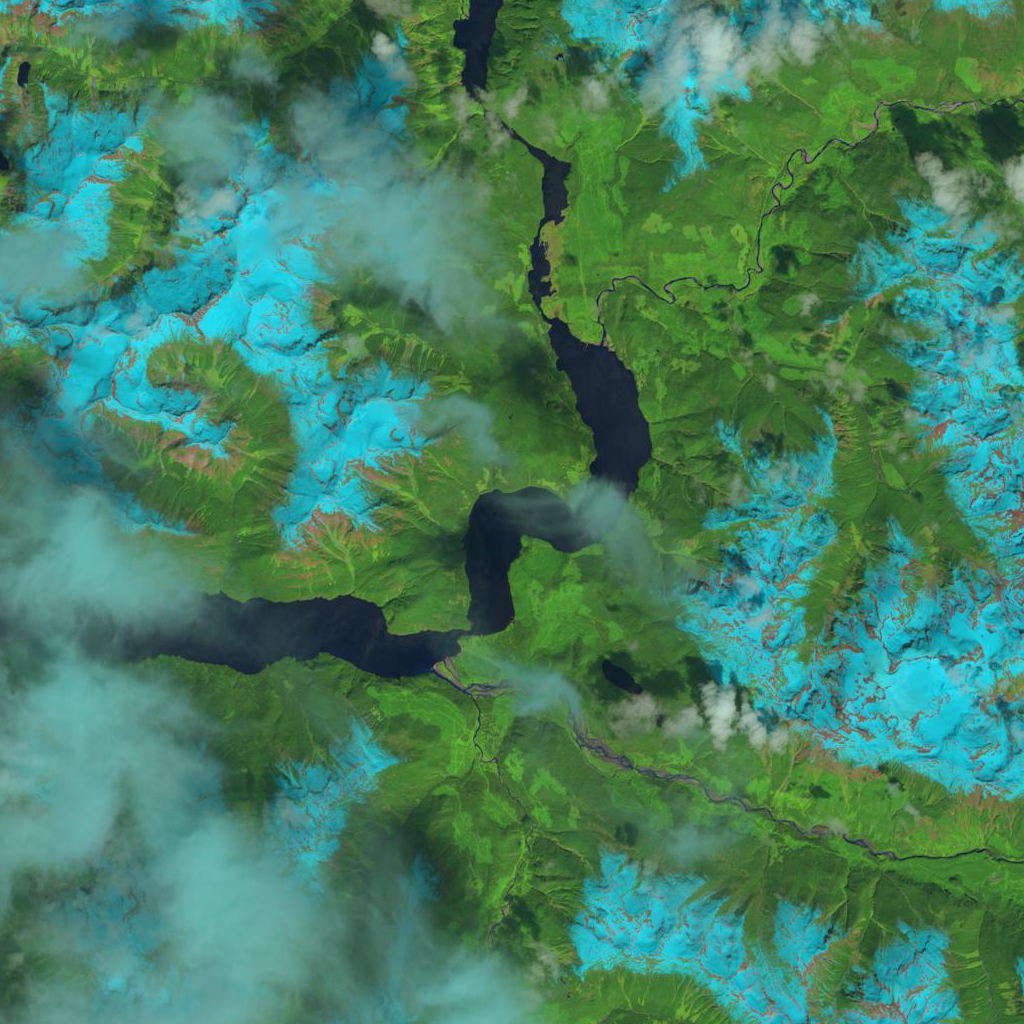} & 
        \includegraphics[width=0.16\columnwidth]{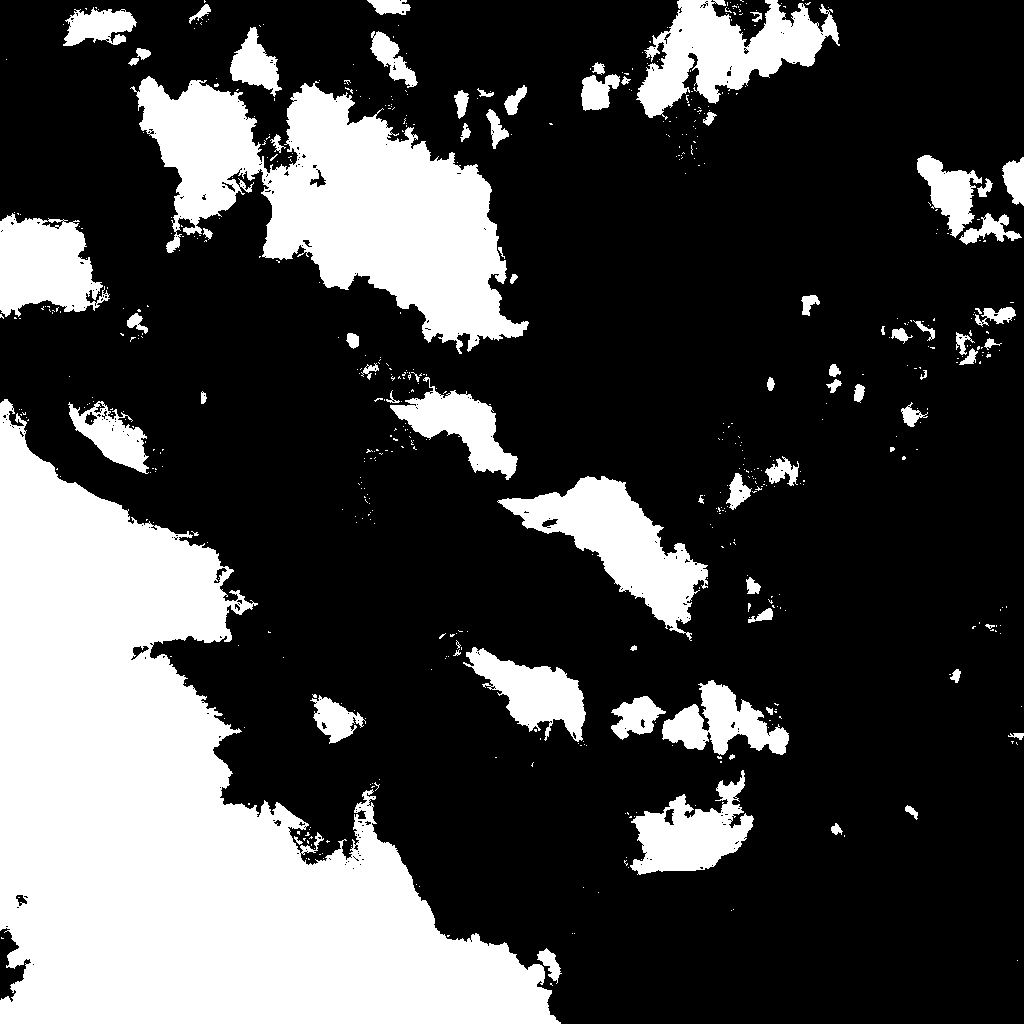} & 
        \includegraphics[width=0.16\columnwidth]{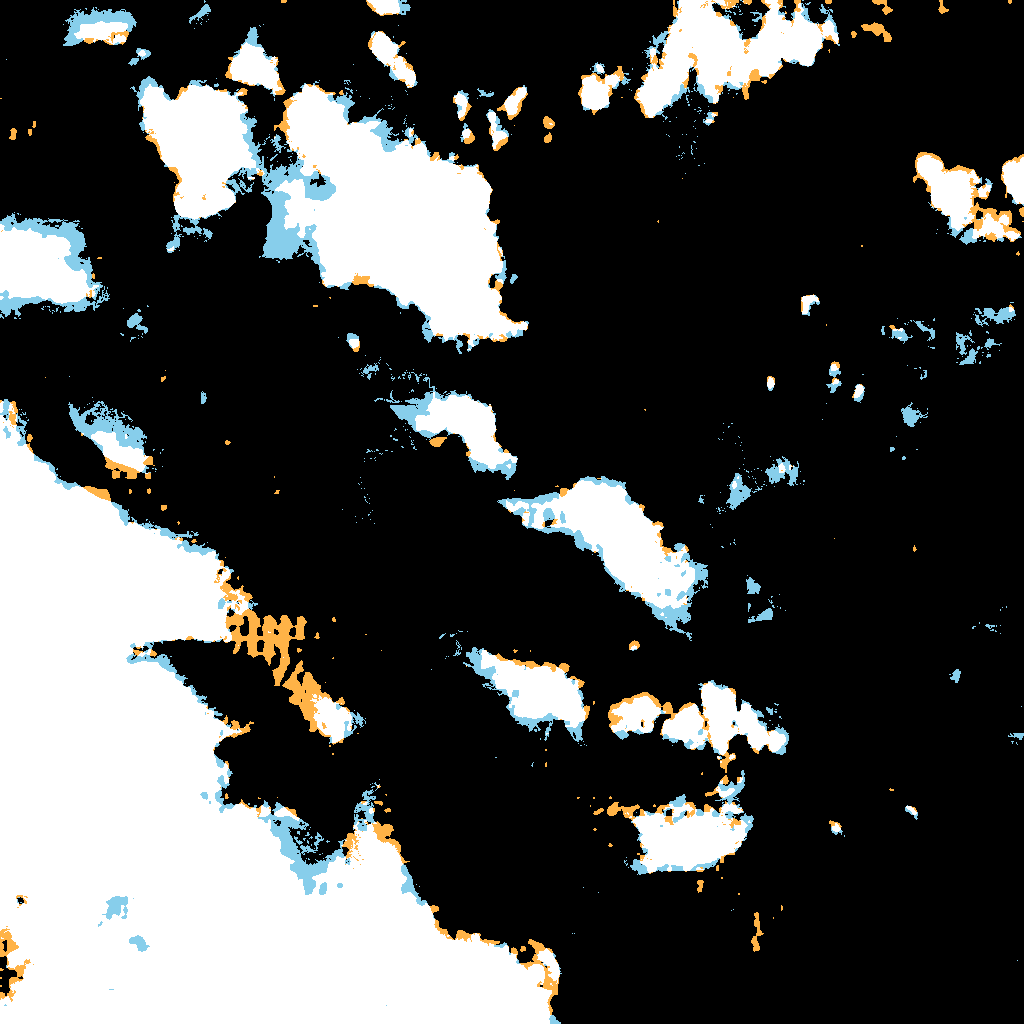} & 
        \includegraphics[width=0.16\columnwidth]{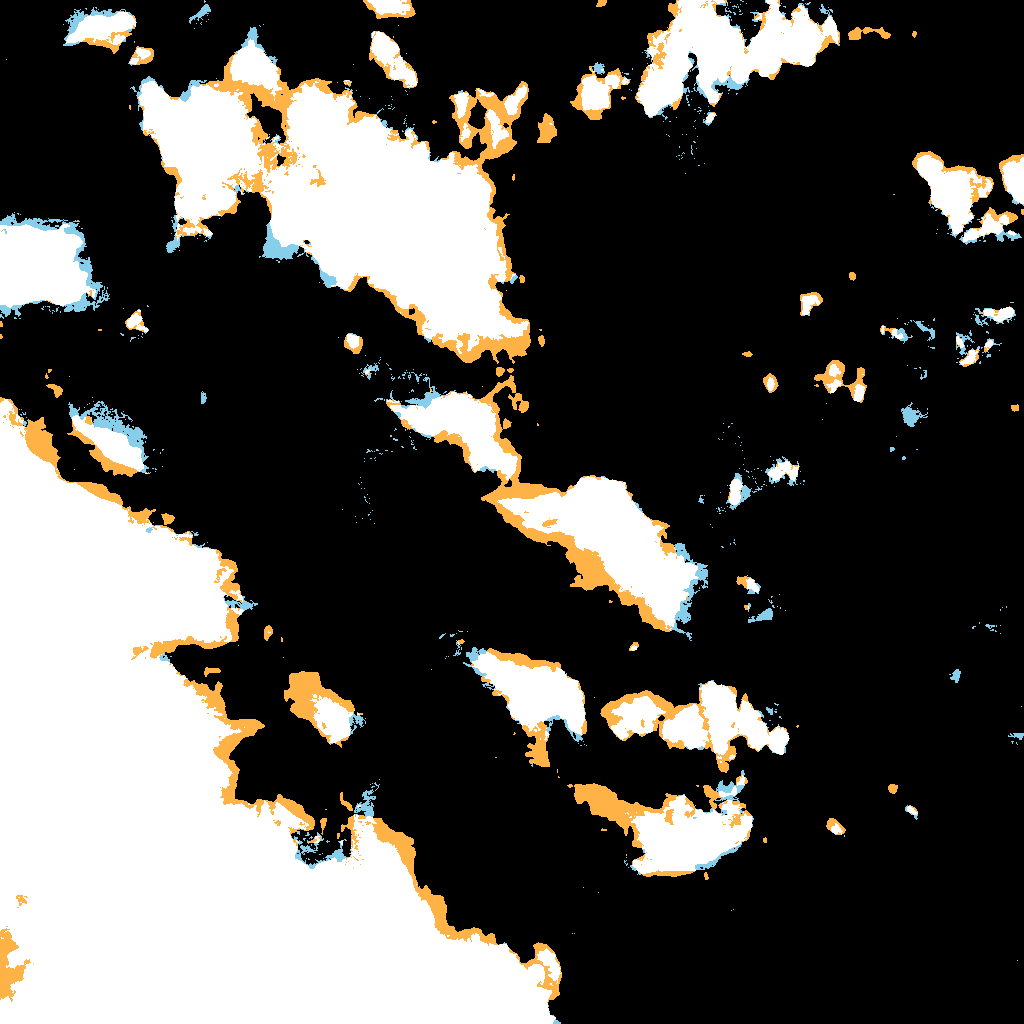} \\[-2pt] 
        
        \scriptsize Image & 
        \scriptsize Ground Truth & 
        \scriptsize \shortstack{GeoSAM-Lite} & 
        \scriptsize \shortstack{RSAM-Seg}
    \end{tabular}
    \vspace{-0.5em} 
    \caption{Visual comparison between student and teacher model.}
    \label{fig:RScomp} 
\end{figure}

We compare the resource consumption of GeoSAM-Lite (ViT-S) against the baseline RSAM-Seg (ViT-L) to validate its suitability for onboard deployment. As shown in Table~\ref{tab:lightweight_table}, GeoSAM-Lite achieves an order-of-magnitude reduction in model complexity: parameters decrease by 92.8\% and FLOPs by 92.5\%. This efficiency results from halving the network depth and condensing the embedding dimension, which directly translates to a 58.3\% reduction in GPU memory usage. Despite these aggressive simplifications, GeoSAM-Lite maintains high visual fidelity with the baseline in Fig.~\ref{fig:RScomp} and Fig. \ref{fig:sparcs_multi}, demonstrating that the Geo-Init pre-training strategy effectively bridges the gap between extreme lightweight design and segmentation accuracy, making it an ideal candidate for resource-constrained edge devices.

\begin{table}[!t]
\centering
\caption{Comparative Analysis of Model Efficiency}
\label{tab:lightweight_table}
\renewcommand{\arraystretch}{0.9}
\setlength{\tabcolsep}{7pt}  
\scriptsize

\begin{tabular}{lccc}
\toprule
\textbf{Metric} & \textbf{RSAM-Seg} & \textbf{GeoSAM-Lite} & \textbf{Reduction} \\ 
& \textbf{(ViT-L)} & \textbf{(ViT-S)} & \textbf{($\downarrow$)} \\ 
\midrule
\rowcolor[gray]{0.95} \multicolumn{4}{l}{\textit{Model Complexity Analysis}} \\
Parameters (M)      & 307.4  & 22.05 & \textbf{92.8\%} $\downarrow$ \\
FLOPs (G)     & 61.6   & 4.6   & \textbf{92.5\%} $\downarrow$ \\
Network Depth             & 24     & 12    & \textbf{50.0\%} $\downarrow$ \\
Feature Dimension         & 1024   & 384   & \textbf{62.5\%} $\downarrow$ \\
\midrule
\rowcolor[gray]{0.95} \multicolumn{4}{l}{\textit{Hardware Resource Consumption}} \\
System RAM (GB)     & 16     & 8     & \textbf{50.0\%} $\downarrow$ \\
GPU Memory (GB)     & 48     & 20    & \textbf{58.3\%} $\downarrow$ \\
Storage Space (MB)    & 1228.8$^{\dagger}$ & 595   & \textbf{51.6\%} $\downarrow$ \\
\bottomrule
\multicolumn{4}{l}{\scriptsize $^{\dagger}$ \textit{Converted from 1.2 GB for consistent comparison units.}}
\end{tabular}
\end{table}

\subsection{Generalization Analysis on Farmland Scenario}

We further evaluate GeoSAM-Lite on farmland segmentation to assess its generalization beyond cloud detection. The Sentinel-2 farmland dataset \cite{farm_dataset} contains 13 spectral channels with public releases capturing agricultural blocks in France. The $224 \times 224$ images are split 8:1:1 into 1,566 training, 200 validation, and 200 testing patches. As shown in Table \ref{tab:farmland_res}, GeoSAM-Lite achieves competitive performance against other lightweight baselines, maintaining a superior balance between efficiency and accuracy for new domains. Qualitatively, the results in Fig. \ref{fig:dis_image} reveal that GeoSAM-Lite can effectively delineate key structural footprints and boundaries of agricultural parcels.

\begin{table}[!t]
    \centering
    \caption{Quantitative Comparison of Segmentation Performance on Farmland Scenario}
    \label{tab:farmland_res}
    \setlength{\tabcolsep}{2.4pt} 
    \renewcommand{\arraystretch}{0.9} 
    \scriptsize
    \begin{tabular}{llccccc}
        \toprule
        \textbf{Dataset} & \textbf{Method} & \textbf{Jaccard} & \textbf{F1 Score} & \textbf{OA}& \textbf{Precision} & \textbf{Recall} \\ 
        \midrule
        \multirow{4}{*}{Farmland} 
        & EfficientSAM    & 0.5451 & 0.6807 & 0.7810 & \underline{0.7530} & 0.6557 \\
        & MobileSAM       & 0.5418 & 0.6799 & 0.7761 & 0.7446 & 0.6578 \\
        & RSAM-Seg & \textbf{0.6346} & \textbf{0.7592} & \textbf{0.8201} & \textbf{0.7600} & \textbf{0.7818} \\
        & GeoSAM-Lite     & \underline{0.5683} & \underline{0.7036} & \underline{0.7848} & 0.7323 & \underline{0.7079} \\
        \bottomrule
        \multicolumn{5}{l}{\textit{Note: Best results are in bold; second best are underlined.}}
    \end{tabular}
\end{table}

\begin{figure}[!t]
    \centering
    \renewcommand{\arraystretch}{0.4}
    \setlength{\tabcolsep}{1.0pt}
    \scriptsize

    \begin{tabular}{cccccc}
        \includegraphics[width=0.158\columnwidth]{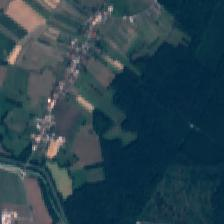} & 
        \includegraphics[width=0.158\columnwidth]{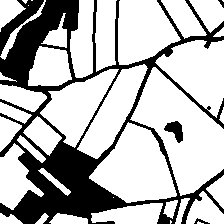} & 
        \includegraphics[width=0.158\columnwidth]{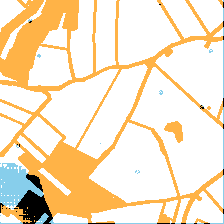} & 
        \includegraphics[width=0.158\columnwidth]{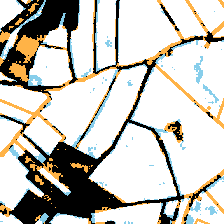} & 
        \includegraphics[width=0.158\columnwidth]{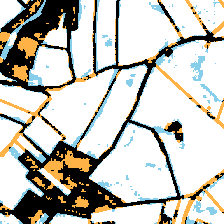} &
        \includegraphics[width=0.158\columnwidth]{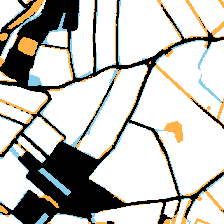} \\
        
        (a) & (b) & (c) & (d) & (e) & (f)
    \end{tabular}
    \vspace{-1em}
    \caption{Visualization comparison of farmland scenario. (a) Original image, (b) ground truth, (c) SAM (manual), (d) EfficientSAM, (e) MobileSAM, (f) GeoSAM-Lite.}
    \label{fig:dis_image}
\end{figure}

\section{Conclusion}
 
In this letter, we presented GeoSAM-Lite for efficient onboard remote sensing segmentation. Ablation results show that frequency-domain reconstruction (FFL-F) is more critical than spatial recalibration (FFL-S) for boundary fidelity, while Domain-Expert initialization (Geo-Init) proves indispensable even under full supervision. Cross-scenario experiments further confirm that the learned representations generalize from cloud detection to farmland segmentation without architectural modification. Future work will focus on narrowing the performance gap with heavyweight models on cross-sensor benchmarks and extending the framework to more RS scenarios.

\ifCLASSOPTIONcaptionsoff
\newpage
\fi

\bibliographystyle{IEEEtran}

\bibliography{references}

\end{document}